\documentclass{article}

\usepackage{microtype}
\usepackage{graphicx}
\usepackage{subfigure}
\usepackage{booktabs} 

\usepackage{hyperref}

\usepackage{arxiv2026}

\makeatletter
\icmlshowauthorstrue
\renewcommand{\Notice@String}{}  
\makeatother

\usepackage{amsmath}
\usepackage{amssymb}
\usepackage{mathtools}
\usepackage{amsthm}

\usepackage[capitalize,noabbrev]{cleveref}

\theoremstyle{plain}

\theoremstyle{definition}

\theoremstyle{remark}

\usepackage[textsize=tiny]{todonotes}

\usepackage{graphicx}
\usepackage{wrapfig}
\usepackage{amsmath} 
\usepackage{longtable}
\usepackage{multirow}
\usepackage{todonotes}
\usepackage{pifont}
\usepackage{enumitem}
\usepackage{csvsimple}
\newcommand{\SE}[1]{\scriptsize{(#1)}} 
\usepackage{makecell} 
\usepackage{xstring}
\usepackage[utf8]{inputenc}
\usepackage[T1]{fontenc}

\newcommand{\model}[1]{\texttt{#1}}
\newcommand{\Moment}{\model{MOMENT}}
\newcommand{\PatchTST}{\model{PatchTST}}
\newcommand{\Moirai}{\model{Moirai}}
\newcommand{\Chronos}{\model{Chronos}}

\newcommand{\TimesFM}{\model{TimesFM}}

\newcommand{\TSMixer}{\model{TSMixer}}
\newcommand{\TimeMixer}{\model{TimeMixer}}
\newcommand{\iTransformer}{\model{iTransformer}}
\newcommand{\MLP}{\model{MLP}}

\newcommand{\TinyTimeMixers}{\model{TinyTimeMixers}}
\newcommand{\TimerXL}{\model{Timer-XL}}
\newcommand{\Crossformer}{\model{Crossformer}}
\newcommand{\PCA}{\model{PCA}}

\newcommand{\UPtwoME}{\model{UP2ME}}
\newcommand{\SOFTS}{\model{SOFTS}}
\newcommand{\ModernTCN}{\model{ModernTCN}}
\newcommand{\TimesNet}{\model{TimesNet}}
\newcommand{\StemGNN}{\model{StemGNN}}
\newcommand{\ChronosTwo}{\model{Chronos-2}}

\newcommand{\AutoETS}{\model{AutoETS}}
\newcommand{\MICA}{\model{MICA}}
\newcommand{\TimeFilter}{\model{TimeFilter}}
\newcommand{\TimePro}{\texttt{TimePro}}
\newcommand{\cosFormer}{\texttt{cosFormer}}
\newcommand{\Performers}{\texttt{Performers}}
\newcommand{\Informer}{\texttt{Informer}}
\newcommand{\Autoformer}{\texttt{Autoformer}}
\newcommand{\Fedformer}{\texttt{Fedformer}}

\icmltitlerunning{MICA: Multivariate Infini Compressive Attention for Time Series Forecasting}

\begin{document}

\twocolumn[
\icmltitle{MICA: Multivariate Infini Compressive Attention for Time Series Forecasting}




\icmlsetsymbol{equal}{*}
\icmlsetsymbol{dagger}{\dag}

\begin{icmlauthorlist}
\icmlauthor{Willa Potosnak}{cmu}
\icmlauthor{Nina \.Zukowska}{cmu,equal}
\icmlauthor{Micha\l{} Wili\'nski}{cmu,equal}
\icmlauthor{Dan Howarth}{cmu,dagger}
\icmlauthor{Ignacy St\k{e}pka}{cmu,dagger}
\icmlauthor{Mononito Goswami}{cmu,amazon}
\icmlauthor{Artur Dubrawski}{cmu}
\end{icmlauthorlist}

\icmlaffiliation{cmu}{Carnegie Mellon University, Pittsburgh, PA, USA}
\icmlaffiliation{amazon}{Amazon, Seattle, WA, USA; this work does not relate to the author's position at Amazon}

\icmlcorrespondingauthor{Willa Potosnak}{wpotosna@cs.cmu.edu}

\vskip 0.3in
]



\printAffiliationsAndNotice{\icmlEqualContribution} 

\begin{abstract}
Multivariate forecasting with Transformers faces a core scalability challenge: modeling cross-channel dependencies via attention compounds attention's quadratic sequence complexity with quadratic channel scaling, making full cross-channel attention impractical for high-dimensional time series. We propose Multivariate Infini Compressive Attention (\MICA), an architectural design to extend channel-independent Transformers to channel-dependent forecasting. By adapting efficient attention techniques from the sequence dimension to the channel dimension, \MICA\ adds a cross-channel attention mechanism to channel-independent backbones that scales linearly with channel count and context length. We evaluate channel-independent Transformer architectures with and without \MICA\ across multiple forecasting benchmarks. \MICA\ reduces forecast error over its channel-independent counterparts by 5.4\% on average and up to 25.4\% on individual datasets, highlighting the importance of explicit cross-channel modeling. Moreover, models with \MICA\ rank first among deep multivariate Transformer and MLP baselines. \MICA\ models also scale more efficiently with respect to both channel count and context length than Transformer baselines that compute attention across both the temporal and channel dimensions, establishing compressive attention as a practical solution for scalable multivariate forecasting.
\end{abstract}

\section{Introduction}
\label{section:introduction}
Leveraging relationships across multiple time series, whether through cross-channel dependencies or exogenous covariates, has been shown to improve forecasting performance across diverse domains including weather \citep{cachay2021world}, energy \citep{OlivaresChallu2022nbeatsx, liu2025timerxllongcontexttransformersunified}, supply chain \citep{goncalves2021multivariate}, healthcare \citep{potosnak2025pkforecast}, and finance \citep{santos2013var}. For example, weather patterns observed at geographically distributed stations provide predictive signals for sudden changes at other locations that cannot be captured by seasonality or local trends alone. Similarly, regional infection rates exhibit spatial dependencies where disease outbreaks in one area serve as indicators for imminent surges in neighboring regions.

\begin{figure}[t!]
    \centering
    \includegraphics[width=0.5\textwidth, trim=65 5 80 30, clip]{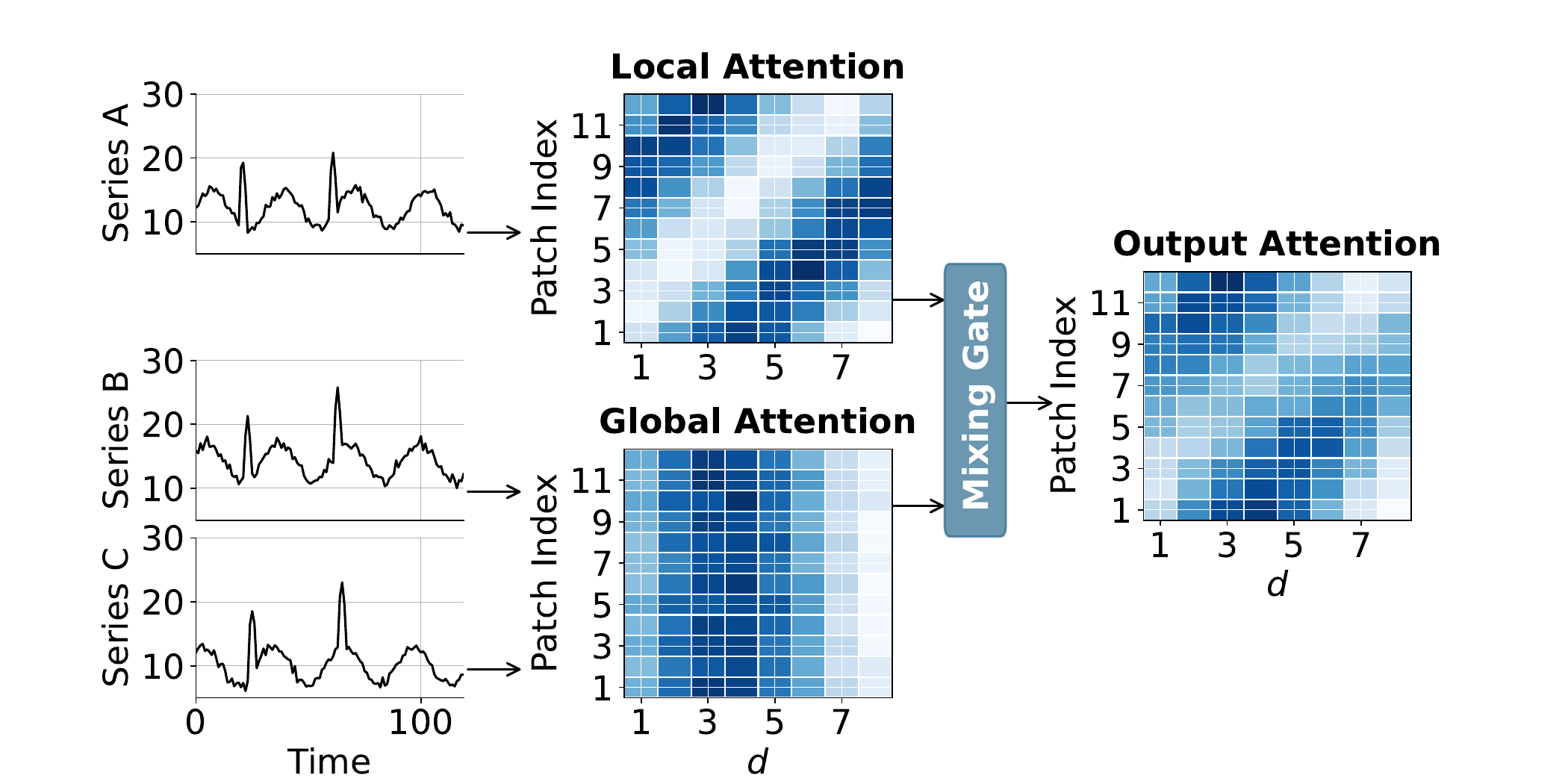}
    \caption{\MICA~consists of local attention to capture temporal dependencies and global linear attention to capture cross-channel information, combined via a mixing gate. Attention outputs are shown as patch (token) $\times$ $d$ matrices, where darker cells indicate stronger weighted values.}
    \label{fig:intro_summary}
\end{figure}

Several influential Transformer-based forecasting models treat each time series channel independently, learning shared parameters across all series without explicitly modeling cross-channel dependencies \citep{nie2023patchtst, goswami2024moment, wu2021autoformer, zhou2021informer}. Channel-independent methods already achieve strong performance, owing to their computational efficiency and robustness to spurious cross-channel correlations. Multivariate architectures must therefore demonstrate both superior accuracy and computational efficiency to justify the cost of explicit cross-channel modeling. While recent work has begun to explore cross-channel modeling for Transformer architectures \citep{liu2024itransformer, ansari2025chronos2}, scalable and effective methods for capturing these dependencies remain an open challenge. Specifically, extending attention to cross-channel dependencies compounds its quadratic sequence complexity with quadratic channel scaling, making full cross-channel attention impractical for high-dimensional time series. Thus, computationally efficient methods must be leveraged for channel-dependent Transformers to be practical in real-world, large-dimensional settings.

This challenge mirrors the long-context problem in large language models (LLMs), where quadratic sequence complexity limits the context window Transformers can effectively process. Recent advances in LLMs have attempted to address this problem through compressive memory techniques that enable the processing of arbitrarily long contexts \citep{munkhdalai2024leave}. Drawing inspiration from this parallel, we propose Multivariate Infini Compressive Attention (\MICA), an architectural design to extend channel-independent Transformers to channel-dependent forecasting. By adapting efficient attention techniques used in Infini-Attention from the context dimension to the channel dimension, \MICA\ adds a cross-channel attention mechanism to channel-independent backbones that scales linearly with channel count and context length.

\begin{table*}[t!]
\centering
\caption{Comparison of multivariate time series forecasting methods by cross-channel processing technique and complexity. Complexity refers to how computational cost scales with the context length ($P$ tokens/patches) and number of channels $C$. Complexities assume constant hidden dimensions and number of layers.} 
\resizebox{1.0\textwidth}{!}{\begin{tabular}{lllll}
\toprule
\textbf{Method} & \textbf{Architecture Type} & \textbf{Cross-Channel Method} & \textbf{Application Stage} & \textbf{Complexity} \\
\midrule
Decorrelation~\citep{kessy2018optimal, zheng2024mvgcrps} & Universal & \PCA & Preprocessing & $\mathcal{O}(PC^2 + C^3)$ \\
Decorrelation~\citep{ica2003multivarfcst} & Universal & \texttt{FastICA} & Preprocessing & $\mathcal{O}(PC^2)$ \\
\midrule
\MLP-Multivariate~\citep{rosenblatt1958_mlp} & MLP & Dense Layer Mixing & Model Encoder & $\mathcal{O}(PC)$ \\
\TSMixer~\citep{chen2023tsmixer} & MLP & Channel-wise Linear Projection & Model Encoder & $\mathcal{O}(PC)$ \\
\TinyTimeMixers~\citep{ekambaram2024ttms} & MLP & Channel-wise Linear Projection & Model Encoder & $\mathcal{O}(PC)$ \\
\TimeMixer~\citep{wang2023timemixer} & MLP & Multi-scale Mixing & Model Encoder & $\mathcal{O}(P^2C)$ \\
\TimeMixer$++$~\citep{wang2024timemixer++} & MLP & Channel-wise Attention & Model Encoder & $\mathcal{O}(P^2C + PC^2)$ \\
\SOFTS~\citep{han2024softs} & MLP & Global channel-wise representation & Model Encoder & $\mathcal{O}(PC)$ \\
\midrule
\ModernTCN~\citep{luo2024moderntcn} & CNN & Depthwise-Separable Convolutions & Model Encoder & $\mathcal{O}(PC)$ \\
\TimesNet~\citep{wu2023timesnettemporal2dvariationmodeling} & CNN & 2D Convolutions on Reshaped Tensors & Model Encoder & $\mathcal{O}(P \log(P)C)$ \\
\midrule
\UPtwoME~\cite{zhangup2me} & Graph & Graph Transformer Layers & Model Encoder & $\mathcal{O}(C^2 + P^2C)$ \\
\StemGNN~\citep{cao2020stemgnn} & Graph & Spectral Graph Convolution & Model Encoder & $\mathcal{O}(C^3 + PC^2 + P\log(P)C)$ \\
\TimeFilter~\citep{hu2025timefilter} & Graph & Patch-Specific Spatial-Temporal Graph Filtration & Model Encoder & $\mathcal{O}(P^2C + PC^2)$ \\
\midrule
\TimePro~\citep{timepro2025}& State-Space & Channel-wise Hyper-state & Model Encoder & $\mathcal{O}(PC)$ \\
\midrule
\Moirai~\citep{woo2024moirai} & Transformer & Channel token concatenation & Tokenization & $\mathcal{O}(P^2C^2)$ \\
\TimerXL~\citep{liu2025timerxllongcontexttransformersunified} & Transformer & Channel token concatenation & Tokenization & $\mathcal{O}(P^2C^2)$ \\
\iTransformer~\citep{liu2024itransformer} & Transformer & Channel-wise Attention & Channel Inversion & $\mathcal{O}(PC^2)$ \\
\Crossformer~\citep{zhang2023crossformer} & Transformer & Cross-channel attention & Attention Module & $\mathcal{O}(P^2C)$ \\
\ChronosTwo~\citep{ansari2025chronos2} & Transformer & Cross-channel attention & Attention Module & $\mathcal{O}(P^2C + PC^2)$ \\
\texttt{Correlated Attention}~\citep{nguyen2023correlatedattentiontransformers} & Transformer & Lagged Cross-Correlation (FFT) & Attention Module & $\mathcal{O}(P \log(P)C^2)$ \\
\midrule
\textbf{\MICA\ (Ours)} & \textbf{Transformer} & \textbf{Channel-Aware Compressive Attention} & \textbf{Attention Module} & $\mathbf{\mathcal{O}(P^2C + PC)}$ \\
\bottomrule
\end{tabular}}
\label{tab:prior_work_comparison}
\end{table*}

Our main contributions are as follows:
\begin{enumerate}[label=(\roman*)]
    \item \textbf{\MICA~(Multivariate Infini Compressive Attention).} An architectural design to extend channel-independent Transformers to channel-dependent forecasting. \MICA\ adds a cross-channel attention mechanism to channel-independent backbones that scales linearly with channel count and context length.

    \item \textbf{Cross-Channel Mechanism Taxonomy.} We present a taxonomy of multivariate forecasting methods organized by cross-channel processing technique and complexity in terms of channel count and context length (Table~\ref{tab:prior_work_comparison}), enabling comparison of computational trade-offs.

    \item \textbf{Multivariate Forecasting Benchmark.} We curate a diverse benchmark spanning climate, energy, traffic, and healthcare domains, including two contributed high-frequency meteorological datasets from the Earth Observing Laboratory Field Data Archive with measurements across distributed stations. Models with \MICA\ rank first among deep Transformer and MLP baselines. \MICA\ models also scale more efficiently with respect to both channel count and context length than Transformer baselines that compute attention across both the temporal and channel dimensions.
    
    \item \textbf{\MICA\ Design Ablation Studies.} We ablate key design choices in \MICA, including channel exclusion and weighting strategies for the global linear attention component, and nonlinear gating mechanisms.
\end{enumerate}

\section{Related Work}
\label{section:related_work}

\paragraph{Channel-independent Adapters.}
Approaches that leverage \textit{Adapters} generally learn representations for each channel using a channel-independent backbone and then combine them using a multi-channel adapter. Examples include multivariate decoders~\cite {ekambaram2024ttms} and Graph Transformer layers~\cite{zhangup2me}.

\paragraph{Channel-dependent Forecasting Models.}
Alternatively, Channel-dependent forecasting methods incorporate channel mixing into the model architecture. For instance, both \TimerXL~\citep{liu2025timerxllongcontexttransformersunified} and \Moirai~\citep{woo2024moirai} flatten 2D multivariate data into a single 1D token sequence and apply attention over this expanded set, incurring $\mathcal{O}(P^2C^2)$ complexity; however, \TimerXL\ employs an attention mask to preserve causal dependencies, whereas \Moirai\ applies standard unmasked attention over concatenated channel tokens. \iTransformer~\citep{liu2024itransformer} inverts the problem by attending to channel tokens rather than temporal tokens, achieving $\mathcal{O}(PC^2)$ complexity. \Crossformer~\citep{zhang2023crossformer} employs a two-stage attention design with dedicated cross-time and cross-dimension layers, using a router mechanism with learnable vectors to reduce cross-dimension complexity from $\mathcal{O}(P^2C^2)$ to $\mathcal{O}(P^2C)$. \ChronosTwo~\citep{ansari2025chronos2} uses time attention ($\mathcal{O}(P^2C)$) in addition to group attention ($\mathcal{O}(PC^2)$) to model cross-channel dependencies, yielding $\mathcal{O}(P^2C + PC^2)$ total complexity. Among these methods, \MICA\ models are the only multivariate Transformers that scale linearly with channel count; \Crossformer\ also achieves linear scaling, but only when its learnable vector count satisfies $R \ll C$. We provide a detailed comparison of these methods in Appendix~\ref{section:apd_related_work_comparison}.

Beyond Transformers, MLP-based approaches employ various channel mixing strategies. 
\TSMixer~\citep{chen2023tsmixer} and \TinyTimeMixers~\citep{ekambaram2024ttms} achieve $\mathcal{O}(PC)$ 
complexity through different mechanisms: \TSMixer\ alternates between time-mixing 
and feature-mixing layers; \TinyTimeMixers\ is a foundation model based on the 
\TSMixer\ architecture. \TimeMixer\ achieves $\mathcal{O}(P^2C)$ 
complexity by applying multi-scale temporal mixing with decomposition. \TimeMixer\texttt{++} is a variation of \TimeMixer\ with channel-wise attention achieving $\mathcal{O}(P^2C + PC^2)$ overall. \SOFTS\ aggregates channels into a core representation before redistributing 
global information with $\mathcal{O}(PC)$ complexity. Table~\ref{tab:prior_work_comparison} presents a taxonomy organized by 
cross-channel mechanism and complexity. We discuss convolution-based, state-space, and graph-based models in Appendix~\ref{section:apd_other_multivariate_architectures}.

\paragraph{Infini Attention.} 
Infini-Attention~\cite{munkhdalai2024leave} is a technique leveraging 
compressive memory to enable LLMs to process long contexts efficiently. Infini-Attention combines two mechanisms within a single Transformer block: a scaled dot-product attention mechanism for local attention over the current context segment and a linear attention mechanism to capture and update a compressive memory state from prior context segments. This memory is incrementally updated and queried when processing each new context segment to retrieve relevant long-range information. The scaled dot-product and memory-based attention outputs are then combined via learnable mixing weights. More information on Infini-Attention is in Appendix~\ref{apd:infini_attn_background}. 

While Infini-Attention provides a blueprint for efficient compression via linear attention, it was developed for long-context language modeling and does not address multivariate forecasting. Adapting it requires modifications. Extending our prior work \citep{infinichannelmixer}, our approach makes four key modifications: (1) we adapt the use of linear attention along the channel dimension to perform channel mixing and compression (2) we remove the memory update mechanism since common forecasting implementations use window-sampling during training which does not ensure sequential temporal context across forward passes required to leverage memory~\citep{olivares2022library_neuralforecast, gluonts2019, tslib2024}, (3) we introduce new design choices including channel exclusion mechanisms and channel weighting strategies for the global linear attention component, and (4) we propose multiple new gating mechanisms to control the mixing of local and global attention outputs. We include an extended discussion comparing alternative efficient attention methods as candidates for multivariate channel mixing and compression in Appendix~\ref{section:apd_efficient_attention_architectures}.

\section{Methods}
\label{section:methods}
In this section, we first formalize the multi-horizon forecasting task from both univariate and multivariate perspectives. We then present our proposed approach, \MICA.

\begin{figure*}[t!]
    \centering
    \includegraphics[width=0.8\textwidth, trim=0 7 0 170, clip]{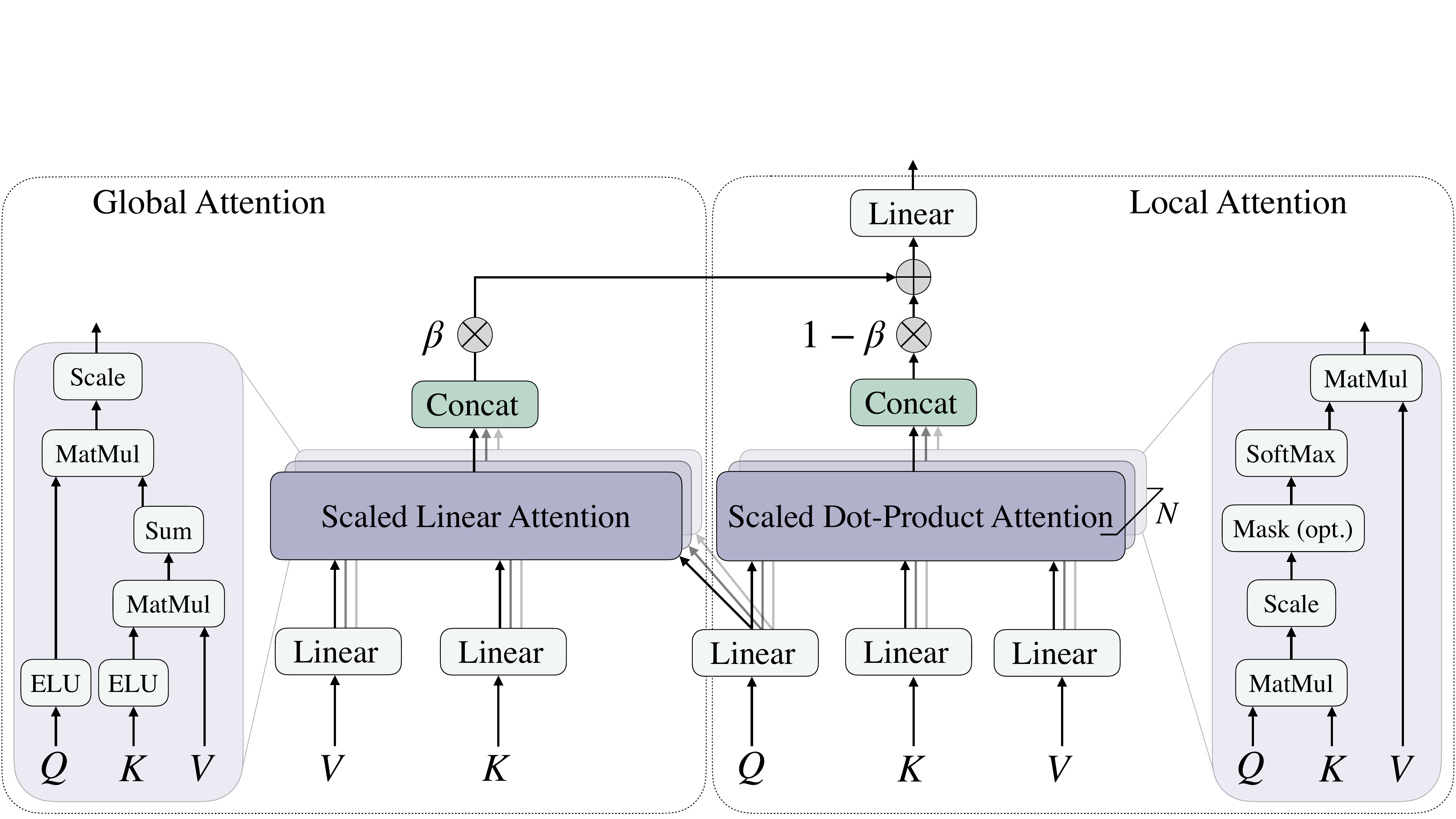}
    \caption{\MICA\ is an attention-based architectural design proposed to model both local patterns and global cross-channel interactions. $\textbf{Q}$, $\textbf{K}$, and $\textbf{V}$ denote the query, key and value matrices and $N$ denotes the number of attention heads. \MICA\ consists of three complementary components: (1) a scaled dot-product attention module that models detailed temporal relationships locally within individual time series, (2) a linear attention module that efficiently aggregates information globally across channels, and (3) a learnable attention mixing gate that adaptively balances local and global information in a computationally efficient manner. \MICA\ variants, such as the non-linear \MLP\ gate, are discussed in Appendix~\ref{apd:infini_gate_variants}.}
    \label{fig:method_infini_main}
\end{figure*}

\subsection{Multi-horizon Forecasting}
We begin by comparing a multi-horizon forecasting task in univariate and multivariate settings. The multi-horizon forecasting task refers to the forecasting $H$ timesteps in the future at time $t$. Consider a multivariate time series~$\mathbf{Y} \in \mathbb{R}^{B \times C \times T}$, where~$B$ is the window batch size,~$C$ is the number of channels and~$T$ is the total number of time steps. Each channel~$c \in \{1, \dots, C\}$ represents unique variate, such as a distinct sensor.

\paragraph{Univariate Forecasting.} 
In the univariate setting, we forecast each channel independently without modeling cross-channel dependencies. The forecasting task is:
\begin{equation}\label{eqn:fcst_task}
    \mathbf{\hat{Y}}_{t+1:t+H} = f(\mathbf{Y}_{t-L+1:t}),
\end{equation}
where~$\mathbf{Y}_{t-L+1:t} \in \mathbb{R}^{B \times C \times L}$ represents the context window of length~$L$. Although the data retains its $B \times C \times L$ shape, each channel is processed independently (the channel dimension is absorbed into the batch dimension during processing, effectively treating it as a flattened batch of size $BC$). The forecast output $\mathbf{\hat{Y}}_{t+1:t+H} \in \mathbb{R}^{B \times C \times H}$ of horizon length~$H$ is produced independently for each channel using the model~$f$.

\paragraph{Multivariate Forecasting.}
In the multivariate setting, we leverage information across all~$C$ channels to model cross-channel dependencies. The forecasting task is the same as in Eqn.~\ref{eqn:fcst_task}, but now~$\mathbf{Y}_{t-L+1:t} \in \mathbb{R}^{B \times C \times L}$ preserves the channel dimension rather than reshaping channels into the batch. Correspondingly, the model~$f$ is equipped with architectural components to leverage this channel dimension and learn cross-channel dependencies. The model produces forecast output $\mathbf{\hat{Y}}_{t+1:t+H} \in \mathbb{R}^{B \times C \times H}$ for all channels.

Training loss is computed and averaged over all predictions. The objective can be specified using standard forecasting loss functions. We use a pointwise loss function following prior work \citep{goswami2024moment, nie2023patchtst}, opting for mean absolute error (MAE) over mean squared error (MSE) as it is more robust to outliers and provides interpretable error metrics in the original data units:
\begin{equation}\label{eqn:mae}
    \mathcal{L}(\mathbf{Y}, \mathbf{\hat{Y}}) = \frac{1}{B C H} \sum_{b=1}^{B} \sum_{c=1}^{C} \sum_{h=1}^{H} |\mathbf{Y}_{b, c, t+h} - \mathbf{\hat{Y}}_{b, c, t+h}|
\end{equation}

\subsection{Multivariate Infini Compressive Attention (\MICA)}
Our proposed architecture models both local channel-independent patterns and global cross-channel dependencies. \MICA\ consists of three components: (1) the context-wise (local) attention mechanism of the channel-independent backbone, which models temporal relationships within individual time series, (2) a channel-wise (global) linear attention mechanism that efficiently aggregates information across channels, and (3) a learnable gate that adaptively balances the two attention outputs.

\paragraph{Local Attention.}
Channel-independent backbones such as \PatchTST~\citep{nie2023patchtst} adopt scaled dot-product attention~\citep{vaswani_2021_attentionisallyouneed} for local temporal modeling, outperforming more efficient variants such as \Informer~\citep{zhou2021informer}, \Autoformer~\citep{wu2021autoformer}, and \Fedformer~\citep{zhou2022fedformer}. Since \MICA\ extends these backbones, we retain this as the local temporal attention mechanism, operating independently per channel:
\begin{align}
\textbf{A}_{\text{local}} = \text{softmax}\left(\frac{\textbf{Q}\textbf{K}^\top}{\sqrt{d_k}}\right)\textbf{V},
\end{align}
where $\mathbf{Q} \in \mathbb{R}^{B \times C \times N \times P \times d_q}$, $\mathbf{K} \in \mathbb{R}^{B \times C \times N \times P \times d_k}$, and $\mathbf{V} \in \mathbb{R}^{B \times C \times N \times P \times d_v}$ denote the query, key, and value matrices obtained by passing the encoder inputs $\textbf{x}\in \mathbb{R}^{B \times C \times P \times d}$ through linear projection layers $\mathbf{W}_Q$, $\mathbf{W}_K$, and $\mathbf{W}_V$, respectively. Here, $N$ denotes the number of attention heads, $P$ denotes the number of tokens, and $d_q$, $d_k$, and $d_v$ represent the per-head hidden dimensions of the query, key, and value vectors, respectively. Multi-head scaled dot-product attention is illustrated in Fig.~\ref{fig:method_infini_main} with attention output $\mathbf{A}_{\text{local}} \in \mathbb{R}^{B \times C \times N \times P \times d_v}$.

\paragraph{Global Attention.}
To capture cross-channel dependencies, we adapt the linear attention mechanism from Infini-Attention~\citep{munkhdalai2024leave}, originally designed to compress information across text beyond context length, to instead compress information across channels. This approach computes attention output~$\mathbf{A}_{\text{global}}$ that aggregates information from all channels, with computation that scales linearly rather than quadratically with context length. The key computational advantage comes from avoiding the~$\mathbf{Q}\mathbf{K}^\top$ matrix multiplication: instead, we directly compute~$\mathbf{K}^\top\mathbf{V}$ across all channels and then use the query matrix to retrieve cross-channel information. Formally, $\mathbf{M}$ aggregates keys and values from $C$ channels:
\begin{align} 
\label{eq:M_z_update}
    \mathbf{M} &= \sum_{c=1}^C \phi(\mathbf{K}^{(c)})^\top \mathbf{V}^{(c)}, \\
    \mathbf{z} &= \sum_{c=1}^C \sum_{p=1}^{P} \phi(\mathbf{K}^{(c)}_{p}),\\
    \mathbf{A}_{\text{global}} &= \frac{\phi(\mathbf{Q})\mathbf{M}}{\phi(\mathbf{Q})\mathbf{z} + \epsilon}.
\end{align}
where~$\phi$ corresponds to the nonlinear activation $\text{ELU}(\mathbf{X}) + 1$, $\mathbf{M} \in \mathbb{R}^{B \times 1 \times N \times d_k \times d_v}$ is the global context matrix aggregating key-value information across all channels, $\mathbf{z} \in \mathbb{R}^{B \times 1 \times N \times d_k \times 1}$ is a normalization term accumulated across all channels and their sequence positions, and~$\epsilon$ is a small constant for numerical stability \cite{munkhdalai2024leave, pmlr-v119-katharopoulos20a}. Multi-head linear attention is shown in Fig.~\ref{fig:method_infini_main} with global attention output $\textbf{A}_{\text{global}} \in \mathbb{R}^{B \times C \times N \times P \times d_v}$.

While Infini-Attention maintains a memory state $\mathbf{M}$ to incorporate historical context across forward passes as described in Appendix~\ref{apd:infini_attn_background}, oour approach recomputes $\mathbf{M}$ independently for each forward pass without persisting or updating it. This aligns with common forecasting libraries that use window-sampling during training, which does not guarantee sequential temporal context across forward passes, as required to leverage memory~\citep{olivares2022library_neuralforecast, gluonts2019, tslib2024}. Additionally, more recent architectural designs for model training, such as forking-sequences~\citep{potosnak2025forkingsequences}, support processing entire series in a single forward pass, removing the need for persistent temporal memory altogether.

\paragraph{Global-Local Attention Mixing Gate.}
The final component of \MICA~is an attention mixing gate that balances local quadratic attention and global linear attention to produce the final attention output $\mathbf{A}_{\text{mixed}} \in \mathbb{R}^{B \times C \times N \times P \times d_v}$ as shown in Fig.~\ref{fig:intro_summary}. We evaluate two types of mixing gates: linear and non-linear. 

The linear mixing gate aligns with the original infini-attention implementation \citep{munkhdalai2024leave} and consists of a learnable parameter~$\beta$ that balances local and global attention contributions as shown in Fig.~\ref{fig:method_infini_main}:
\begin{equation}
   \mathbf{A}_{\text{mixed}} = \sigma(\beta) \odot \mathbf{A}_{\text{global}} + (1 - \sigma(\beta)) \odot \mathbf{A}_{\text{local}}.
\end{equation}
We follow prior work \citep{munkhdalai2024leave} and initialize $\beta \in \mathbb{R}^{1 \times 1 \times N \times 1 \times 1}$ with one learnable parameter per head by sampling from $\mathcal{U}(0, 10^{-2})$ for each attention block. We additionally center $\beta$ across attention heads to ensure balanced initial mixing preferences:
\begin{align}
\hat{\beta}_k = \beta_k - \mathbb{E}_k[\beta_k] \quad \text{for each head } k \in \{1, \dots, N\}.
\end{align}
We pass $\beta$ through a sigmoid function $\sigma$ to ensure the mixing weights are between 0 and 1. The $\beta$ parameters can be extended to other configurations which we discuss in Section~\ref{section:mica_ablation_study}.

Alternatively, we explore novel mixing methods, such as a non-linear \MLP-based gate that computes channel-specific mixing weights conditioned on the attention outputs:
\begin{equation}
   \mathbf{A}_{\text{mixed}} = \text{MLP}(\mathbf{A}_{\text{global}}, \mathbf{A}_{\text{local}}).
\end{equation}
In the simplest case, the \MLP~takes both local and global attention outputs as input. We additionally explore extensions that incorporate query information~$\mathbf{Q}$ for context-guided mixing, enabling the gate to adapt based on the semantic content of the queries:
\begin{equation}
   \mathbf{A}_{\text{mixed}} = \text{MLP}(\mathbf{A}_{\text{global}}, \mathbf{A}_{\text{local}}, \mathbf{Q}).
\end{equation}
The mixed attention output $\mathbf{A}_{\text{mixed}}$ from the final transformer layer is processed through a decoder to produce the final forecast,
\begin{align}
    \mathbf{\hat{Y}} = \text{Decoder}(\mathbf{A}_{\text{mixed}}),
\end{align}
where $\text{Decoder}$ can be a lightweight linear projection as in \Moment\ and \PatchTST, or a more complex architecture.

\paragraph{Complexity.}
For the standard \MICA, the local attention mechanism operates as standard quadratic self-attention with time and memory complexity $\mathcal{O}(P^2C)$, where $P$ is the number of tokens (patches). The global cross-channel attention has linear complexity $\mathcal{O}(PC)$ for both constructing the memory matrix via $\mathbf{K}^\top\mathbf{V}$ and querying it. Combining both, \MICA\  complexity is thus $\mathcal{O}(P^2C + PC)$. The local attention component is compatible with FlashAttention~\citep{dao2022flashattention} for further memory efficiency. We also provide an algorithm sketch for more memory-efficient global attention in Appendix~\ref{section:apd_flashatt}. Alternatively, the global attention memory complexity can be reduced to $\mathcal{O}(P^2C + P)$ with a time-memory tradeoff by processing channels sequentially.

\section{Experiments}
\label{section:experiments}
Our empirical evaluation addresses the following research questions. (RQ1) Does \MICA\ reduce forecasting error compared with its channel-independent counterparts? We compare \PatchTST-\MICA\ and \Moment-\MICA\ against channel-independent Transformer baselines (\PatchTST\ and \Moment) to assess whether incorporating cross-channel information through compressive attention enhances forecasting performance across diverse datasets. (RQ2) How does \MICA\ compare to state-of-the-art multivariate methods in both empirical performance and efficiency? We evaluate \MICA\ against multiple multivariate deep learning models (\iTransformer, \iTransformer-\texttt{T5}, \Crossformer, \TimerXL, \TSMixer, \TimeMixer, Multivariate \MLP, \Chronos\texttt{-2}) to assess whether \MICA\ achieves competitive performance and efficiency relative to state-of-the-art methods. (RQ3) How does \MICA's cross-channel attention compare to alternative strategies for processing multivariate data without modifying the channel-independent backbone? We compare \MICA\ against channel-independent backbones (\PatchTST\ and \Moment) augmented with a channel decorrelation-based preprocessing adapter and a multivariate output layer. (RQ4) Which design choices impact \MICA's performance? Through ablation studies, we examine the impact of linear attention channel exclusion and channel weighting strategies as well as different attention gating types to identify the most effective architectural configurations.

\subsection{Data}
We demonstrate \MICA~in two complementary settings: (1) modeling a single unique identifier across multiple variates (e.g., air temperature, atmospheric pressure, humidity, and wind direction for a single weather station), and (2) modeling a single variate across multiple unique identifier series (e.g., windspeed measurements across different weather stations). An exciting direction for future work is extending \MICA\ to combine both settings: modeling multiple variates for multiple unique identifiers (e.g., temperature, humidity, and pressure across multiple weather stations). We leave this as the next step in exploring and scaling \MICA~capabilities. 

We select datasets that meet the above criteria, including: ETT1, ETT2, Jena Weather, COVID Deaths, Loop-Seattle, Solar, and M-Dense from the Gift-Eval repository \citep{aksu2024giftevalbenchmark}. We also supplement these Gift-Eval datasets with an additional healthcare dataset containing simulated blood glucose data. Additionally, we obtain and preprocess two datasets from the Earth Observing Laboratory Field Data Archive \citep{ISU_SMEX02_AWOS_2005, ISU_IHOP2002_AWOS_2008, ISU_PLOWS_AWOS_2010} that contain 1-minute measurements of wind speed across geographically distinct stations. These open-source datasets provide well-suited test cases for multivariate modeling across geographically distributed stations, where accurate wind forecasting requires capturing cross-channel spatial dependencies from upwind measurement stations. More information on datasets is provided in Appendix~\ref{section:apd_data}.

\subsection{Models}
We integrate \MICA\ into two encoder-only, channel-independent, patch-based Transformer 
architectures: \Moment~\citep{wolf2020transformers} and \PatchTST, representative of the 
dominant tokenization strategy in modern TSFMs~\citep{goswami2024moment, ansari2025chronos2, 
woo2024moirai, das2024timesfm}. All models are implemented in \texttt{NeuralForecast} 
\citep{olivares2022library_neuralforecast} with aligned design choices (standardization, 
positional encodings, output projection); details are in Appendix~\ref{section:apd_training}. 
We compare against: \textbf{univariate Transformers} (\Moment, \PatchTST); \textbf{multivariate 
Transformers} (\iTransformer~\citep{liu2024itransformer}, \iTransformer-\texttt{T5} 
\citep{wolf2020transformers}, \Crossformer~\citep{zhang2023crossformer}, 
\TimerXL~\citep{liu2025timerxllongcontexttransformersunified}, \texttt{Chronos-2 zero-shot}~\citep{ansari2025chronos2}); \textbf{MLPs} (\TSMixer~\citep{chen2023tsmixer}, 
\TimeMixer~\citep{wang2023timemixer}, \MLP~\citep{rosenblatt1958_mlp}); and \textbf{statistical} 
(\AutoETS~\citep{ets_2008}).

\newcommand{\prevdatasetone}{}
\begin{table*}[ht!]
\centering
\caption{Forecasting MAE averaged over 5 random seeds with standard deviation in parentheses. Methods without standard deviation have deterministic solutions. \MICA~results correspond to the MLP-Query Gate. Best results are shown in \textbf{bold}. Second best results are \underline{underlined}. \textcolor{blue}{Blue} results indicate lower forecast error of \MICA~compared with the univariate model counterpart. The average rank across datasets for deep learning models is presented at the bottom of the table, with the best result in \textbf{bold} and second best \underline{underlined}.}
\scriptsize
\resizebox{1.0\textwidth}{!}{
\begin{tabular}{ll|ccccccccccccc}
\toprule
\textbf{Dataset} & \textbf{Freq.} & \textbf{\Moment} & \textbf{\Moment-\MICA} & \textbf{\PatchTST} & \textbf{\PatchTST-\MICA} & \textbf{\iTransformer} & \textbf{\iTransformer-T5} & \textbf{\Crossformer} & \textbf{\TimerXL} &\textbf{\TSMixer} & \textbf{\TimeMixer} & \textbf{\MLP} & \textbf{\texttt{\Chronos-2}} & \textbf{\AutoETS} \\
\midrule
\addlinespace[6pt]
\begin{filecontents*}{tables/mae_main.csv}
Dataset,Frequency,moment_u,moment_u_se,moment_m,moment_m_se,patchtst_u,patchtst_u_se,patchtst_m,patchtst_m_se,itransformer_m,itransformer_m_se,itransformer_t5_m,itransformer_t5_m_se,crossformer_m,crossformer_m_se,timerxl_m,timerxl_m_se,tsmixer_m,tsmixer_m_se,timemixer_m,timemixer_m_se,mlp_m,mlp_m_se,chronos_m,chronos_m_se,autoets,autoets_se
Simglucose,5min,5.136,0.057,\textcolor{blue}{4.347},0.035,5.593,0.577,\textcolor{blue}{\underline{4.241}},0.097,4.261,0.085,4.446,0.036,\textbf{4.164},0.201,4.662,0.067,6.458,0.048,6.670,0.074,8.220,0.029,8.835,-,9.362,-
COVID Deaths,D,157.292,25.536,\textcolor{blue}{104.172},15.977,141.437,43.419,\textcolor{blue}{135.718},40.678,297.885,247.687,165.436,52.466,156.161,65.810,174.528,59.192,561.950,200.411,96.065,1.143,483.689,6.254,\underline{93.739},-,\textbf{91.579},-
Iowa IHOP SMEX02,5min,1.733,0.005,\textcolor{blue}{1.666},0.005,1.765,0.019,\textcolor{blue}{\underline{1.662}},0.008,1.953,0.003,1.694,0.007,1.746,0.059,1.939,0.010,\textbf{1.661},0.002,1.710,0.008,1.821,0.012,1.776,-,1.781,-
Iowa PLOWS,5min,1.369,0.002,\textcolor{blue}{\underline{1.332}},0.005,1.382,0.009,\textcolor{blue}{\textbf{1.327}},0.006,1.517,0.003,1.344,0.004,1.358,0.009,1.487,0.012,1.334,0.002,1.351,0.001,1.479,0.004,1.431,-,1.405,-
Jena Weather,H,9.682,0.131,\textcolor{blue}{\textbf{9.387}},0.097,9.911,0.225,\textcolor{blue}{9.543},0.298,10.794,0.094,10.364,0.133,\underline{9.503},0.109,10.520,0.102,13.941,0.169,13.125,0.200,13.483,0.158,9.591,-,15.030,-
,D,\underline{13.155},0.438,14.907,0.271,14.005,0.185,\textcolor{blue}{13.799},0.328,15.057,0.417,13.670,0.502,14.484,0.638,14.224,0.340,16.538,0.664,15.569,0.954,18.853,0.437,13.857,-,\textbf{13.057},-
M-DENSE,H,92.637,0.916,\textcolor{blue}{\textbf{87.412}},1.880,95.861,2.105,\textcolor{blue}{\underline{88.020}},1.220,93.169,0.308,95.760,0.399,173.078,72.218,95.153,0.420,103.719,0.726,107.922,0.717,117.729,0.565,121.965,-,163.191,-
,D,52.927,0.849,\textcolor{blue}{51.740},0.413,53.723,0.236,\textcolor{blue}{51.959},0.203,52.650,0.805,\underline{49.861},0.174,51.195,2.098,51.891,0.611,50.529,0.504,58.068,1.774,55.281,0.266,\textbf{43.167},-,50.577,-
Loop-Seattle,D,3.010,0.007,\textcolor{blue}{\textbf{2.939}},0.014,3.246,0.011,\textcolor{blue}{3.009},0.003,3.496,0.030,3.127,0.031,3.158,0.016,3.215,0.019,3.115,0.042,2.962,0.047,3.045,0.017,\underline{2.944},-,3.032,-
ETT1,H,5.683,0.055,5.851,0.132,5.454,0.035,\textcolor{blue}{\underline{5.403}},0.199,6.010,0.043,5.758,0.053,5.682,0.186,5.806,0.060,5.506,0.045,6.346,0.109,5.747,0.075,\textbf{5.258},-,12.092,-
,D,157.189,2.469,\textcolor{blue}{157.090},1.833,155.534,2.508,157.871,2.663,163.198,2.165,168.808,2.789,208.948,23.234,\underline{154.405},1.135,185.805,3.861,162.778,5.350,185.507,5.402,\textbf{144.588},-,165.377,-
,W,982.317,37.048,994.716,34.867,1003.563,29.109,\textcolor{blue}{\underline{959.206}},42.454,998.254,46.672,1027.069,49.182,1252.076,121.811,1000.988,40.177,1076.143,48.788,980.674,37.099,1190.755,95.276,997.564,-,\textbf{874.218},-
ETT2,H,7.618,0.071,7.720,0.100,7.452,0.127,\textcolor{blue}{7.334},0.013,7.811,0.055,7.677,0.094,7.636,0.568,7.932,0.019,\underline{7.279},0.057,8.022,0.278,7.660,0.080,\textbf{7.217},-,10.034,-
,D,\textbf{228.246},11.983,269.353,16.031,260.281,11.639,\textcolor{blue}{254.152},5.445,287.942,31.051,282.243,21.453,416.306,51.361,248.978,4.098,380.410,16.529,271.592,10.638,472.509,48.700,\underline{236.354},-,250.327,-
,W,2868.244,359.457,\textcolor{blue}{\underline{2211.349}},203.851,3016.452,686.651,\textcolor{blue}{2249.077},188.625,2870.885,414.512,2315.718,121.577,3337.477,502.711,2647.705,221.390,2645.104,379.367,2456.773,77.993,3048.120,159.123,2399.760,-,\textbf{1597.126},-
Solar,H,12.333,0.175,12.914,0.828,12.073,1.072,12.847,0.402,12.012,0.241,13.492,0.065,25.142,5.277,\textbf{10.872},0.236,\underline{11.078},0.062,11.986,0.284,13.434,0.333,11.365,-,27.067,-
,D,257.150,2.393,\textcolor{blue}{240.492},1.191,258.389,1.663,\textcolor{blue}{252.718},1.232,254.190,1.127,277.509,1.765,259.512,13.088,250.915,2.563,271.532,4.187,299.295,13.156,255.307,2.810,\textbf{231.535},-,\underline{237.917},-
,W,991.329,12.567,\textcolor{blue}{959.544},70.282,1127.889,20.369,\textcolor{blue}{1058.990},38.140,939.926,51.200,855.893,37.860,1014.611,30.887,815.778,35.372,\underline{812.396},83.750,1248.227,147.094,\textbf{788.671},22.443,1283.383,-,927.811,-
\end{filecontents*}
\csvreader[before reading=\gdef\prevdatasetone{},late after line=\\,late after last line=\vspace{3pt}\\\bottomrule]{tables/mae_main.csv}
{Dataset=\dataset,
Frequency=\freq,
moment_u=\momentU,
moment_u_se=\momentUSE,
moment_m=\momentM,
moment_m_se=\momentMSE,
patchtst_u=\patchtstU,
patchtst_u_se=\patchtstUSE,
patchtst_m=\patchtstM,
patchtst_m_se=\patchtstMSE,
itransformer_m=\itransformerM,
itransformer_m_se=\itransformerMSE,
itransformer_t5_m=\itransformerTfiveM,
itransformer_t5_m_se=\itransformerTfiveMSE,
crossformer_m=\crossformerM,
crossformer_m_se=\crossformerMSE,
chronos_m=\chronosM,
chronos_m_se=\chronosMSE,
timerxl_m=\timerxlM,
timerxl_m_se=\timerxlMSE,
tsmixer_m=\tsmixerM,
tsmixer_m_se=\tsmixerMSE,
timemixer_m=\timemixerM,
timemixer_m_se=\timemixerMSE,
mlp_m=\mlpM,
mlp_m_se=\mlpMSE,
autoets=\autoets,
autoets_se=\autoetsSE
}
{%
\IfStrEq{\dataset}{}{}{%
  \IfStrEq{\prevdatasetone}{}{}{\\\addlinespace[-3pt]\hline\addlinespace[6pt]}%
  \gdef\prevdatasetone{\dataset}%
}%
\multirow{2}{*}{\dataset} & \multirow{2}{*}{\freq} & \momentU & \momentM & \patchtstU & \patchtstM & \itransformerM & \itransformerTfiveM & \crossformerM & \timerxlM & \tsmixerM & \timemixerM & \mlpM & \chronosM & \autoets  \\
{} & {} & \SE{\momentUSE} & \SE{\momentMSE} & \SE{\patchtstUSE} & \SE{\patchtstMSE} & \SE{\itransformerMSE} & \SE{\itransformerTfiveMSE} & \SE{\crossformerMSE} & \SE{\timerxlMSE} & \SE{\tsmixerMSE} & \SE{\timemixerMSE} & \SE{\mlpMSE} & -- & -- }
\multicolumn{2}{l|}{\textbf{Average Rank}} & 5.500 & \underline{4.389} & 6.944 & \textbf{3.722} & 8.167 & 6.500 & 7.833 & 6.556 & 6.833 & 7.667 & 9.278 & 4.611 & -- \\
\bottomrule
\end{tabular}}
\label{tab:mae_main}
\end{table*}

\begin{figure*}[ht!]
    \centering
    \includegraphics[width=0.9\textwidth, trim=0 10 0 0, clip]{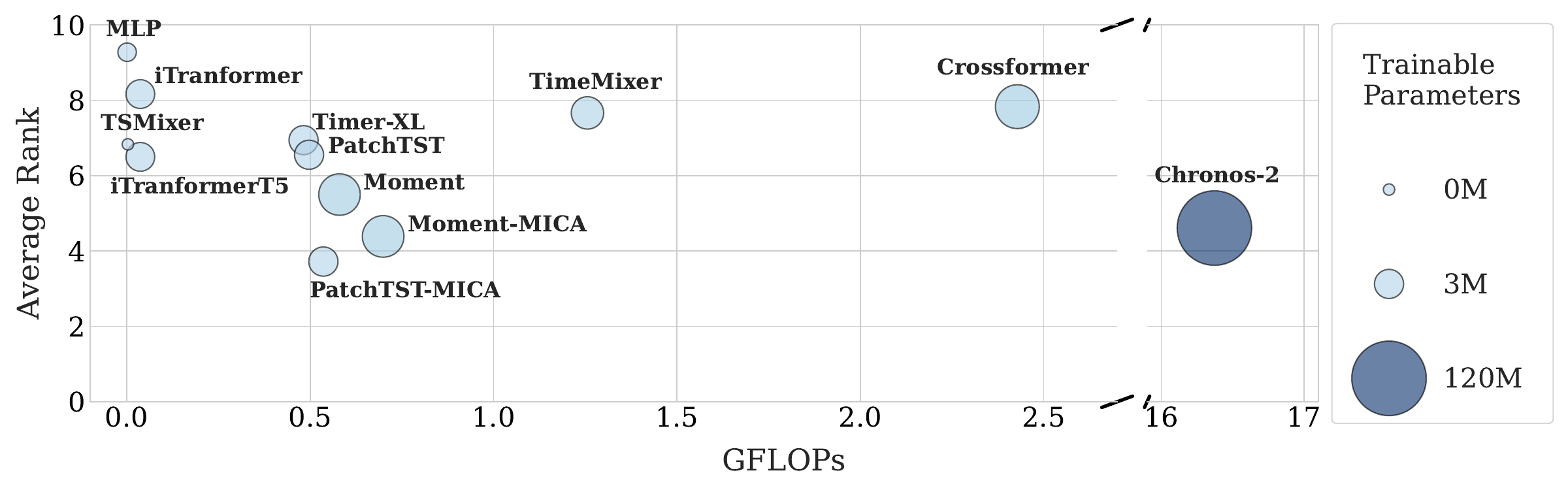}
    \caption{Average rank across datasets vs. computational cost (GFLOPs). \PatchTST-\MICA and \Moment-\MICA\ achieve best average rank across datasets in terms of MAE with relatively small computational overhead increase over their univariate counterparts.
    Marker size reflects trainable parameter count. GFLOPs are estimated at $C=7$, $H=48$, and $L=96$ as a representative low-dimensional setting.
}
    \label{fig:model_size_comparison_main}
\end{figure*}

\subsection{Channel-Independent Ablations}
We assess the relative value of \MICA’s channel-dependent backbone compared to alternative strategies for processing multivariate data with channel-independent backbones.

\paragraph{Decorrelation preprocessing.} To assess whether decorrelating channels can improve channel-independent architectures without modifying the model architecture itself, we apply \PCA\ as a decorrelation transform to the training data, retaining all principal components (i.e., no dimensionality reduction). We train \PatchTST\ and \Moment\ on the decorrelated channels and inverse-transform predictions to the original coordinate frame.

\paragraph{Multivariate Output Layer.}
We evaluate a \textit{multivariate output layer} as an alternative final output layer, keeping the channel-independent backbone unchanged. This layer learns separate linear projections $\mathbf{W}_c \in \mathbb{R}^{d \times H}$ for each channel $c \in \{1, \ldots, C\}$, where $d$ is the flattened representation dimension and $H$ is the forecast horizon. This contrasts with the standard \textit{shared output layer} that applies a single transformation across all channels.

\begin{figure*}[ht!]
    \centering
    \begin{tabular}{@{}c@{}}
        \includegraphics[width=0.9\textwidth, trim=0 10 0 0, clip]{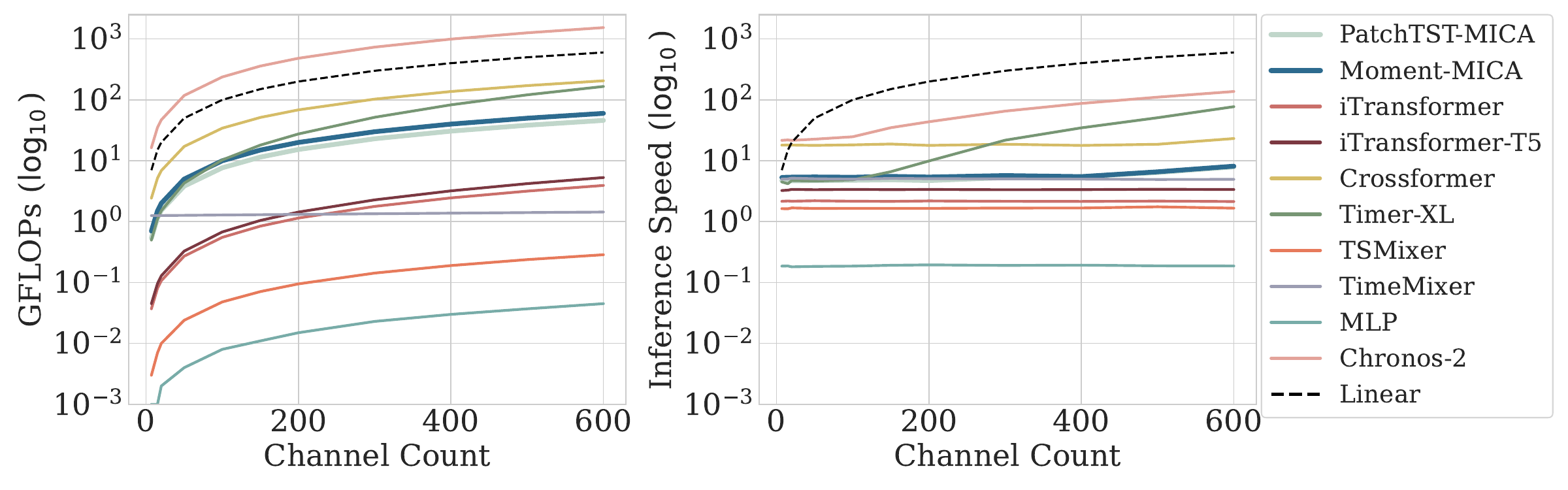} \\
        {\small (a) $\log_{10}$ GFLOPs and inference speed (ms) vs.\ channel count $C \in [7, 600]$ ($L=96, H=48$).}
        \label{fig:scaling_channel_count_comparison_log} \\[1em]
        \includegraphics[width=0.9\textwidth, trim=0 10 0 0, clip]{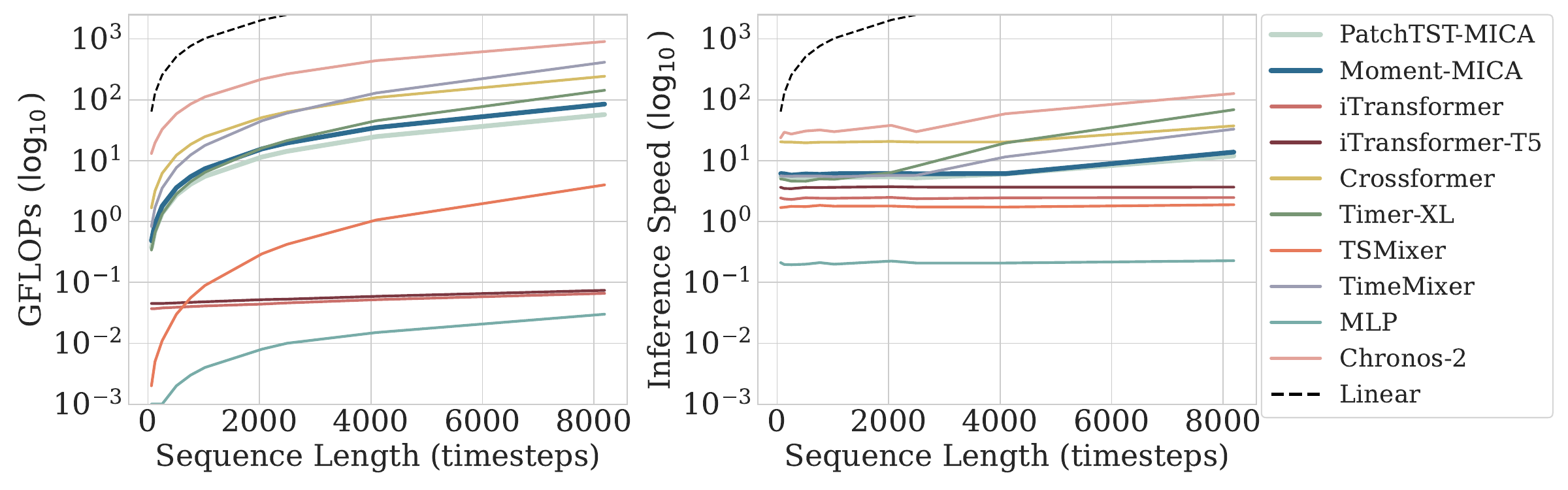} \\
        {\small (b) $\log_{10}$ GFLOPs and inference speed (ms) vs.\ context length $L \in [64, 8192]$ ($C=7$, $H=48$).}
        \label{fig:scaling_seqlen_comparison_log}
    \end{tabular}
    \caption{\MLP-based methods are the most computationally efficient overall. Among 
    Transformer-based models, \MICA\ variants scale best (except \iTransformer), 
    substantially outperforming \Crossformer, \TimerXL, and \Chronos\texttt{-2} in both 
    GFLOPs and inference speed; \MICA\ also outperforms \TimeMixer\ with context length. 
    Dashed line indicates linear scaling. Note the logarithmic y-axis; see 
    Figs.~\ref{fig:scaling_channels_comparison}--\ref{fig:scaling_seqlen_comparison} 
    for linear scale.}
    \label{fig:scaling_combined_log}
\end{figure*}

\subsection{\MICA\ Design Ablations}\label{section:mica_ablation_study}

\paragraph{Gating mechanisms.} We explore multiple strategies for combining local and global attention: (1)~\textit{shared $\beta$} uses a single scalar $\beta$ across all layers and channels; (2)~\textit{channel-wise $\beta$} learns separate $\beta$ per channel; (3)~\textit{layer-wise $\beta$} learns separate $\beta$ per layer; (4)~\textit{layer-wise channel-wise $\beta$} learns separate $\beta$ for each layer-channel combination; (5)~\textit{MLP gating} uses a multi-layer perceptron for non-linear mixing; and (6)~\textit{MLP query gating} additionally incorporates query information. We provide further details in Appendix~\ref{apd:infini_gate_variants}.

\paragraph{Channel exclusion strategy.} We evaluate two configurations for computing $\mathbf{A}_{\text{global}}$. \textit{Without channel exclusion}, all channels are retained in $\textbf{M}$ and $\textbf{z}$, and $\mathbf{A}_{\text{global}}$ mixes information across all channels. \textit{With channel exclusion}, when computing $\mathbf{A}_{\text{global}}$ for a given channel, that channel's information is excluded from $\textbf{M}$ and $\textbf{z}$ to avoid redundancy with information already captured through local attention in $\mathbf{A}_{\text{local}}$. We provide further details in Appendix~\ref{apd:infini_ciexcl_variants}.

\paragraph{Weighted Channel Aggregation.} We evaluate three configurations of weighted channel aggregation within the global attention component: \textit{uniform}, \textit{static}, and \textit{dynamic} channel weights. Uniform weights apply equal weights across channels, following the default linear attention formulation. Static weights are learnable parameters $w \in \mathbb{R}^C$ that allow channel-specific scaling without input dependence. Dynamic weights are derived from the query via a linear projection $W_{d_q}$, where $d_q$ is the per-head query hidden dimension, enabling input-adaptive aggregation. We provide further details in Appendix~\ref{apd:infini_weighted_channels_variants}.

\section{Results}
\label{section:results}
\begin{figure}[t!]
    \centering
    \includegraphics[width=0.495\textwidth, trim=0 15 0 0, clip]{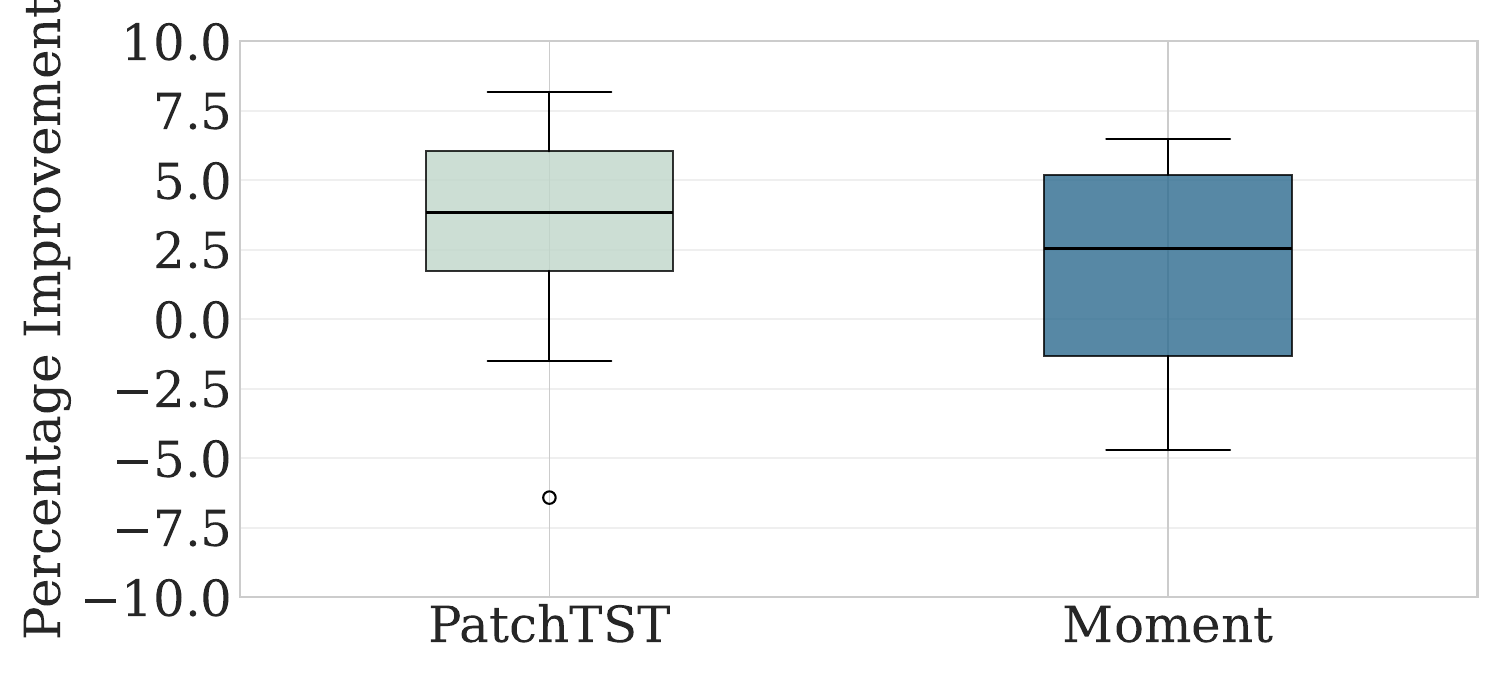}
    \caption{Distribution of percentage error reduction across datasets achieved by \PatchTST-\MICA\ and \Moment-\MICA\ (MLP w/ Query gate) relative to their univariate implementations. \MICA\ variants achieve lower median forecasting error across datasets.}
    \label{fig:percentage_improvment_main}
\end{figure}

\paragraph{\MICA\ models outperform their univariate counterparts.}
\PatchTST-\MICA\ and \Moment-\MICA\ reduce MAE by 5.4\% and 3.3\% on average 
(up to 25.4\% and 33.8\% on individual datasets) over their channel-independent 
counterparts (Fig.~\ref{fig:percentage_improvment_main}). Paired Wilcoxon signed-rank 
tests confirm \PatchTST-\MICA\ significantly outperforms its univariate counterpart 
($p{=}0.001$), while \Moment-\MICA\ improvements do not reach significance ($p{=}0.13$). Blue results in Table~\ref{tab:mae_main} indicate lower forecast error of MICA models compared with their channel-independent counterparts. The distribution of percentage error reduction across
datasets achieved by \MICA\ (MLP-query gate) is shown in Fig.~\ref{fig:percentage_improvment_main}. RMSE results are provided in Table~\ref{tab:rmse_main} in Appendix~\ref{section:apd_results}.

\paragraph{\MICA\ models rank first among multivariate baselines.}
\PatchTST-\MICA\ and \Moment-\MICA\ achieve average MAE ranks of 3.7 and 4.4, 
outranking all deep multivariate baselines. \texttt{Chronos-2} (zero-shot) ranks 
third (4.6), followed by \Moment, \iTransformer-\texttt{T5}, and \TimerXL\ (respective average ranks: 5.5, 
6.5, 6.6). Fig.~\ref{fig:model_size_comparison_main} shows that \MICA\ achieves 
these gains with relatively small computational overhead over channel-independent counterparts.

\paragraph{\MICA\ models outperform univariate model adaptations.}
\PatchTST-\MICA\ and \Moment-\MICA\ achieve average MAE ranks of 2.0 and 3.2 in Table~\ref{tab:mica_vs_pca}, outranking channel-independent models (\PatchTST: 5.2; \Moment: 3.8), multivariate output layer variants (\PatchTST: 4.8; \Moment: 4.3), and PCA preprocessing (ranked last). \MICA\ models also scale more efficiently in channel count than the multivariate output layer variants (Appendix~\ref{apd:channel_count_efficiency_analysis}).

\paragraph{\MICA\ models are compact and efficient relative to channel-independent counterparts.}
At $C{=}600$ (Table~\ref{tab:mica_variant_flops_n600}), \MICA\ (\MLP-query gate) requires 
only $1.11\times$/$1.21\times$ the GFLOPs and $1.02\times$/$1.01\times$ the parameters 
of channel-independent \PatchTST/\Moment. The multivariate output layer variant, requires 
$35.2\times$/$9.3\times$ more parameters and is $11.2\times$/$9.86\times$ slower than channel-independent models.

\paragraph{\MICA\ models are more efficient than channel-dependent Transformer baselines.}
At $C{=}600$ (Table~\ref{tab:baseline_flops_n600}; 
Appendix~\ref{apd:channel_count_efficiency_analysis}), \PatchTST-\MICA\ remains below 60 GFLOPs and 9ms, requiring $33.4\times$, $3.6\times$, and $4.5\times$ fewer GFLOPs and achieving $17.3\times$, $9.7\times$, and $2.9\times$ faster inference than \Chronos\texttt{-2}, \TimerXL, and \Crossformer, respectively. At $L{=}8192$ (Table~\ref{tab:baseline_flops_n7_is8192}; Appendix~\ref{apd:seqlen_efficiency_analysis}), 
\PatchTST-\MICA\ uses $15.9\times$, $2.5\times$, $4.3\times$, and $7.2\times$ fewer GFLOPs and achieves $10.6\times$, $5.7\times$, $3.1\times$, and $2.7\times$ faster inference than \Chronos\texttt{-2}, \TimerXL, \Crossformer, and \TimeMixer, respectively. Fig.~\ref{fig:scaling_combined_log} shows GFLOPs and inference speed as channel count and context length scales.

\paragraph{Global attention: simpler configurations suffice.}
Channel inclusion and exclusion variants perform comparably across datasets with no statistically significant 
differences. Uniform channel 
weighting achieves the best average rank (1.7), outperforming static (2.5) and dynamic (2.4) variants, both still improving over the channel-independent \PatchTST\ baseline (3.4), as shown in Table~\ref{tab:mae_weighted_channels_ablation}.

\paragraph{Gate type effectiveness varies across architectures.}
$\beta$-based gates achieve the best average ranks for \Moment\ (4.7--6.2) while \MLP\ gates excel 
for \PatchTST\ (3.2--4.5), though no statistically significant differences emerge 
(Figs.~\ref{fig:mica_variant_cd_moment}, \ref{fig:mica_variant_cd_patchtst}; Tables~\ref{tab:mae_moment_infini_ablation},~\ref{tab:mae_patchtst_infini_ablation}). We hypothesize that \PatchTST's residual connections favor \MLP\ gates, whereas \Moment\ benefits from simpler $\beta$ gates to avoid overfitting.

\section{Discussion}
\label{section:discussion}
We introduced \MICA, a scalable cross-channel attention mechanism that extends channel-independent Transformers to multivariate forecasting, achieving the lowest 
average forecast error rank across evaluated models while scaling more efficiently than Transformer baselines that attend across both temporal and channel dimensions.

Four key insights emerge from ablation studies: 1) structural integration of cross-channel dependencies within the attention mechanism outperforms adapter-based alternatives such as decorrelation preprocessing and multivariate projection heads; 2) channel inclusion can be prioritized over exclusion without sacrificing performance, reducing computation; 3) uniform channel weighting suffices, outperforming learned channel-specific strategies on average; and 4) gate effectiveness is architecture-dependent.

Future work includes large-scale cross-frequency pretraining to evaluate \MICA's 
zero-shot capabilities and scaling behavior. Configurations without channel-specific parameters ($\beta$ gates excluding channelwise, \MLP-based gates, uniform and 
dynamic weighting) already support training across varying channel counts. Additionally, future work could explore alternative gating mechanisms, such as cross-attention-based gating, where mixing weights are computed via explicit attention over local and global representations rather than a feedforward transformation. While the present work focuses on Transformer- and MLP-based architectures, broader evaluation against state-space and graph-based models remains an open direction.


\section*{Impact Statement}
\MICA\ has the potential to positively impact domains that rely on multivariate temporal data, such as healthcare monitoring, climate modeling, industrial sensing, and financial analysis, where accuracy and computational efficiency are critical. Improved efficiency may also lower barriers to deploying attention-based models in resource-constrained environments. As with any deep learning architecture, \MICA~is sensitive to data quality and to potential biases present in the training data. Responsible deployment should include appropriate validation.

\section*{Software and Data}
To ensure reproducibility, we have made our code accessible through \href{https://github.com/PotosnakW/neuralforecast/tree/MICA}{\texttt{GitHub}}. We implement our method and various baselines within the \texttt{neuralforecast} repository \citep{olivares2022library_neuralforecast}, leveraging its available models and standardized training infrastructure for consistent benchmarking. All baseline models and datasets used in our experiments are publicly available and open-source within our code repository. We evaluate on diverse multivariate time series benchmarks from the Gift-Eval repository \citep{aksu2024giftevalbenchmark} spanning climate, energy, traffic, and healthcare domains (detailed in Section~\ref{section:apd_data}), as well as two newly contributed high-frequency meteorological datasets procured from the Earth Observing Laboratory Field Data Archive. We compare against state-of-the-art methods using publicly available implementations \citep{olivares2022library_neuralforecast, garza2022statsforecast, zhang2023crossformer, liu2025timerxllongcontexttransformersunified, goswami2024moment}. All experiments were conducted on NVIDIA A100-SXM4-80GB GPUs (80GB memory).

\section*{Acknowledgments}
\textbf{Funding.} This work has been partially supported by the National Science Foundation (awards 2427948 and 2406231) and by GE Vernova Advanced Research Center. 

\textbf{Discussions.} We would like to express our sincerest gratitude to Masoud Abbaszadeh, Weizhong Yan, Qiuyi Chen and Varsha Pendyala, for insightful discussions regarding the problem setting, model validation and functionality enhancement suggestions. 

\bibliography{references}

\begin{thebibliography}{61}
\providecommand{\natexlab}[1]{#1}
\providecommand{\url}[1]{\texttt{#1}}
\expandafter\ifx\csname urlstyle\endcsname\relax
  \providecommand{\doi}[1]{doi: #1}\else
  \providecommand{\doi}{doi: \begingroup \urlstyle{rm}\Url}\fi

\bibitem[Aksu et~al.(2024)Aksu, Woo, Liu, Liu, Liu, Savarese, Xiong, and Sahoo]{aksu2024giftevalbenchmark}
Aksu, T., Woo, G., Liu, J., Liu, X., Liu, C., Savarese, S., Xiong, C., and Sahoo, D.
\newblock Gift-eval: A benchmark for general time series forecasting model evaluation, 2024.
\newblock URL \url{https://arxiv.org/abs/2410.10393}.

\bibitem[Alexandrov et~al.(2019)Alexandrov, Benidis, Bohlke-Schneider, Flunkert, Gasthaus, Januschowski, Maddix, Rangapuram, Salinas, Schulz, Stella, Türkmen, and Wang]{gluonts2019}
Alexandrov, A., Benidis, K., Bohlke-Schneider, M., Flunkert, V., Gasthaus, J., Januschowski, T., Maddix, D.~C., Rangapuram, S., Salinas, D., Schulz, J., Stella, L., Türkmen, A.~C., and Wang, Y.
\newblock Gluonts: Probabilistic time series models in python.
\newblock \url{https://github.com/awslabs/gluonts}, 2019.

\bibitem[Ansari et~al.(2025)Ansari, Shchur, Küken, Auer, Han, Mercado, and et~al.]{ansari2025chronos2}
Ansari, A.~F., Shchur, O., Küken, J., Auer, A., Han, B., Mercado, P., and et~al.
\newblock Chronos-2: From univariate to universal forecasting.
\newblock \emph{arXiv preprint}, 2025.
\newblock URL \url{https://arxiv.org/abs/2510.15821}.

\bibitem[Arthur~Harris \& Gano~Benedict(21919)Arthur~Harris and Gano~Benedict]{harris_benedict_basal_metabolism}
Arthur~Harris, J. and Gano~Benedict, F.
\newblock \emph{A biometric study of basal metabolism in man}.
\newblock Carnegie institution of Washington, 21919.

\bibitem[Cachay et~al.(2021)Cachay, Erickson, Bucker, Pokropek, Potosnak, Bire, Osei, and L{\"u}tjens]{cachay2021world}
Cachay, S.~R., Erickson, E., Bucker, A. F.~C., Pokropek, E., Potosnak, W., Bire, S., Osei, S., and L{\"u}tjens, B.
\newblock The world as a graph: Improving el ni{\~n}o forecasts with graph neural networks.
\newblock \emph{arXiv preprint arXiv:2104.05089}, 2021.

\bibitem[Cao et~al.(2020)Cao, Wang, Duan, Zhang, Zhu, Huang, and et. al.]{cao2020stemgnn}
Cao, D., Wang, Y., Duan, J., Zhang, C., Zhu, X., Huang, C., and et. al.
\newblock Spectral temporal graph neural network for multivariate time-series forecasting.
\newblock In \emph{34th Conference on Neural Information Processing Systems}, 2020.

\bibitem[Chen et~al.(2023)Chen, Li, Yoder, $\ddot{\text{O}}$. Arık, and Pfister]{chen2023tsmixer}
Chen, S.-A., Li, C.-L., Yoder, N.~C., $\ddot{\text{O}}$. Arık, S., and Pfister, T.
\newblock {TSMixer}: An all-{MLP} architecture for time series forecasting.
\newblock In \emph{Published in Transactions on Machine Learning Research}, 2023.

\bibitem[Choromanski et~al.(2021)Choromanski, Likhosherstov, Dohan, Song, Gane, Sarlos, and et~al.]{choromanski2021performers}
Choromanski, K., Likhosherstov, V., Dohan, D., Song, X., Gane, A., Sarlos, T., and et~al.
\newblock Rethinking attention with performers.
\newblock In \emph{International Conference on Learning Representations}, 2021.
\newblock URL \url{https://arxiv.org/abs/2009.14794}.

\bibitem[Cui et~al.(2018)Cui, Ke, and Wang]{cui2018loopseattle}
Cui, Z., Ke, R., and Wang, Y.
\newblock Deep bidirectional and unidirectional lstm recurrent neural network for network-wide traffic speed prediction.
\newblock \emph{arXiv preprint arXiv:1801.02143}, 2018.

\bibitem[Cui et~al.(2019)Cui, Henrickson, Ke, and Wang]{cui2019loopseattle}
Cui, Z., Henrickson, K., Ke, R., and Wang, Y.
\newblock Traffic graph convolutional recurrent neural network: A deep learning framework for network-scale traffic learning and forecasting.
\newblock \emph{IEEE Transactions on Intelligent Transportation Systems}, 2019.

\bibitem[Dao et~al.(2022)Dao, Fu, Ermon, Rudra, and R{\'e}]{dao2022flashattention}
Dao, T., Fu, D.~Y., Ermon, S., Rudra, A., and R{\'e}, C.
\newblock Flash{A}ttention: Fast and memory-efficient exact attention with {IO}-awareness.
\newblock In \emph{Advances in Neural Information Processing Systems}, 2022.

\bibitem[Das et~al.(2024)Das, Kong, Sen, and Zhou]{das2024timesfm}
Das, A., Kong, W., Sen, R., and Zhou, Y.
\newblock A decoder-only foundation model for time-series forecasting.
\newblock In \emph{Proceedings of the Forty-First International Conference on Machine Learning (ICML)}, 2024.
\newblock URL \url{https://arxiv.org/abs/2310.10688}.

\bibitem[de~Medrano \& Aznarte(2020)de~Medrano and Aznarte]{medrano2020mdense}
de~Medrano, R. and Aznarte, J.~L.
\newblock A spatio-temporal spot-forecasting framework for urban traffic prediction.
\newblock \emph{Applied Soft Computing}, 2020.

\bibitem[Ekambaram et~al.(2024)Ekambaram, Jati, Nguyen, Dayama, Reddy, Gifford, and Kalagnanam]{ekambaram2024ttms}
Ekambaram, V., Jati, A., Nguyen, N.~H., Dayama, P., Reddy, C., Gifford, W.~M., and Kalagnanam, J.
\newblock {TTMs}: Fast multi-level tiny time mixers for improved zero-shot and few-shot forecasting of multivariate time series.
\newblock \emph{arXiv preprint arXiv:2401.03955}, 2024.

\bibitem[Garza et~al.(2022)Garza, Canseco, Challú, and Olivares]{garza2022statsforecast}
Garza, F., Canseco, M.~M., Challú, C., and Olivares, K.~G.
\newblock {StatsForecast}: Lightning fast forecasting with statistical and econometric models.
\newblock {PyCon} Salt Lake City, Utah, US 2022, 2022.
\newblock URL \url{https://github.com/Nixtla/statsforecast}.

\bibitem[Godahewa et~al.(2021)Godahewa, Bergmeir, Webb, Hyndman, and Montero-Manso]{godahewa2021monash}
Godahewa, R.~W., Bergmeir, C., Webb, G.~I., Hyndman, R., and Montero-Manso, P.
\newblock Monash time series forecasting archive.
\newblock In \emph{Thirty-fifth Conference on Neural Information Processing Systems Datasets and Benchmarks Track (Round 2)}, 2021.
\newblock URL \url{https://openreview.net/forum?id=wEc1mgAjU-}.

\bibitem[Gon{\c{c}}alves et~al.(2021)Gon{\c{c}}alves, Cortez, Carvalho, and Fraz{\~a}o]{goncalves2021multivariate}
Gon{\c{c}}alves, J. N.~C., Cortez, P., Carvalho, M.~S., and Fraz{\~a}o, N.~M.
\newblock A multivariate approach for multi-step demand forecasting in assembly industries: Empirical evidence from an automotive supply chain.
\newblock \emph{Decision Support Systems}, 142, 2021.

\bibitem[Goswami et~al.(2024)Goswami, Szafer, Choudhry, Cai, Li, and Dubrawski]{goswami2024moment}
Goswami, M., Szafer, K., Choudhry, A., Cai, Y., Li, S., and Dubrawski, A.
\newblock {MOMENT}: A family of open time-series foundation models.
\newblock In \emph{International Conference on Machine Learning}, 2024.

\bibitem[Han et~al.(2024)Han, Chen, Ye, and Zhan]{han2024softs}
Han, L., Chen, X.-Y., Ye, H.-J., and Zhan, D.-C.
\newblock Softs: Efficient multivariate time series forecasting with series-core fusion.
\newblock \emph{arXiv preprint arXiv:2404.14197}, 2024.

\bibitem[Hu et~al.(2025)Hu, Zhang, Liu, Lan, Li, Cheng, Dai, Xia, and Pan]{hu2025timefilter}
Hu, Y., Zhang, G., Liu, P., Lan, D., Li, N., Cheng, D., Dai, T., Xia, S.-T., and Pan, S.
\newblock Timefilter: Patch-specific spatial-temporal graph filtration for time series forecasting.
\newblock In \emph{Forty-second International Conference on Machine Learning}, 2025.
\newblock URL \url{https://openreview.net/forum?id=490VcNtjh7}.

\bibitem[Hyndman(2008)]{ets_2008}
Hyndman, R.
\newblock \emph{Forecasting with exponential smoothing: the state space approach}.
\newblock Springer Berlin, Heidelberg, 2008.

\bibitem[{Iowa State University}(2005)]{ISU_SMEX02_AWOS_2005}
{Iowa State University}.
\newblock {SMEX02: Automated Weather Observing System (AWOS) Iowa 1-min Data}, 2005.
\newblock URL \url{https://doi.org/10.26023/SM2V-E9KY-EG10}.
\newblock Accessed: 26 December 2025.

\bibitem[{Iowa State University}(2008)]{ISU_IHOP2002_AWOS_2008}
{Iowa State University}.
\newblock {IHOP\_2002: Automated Weather Observing System (AWOS) Iowa 1-min Data}, 2008.
\newblock URL \url{https://doi.org/10.26023/ZM18-BQWR-0B11}.
\newblock Accessed: 26 December 2025.

\bibitem[{Iowa State University}(2010)]{ISU_PLOWS_AWOS_2010}
{Iowa State University}.
\newblock {PLOWS: Iowa Automated Weather Observing System (AWOS) 1-minute Data}, 2010.
\newblock URL \url{https://doi.org/10.26023/Y9RJ-GH3F-FM01}.
\newblock Accessed: 26 December 2025.

\bibitem[Katharopoulos et~al.(2020)Katharopoulos, Vyas, Pappas, and Fleuret]{pmlr-v119-katharopoulos20a}
Katharopoulos, A., Vyas, A., Pappas, N., and Fleuret, F.
\newblock Transformers are {RNNs}: Fast autoregressive transformers with linear attention.
\newblock In III, H.~D. and Singh, A. (eds.), \emph{Proceedings of the 37th International Conference on Machine Learning}, volume 119 of \emph{Proceedings of Machine Learning Research}, pp.\  5156--5165. PMLR, 13--18 Jul 2020.
\newblock URL \url{https://proceedings.mlr.press/v119/katharopoulos20a.html}.

\bibitem[Kessy et~al.(2018)Kessy, Lewin, and Strimmer]{kessy2018optimal}
Kessy, A., Lewin, A., and Strimmer, K.
\newblock Optimal whitening and decorrelation.
\newblock \emph{The American Statistician}, 72\penalty0 (4):\penalty0 309--314, 2018.
\newblock \doi{10.1080/00031305.2016.1277159}.

\bibitem[Lai et~al.(2018)Lai, Chang, Yang, and Liu]{lai2018solar}
Lai, G., Chang, W.-C., Yang, Y., and Liu, H.
\newblock Modeling long- and short-term temporal patterns with deep neural networks.
\newblock In \emph{The 41st International ACM SIGIR Conference on Research \& Development in Information Retrieval}, 2018.
\newblock URL \url{https://api.semanticscholar.org/CorpusID:4922476}.

\bibitem[Liu et~al.(2024)Liu, Hu, Zhang, Wu, Wang, Ma, and Long]{liu2024itransformer}
Liu, Y., Hu, T., Zhang, H., Wu, H., Wang, S., Ma, L., and Long, M.
\newblock {iTransformer}: Inverted transformers are effective for time series forecasting.
\newblock In \emph{The Twelfth International Conference on Learning Representations}, 2024.
\newblock URL \url{https://openreview.net/forum?id=JePfAI8fah}.

\bibitem[Liu et~al.(2025)Liu, Qin, Huang, Wang, and Long]{liu2025timerxllongcontexttransformersunified}
Liu, Y., Qin, G., Huang, X., Wang, J., and Long, M.
\newblock Timer-xl: Long-context transformers for unified time series forecasting.
\newblock In \emph{Proceedings of the Thirteenth International Conference on Learning Representations}, 2025.

\bibitem[Luo \& Wang(2024)Luo and Wang]{luo2024moderntcn}
Luo, D. and Wang, X.
\newblock Moderntcn: A modern pure convolution structure for general time series analysis.
\newblock In \emph{International Conference on Learning Representations (ICLR)}, 2024.

\bibitem[Ma et~al.(2025)Ma, Ni, Xiao, and Chen]{timepro2025}
Ma, X., Ni, Z.-L., Xiao, S., and Chen, X.
\newblock Timepro: Efficient multivariate long-term time series forecasting with variable- and time-aware hyper-state.
\newblock In \emph{Forty-second International Conference on Machine Learning}, 2025.
\newblock URL \url{https://openreview.net/forum?id=s69Ei2VrIW}.

\bibitem[Munkhdalai et~al.(2024)Munkhdalai, Faruqui, and Gopal]{munkhdalai2024leave}
Munkhdalai, T., Faruqui, M., and Gopal, S.
\newblock Leave no context behind: Efficient infinite context transformers with infini-attention.
\newblock \emph{arXiv preprint arXiv:2404.07143}, 2024.

\bibitem[Nguyen et~al.(2023)Nguyen, Nguyen, and Das]{nguyen2023correlatedattentiontransformers}
Nguyen, Q.~M., Nguyen, L.~M., and Das, S.
\newblock Correlated attention in transformers for multivariate time series, 2023.
\newblock URL \url{https://arxiv.org/abs/2311.11959}.

\bibitem[Nie et~al.(2023)Nie, Nguyen, Sinthong, and Kalagnanam]{nie2023patchtst}
Nie, Y., Nguyen, N.~H., Sinthong, P., and Kalagnanam, J.
\newblock A time series is worth 64 words: Long-term forecasting with transformers.
\newblock \emph{arXiv preprint arXiv:2307.13787}, 2023.
\newblock URL \url{https://arxiv.org/abs/2307.13787}.

\bibitem[Olivares et~al.(2022{\natexlab{a}})Olivares, Challu, Marcjasz, Weron, and Dubrawski]{OlivaresChallu2022nbeatsx}
Olivares, K.~G., Challu, C., Marcjasz, G., Weron, R., and Dubrawski, A.
\newblock Neural basis expansion analysis with exogenous variables: Forecasting electricity prices with nbeatsx.
\newblock \emph{International Journal of Forecasting}, 39\penalty0 (2):\penalty0 884--900, 2022{\natexlab{a}}.

\bibitem[Olivares et~al.(2022{\natexlab{b}})Olivares, Challú, Garza, Canseco, and Dubrawski]{olivares2022library_neuralforecast}
Olivares, K.~G., Challú, C., Garza, F., Canseco, M.~M., and Dubrawski, A.
\newblock {NeuralForecast}: User friendly state-of-the-art neural forecasting models.
\newblock {PyCon} Salt Lake City, Utah, US 2022, 2022{\natexlab{b}}.
\newblock URL \url{https://github.com/Nixtla/neuralforecast}.

\bibitem[Peng et~al.(2021)Peng, Pappas, Yogatama, Schwartz, Smith, and Kong]{peng2021random}
Peng, H., Pappas, N., Yogatama, D., Schwartz, R., Smith, N.~A., and Kong, L.
\newblock Random feature attention.
\newblock In \emph{International Conference on Learning Representations}, 2021.

\bibitem[Popescu(2003)]{ica2003multivarfcst}
Popescu, T.
\newblock Multivariate time series forecasting using independent component analysis.
\newblock In \emph{EFTA 2003. 2003 IEEE Conference on Emerging Technologies and Factory Automation. Proceedings (Cat. No.03TH8696)}, volume~2, pp.\  782--789, 2003.
\newblock \doi{10.1109/ETFA.2003.1248778}.

\bibitem[Potosnak et~al.(2025{\natexlab{a}})Potosnak, Challu, Olivares, Dufendach, and Dubrawski]{potosnak2025pkforecast}
Potosnak, W., Challu, C.~I., Olivares, K.~G., Dufendach, K.~A., and Dubrawski, A.
\newblock Global deep forecasting with patient-specific pharmacokinetics.
\newblock In \emph{Proceedings of Machine Learning Research}, 2025{\natexlab{a}}.
\newblock URL \url{https://raw.githubusercontent.com/mlresearch/v287/main/assets/potosnak25a/potosnak25a.pdf}.

\bibitem[Potosnak et~al.(2025{\natexlab{b}})Potosnak, Wolff, Cao, Ma, Konstantinova, Efimov, Mahoney, Oreshkin, and Olivares]{potosnak2025forkingsequences}
Potosnak, W., Wolff, M., Cao, M., Ma, R., Konstantinova, T., Efimov, D., Mahoney, M.~W., Oreshkin, B., and Olivares, K.~G.
\newblock Forking-sequences, 2025{\natexlab{b}}.
\newblock URL \url{https://arxiv.org/abs/2510.04487}.

\bibitem[Qin et~al.(2022)Qin, Sun, Deng, Li, Wei, Lv, and et~al.]{qin2022cosformer}
Qin, Z., Sun, W., Deng, H., Li, D., Wei, Y., Lv, B., and et~al.
\newblock cos{F}ormer: Rethinking softmax in attention.
\newblock In \emph{International Conference on Learning Representations}, 2022.
\newblock URL \url{https://arxiv.org/abs/2202.08791}.

\bibitem[Rahimi \& Recht(2007)Rahimi and Recht]{rahimi2007random}
Rahimi, A. and Recht, B.
\newblock Random features for large-scale kernel machines.
\newblock In \emph{Advances in Neural Information Processing Systems}, volume~20, 2007.

\bibitem[Rosenblatt(1958)]{rosenblatt1958_mlp}
Rosenblatt, F.
\newblock The {P}erceptron: A probabilistic model for information storage and organization in the brain.
\newblock \emph{Psychological Review}, 65\penalty0 (6):\penalty0 386–--408, 1958.

\bibitem[Santos et~al.(2013)Santos, Nogales, and Ruiz]{santos2013var}
Santos, A. A.~P., Nogales, F.~J., and Ruiz, E.
\newblock Comparing univariate and multivariate models to forecast portfolio value-at-risk.
\newblock \emph{Journal of Financial Econometrics}, 11\penalty0 (2):\penalty0 400--441, 2013.
\newblock \doi{10.1093/jjfinec/nbs015}.

\bibitem[Sun et~al.(2023)Sun, Dong, Huang, Ma, Xia, Xue, Wang, and Wei]{sun2023retnet}
Sun, Y., Dong, L., Huang, S., Ma, S., Xia, Y., Xue, J., Wang, J., and Wei, F.
\newblock Retentive network: A successor to {Transformer} for large language models, 2023.

\bibitem[Vaswani et~al.(2017)Vaswani, Shazeer, Parmar, Uszkoreit, Jones, and Gomez]{vaswani_2021_attentionisallyouneed}
Vaswani, A., Shazeer, N., Parmar, N., Uszkoreit, J., Jones, L., and Gomez, A.~N.
\newblock Attention is all you need.
\newblock In \emph{Proceedings of the 31st Conference on Neural Information Processing Systems}, 2017.

\bibitem[Visentin et~al.(2016)Visentin, Dalla~Man, and Cobelli]{visentin2016towards_single_day_simulator}
Visentin, R., Dalla~Man, C., and Cobelli, C.
\newblock One-day bayesian cloning of type 1 diabetes subjects: Toward a single-day uva/padova type 1 diabetes simulator.
\newblock \emph{IEEE Transactions on Biomedical Engineering}, 63\penalty0 (11), 2016.

\bibitem[Wang et~al.(2024{\natexlab{a}})Wang, Li, Shi, Ye, Mo, Lin, Ju, Chu, and Jin]{wang2024timemixer++}
Wang, S., Li, J., Shi, X., Ye, Z., Mo, B., Lin, W., Ju, S., Chu, Z., and Jin, M.
\newblock Timemixer++: A general time series pattern machine for universal predictive analysis.
\newblock \emph{arXiv preprint arXiv:2410.16032}, 2024{\natexlab{a}}.

\bibitem[Wang et~al.(2024{\natexlab{b}})Wang, Wu, Shi, Hu, Luo, Ma, Zhang, and ZHOU]{wang2023timemixer}
Wang, S., Wu, H., Shi, X., Hu, T., Luo, H., Ma, L., Zhang, J.~Y., and ZHOU, J.
\newblock Timemixer: Decomposable multiscale mixing for time series forecasting.
\newblock In \emph{International Conference on Learning Representations (ICLR)}, 2024{\natexlab{b}}.

\bibitem[Wolf et~al.(2020)Wolf, Debut, Sanh, et~al.]{wolf2020transformers}
Wolf, T., Debut, L., Sanh, V., et~al.
\newblock Transformers: State-of-the-art natural language processing.
\newblock \url{https://github.com/huggingface/transformers}, 2020.

\bibitem[Woo et~al.(2024)Woo, Liu, Kumar, Xiong, Savarese, and Sahoo]{woo2024moirai}
Woo, G., Liu, C., Kumar, A., Xiong, C., Savarese, S., and Sahoo, D.
\newblock Unified training of universal time series forecasting transformers.
\newblock In \emph{Proceedings of the 41st International Conference on Machine Learning}, Vienna, Austria, 2024. International Conference on Machine Learning.

\bibitem[Wu et~al.(2021)Wu, Xu, Wang, and Long]{wu2021autoformer}
Wu, H., Xu, J., Wang, J., and Long, M.
\newblock Autoformer: Decomposition transformers with {Auto-Correlation} for long-term series forecasting.
\newblock In \emph{Advances in Neural Information Processing Systems}, 2021.

\bibitem[Wu et~al.(2023)Wu, Hu, Liu, Zhou, Wang, and Long]{wu2023timesnettemporal2dvariationmodeling}
Wu, H., Hu, T., Liu, Y., Zhou, H., Wang, J., and Long, M.
\newblock {TimesNet}: Temporal 2d-variation modeling for general time series analysis.
\newblock In \emph{Proceedings of the 34th International Conference on Learning Representations}, 2023.

\bibitem[Wu et~al.(2024)Wu, Hu, Liu, Zhou, Wang, and Long]{tslib2024}
Wu, H., Hu, T., Liu, Y., Zhou, H., Wang, J., and Long, M.
\newblock Time series library.
\newblock \url{https://github.com/thuml/Time-Series-Library}, 2024.

\bibitem[Xie(2018)]{simglucose_dataset}
Xie, J.
\newblock Simglucose v0.2.1 (2018), 2018.
\newblock URL \url{https://github.com/jxx123/simglucose}.

\bibitem[Zhang \& Yan(2023)Zhang and Yan]{zhang2023crossformer}
Zhang, Y. and Yan, J.
\newblock Crossformer: Transformer utilizing cross-dimension dependency for multivariate time series forecasting.
\newblock In \emph{The Eleventh International Conference on Learning Representations}, 2023.

\bibitem[Zhang et~al.(2024)Zhang, Liu, Zhou, and Yan]{zhangup2me}
Zhang, Y., Liu, M., Zhou, S., and Yan, J.
\newblock {UP2ME}: Univariate pre-training to multivariate fine-tuning as a general-purpose framework for multivariate time series analysis.
\newblock In \emph{Forty-first International Conference on Machine Learning}, 2024.

\bibitem[Zheng \& Sun(2024)Zheng and Sun]{zheng2024mvgcrps}
Zheng, V.~Z. and Sun, L.
\newblock {MVG-CRPS}: A robust loss function for multivariate probabilistic forecasting.
\newblock \emph{arXiv preprint arXiv:2410.09133}, October 2024.
\newblock URL \url{https://arxiv.org/abs/2410.09133}.

\bibitem[Zhou et~al.(2021)Zhou, Zhang, Peng, Zhang, Li, Xiong, Zhang, Lin, Chu, Zhang, et~al.]{zhou2021informer}
Zhou, H., Zhang, S., Peng, J., Zhang, S., Li, G., Xiong, H., Zhang, W., Lin, T.-J., Chu, X., Zhang, J., et~al.
\newblock Informer: Beyond efficient transformer for long sequence time-series forecasting.
\newblock In \emph{Proceedings of the AAAI Conference on Artificial Intelligence}, 2021.

\bibitem[Zhou et~al.(2022)Zhou, Ma, Wen, Wang, Sun, and Jin]{zhou2022fedformer}
Zhou, T., Ma, Z., Wen, Q., Wang, X., Sun, L., and Jin, R.
\newblock {FEDformer}: Frequency enhanced decomposed transformer for long-term series forecasting.
\newblock In \emph{Proceedings of the 39th International Conference on Machine Learning}, volume 162 of \emph{Proceedings of Machine Learning Research}, pp.\  27268--27286. PMLR, 2022.

\bibitem[{\.Z}ukowska et~al.(2024){\.Z}ukowska, Goswami, Wili{\'n}ski, Potosnak, and Dubrawski]{infinichannelmixer}
{\.Z}ukowska, N., Goswami, M., Wili{\'n}ski, M., Potosnak, W., and Dubrawski, A.
\newblock Towards long-context time series foundation models.
\newblock \emph{arXiv preprint arXiv:2409.13530}, 2024.

\end{thebibliography}
\bibliographystyle{icml2025}

\newpage
\appendix
\onecolumn

\pagebreak
\section{Extended Related Work}
\label{section:apd_related_work}
\subsection{MICA Comparison with Multivariate Transformer Architectures}\label{section:apd_related_work_comparison}

\MICA\ differs from existing multivariate Transformer architectures in both mechanism and complexity.

\Moirai~\citep{woo2024moirai} concatenates channel tokens and applies attention over the expanded token set without modifying the attention mechanism, incurring $\mathcal{O}(P^2C^2)$ complexity that scales quadratically with channel count. \MICA\ differs in that it directly modifies the attention backbone by augmenting scaled dot-product attention with a linear attention mechanism and mixing gate to model cross-channel dependencies. 

\TimerXL~\citep{liu2025timerxllongcontexttransformersunified} flattens the 2D multivariate structure into a 1D token sequence and applies attention over this expanded set, incurring $\mathcal{O}(P^2C^2)$ complexity. Unlike \Moirai's standard attention, \TimerXL\ employs a Kronecker product-based mask to preserve causal dependencies. In contrast, \MICA\ avoids token expansion by compressing cross-channel information through linear attention and fusing it with temporal attention via a learnable gate, achieving $\mathcal{O}(P^2C + PC)$ complexity.

While \iTransformer~\citep{liu2024itransformer} inverts the data representation to treat channels as tokens ($\mathcal{O}(PC^2)$), this approach eliminates temporal attention entirely. \MICA\ preserves the standard temporal token structure while selectively integrating cross-channel information via gating.

\Crossformer~\citep{zhang2023crossformer} employs a two-stage attention design with dedicated cross-time and cross-dimension layers. \MICA\ differs from \Crossformer\ in several key aspects: 1) \textbf{Execution:} \Crossformer's stages must be executed \emph{sequentially}, where the output of the cross-time stage becomes the input to the cross-dimension stage, whereas \MICA's local temporal attention and global cross-channel attention can operate \emph{in parallel}. 2) \textbf{Cross-channel compression:} \Crossformer\ introduces a router mechanism with a small fixed number of learnable vectors ($R \ll C$) that act as an information bottleneck. These router vectors compress cross-channel information by aggregating messages from all $C$ channels, which are then queried by individual channels to retrieve the compressed representation. In contrast, \MICA\ compresses cross-channel information directly via summation across the channel dimension to construct the memory matrix $\mathbf{M}$, which is subsequently queried. 3) \textbf{Complexity scaling:} \Crossformer\ reduces complexity from $\mathcal{O}(PC^2)$ to $\mathcal{O}(PRC)$ only when $R \ll C$, requiring a constraint on the number of router vectors. \MICA\ achieves $\mathcal{O}(P^2C + PC)$ complexity while processing all $C$ channels directly without requiring $R \ll C$ for efficiency. 4) \textbf{Information fusion:} \Crossformer\ combines temporal and cross-dimension attention outputs through addition, whereas \MICA\ employs a learnable gate to compute an adaptive weighted combination of temporal and cross-dimension attention outputs.

\ChronosTwo~\citep{ansari2025chronos2} alternates between time and group attention layers that operate sequentially within each transformer block, achieving $\mathcal{O}(P^2C + PC^2)$ complexity. The time layer aggregates across patches within individual series, while the group layer aggregates across all series at each patch index. In contrast, \MICA\ computes local temporal and global cross-channel attention as independent operations that can be parallelized. \MICA's cross-channel mechanism architecturally differs from group attention: Rather than group attention's full attention over all $C$ channels, \MICA\ employs linear attention to compress cross-channel information via summation, achieving $\mathcal{O}(P^2C + PC)$ complexity. \MICA\ uses a learnable gate to adaptively fuse local and global representations rather than fixed alternation.

\MICA's approach shares conceptual similarities with \SOFTS~\citep{han2024softs}, which also employs gating to fuse local (channel-specific) and global (cross-channel) representations. However, \SOFTS\ operates within an MLP-based architecture and applies channel mixing at the encoder level, whereas \MICA\ integrates cross-channel interaction directly within the transformer attention mechanism. Critically, \SOFTS\ uses an MLP gate to combine local and global representations derived from separate processing pathways, while \MICA's channel-aware gates compress and modulate cross-channel information in a single pass without repeated concatenation operations.

\subsection{Extended Multivariate Architectures}\label{section:apd_other_multivariate_architectures}

\paragraph{Convolution-based models.}
\ModernTCN~\citep{luo2024moderntcn} captures cross-channel dependencies through depthwise-separable convolutions, which decouple spatial and channel-wise filtering to achieve linear complexity with channel count and sequences length $\mathcal{O}(PC)$. \TimesNet~\citep{wu2023timesnettemporal2dvariationmodeling} takes a different approach by reshaping 1D time series into 2D tensors and applying 2D convolutions to model both intra- and inter-period variation with complexity of $\mathcal{O}(P \log(P) C)$.

\paragraph{State-Space models.} We note that \TimePro~\citep{timepro2025} reports a complexity of $\mathcal{O}(PC)$ for the core HyperMamba component; however, the full model includes two linear projections with complexity $\mathcal{O}(P^2C)$, which is quadratic in the number of patches $P$. As $P$ scales with sequence length, the overall complexity is more precisely $\mathcal{O}(P^2C)$ when considering all components.

\paragraph{Graph-based models.} \StemGNN~\citep{cao2020stemgnn} models cross-channel dependencies by applying Graph Fourier Transform (GFT) jointly with Discrete Fourier Transform (DFT) to capture inter-series correlations and temporal dependencies together in the spectral domain; however, the eigendecomposition required by GFT incurs cubic cost in channel count, yielding an overall complexity of $\mathcal{O}(C^3 + PC^2 + P log(P)C)$. \TimeFilter~\citep{hu2025timefilter} constructs a patch-specific spatial-temporal graph that models temporal, inter-channel, and cross-channel dependencies while filtering out irrelevant edges per patch achieving a complexity of $\mathcal{O}(P^2 C + PC^2)$, which is quadratic in both $P$ and $C$. 

\subsection{Efficient Attention Architectures}\label{section:apd_efficient_attention_architectures}

Building on the development of linear attention~\citep{pmlr-v119-katharopoulos20a}, efficient attention mechanisms have been proposed to address the quadratic complexity of standard attention with respect to sequence length in sequence modeling. 

\Performers~\citep{choromanski2021performers} construct an unbiased approximation of the softmax kernel using exponential random feature maps with orthogonal random vector sampling, ensuring non-negative attention weights and achieving $\mathcal{O}(LR)$ complexity, where $R$ is the number of random features. Linear scaling over sequence length $L$ holds only when $R \ll L$, and while increasing $R$ reduces approximation variance, it reintroduces quadratic-like costs as $R\rightarrow L$. 

Random Feature Attention (\texttt{RFA})~\citep{peng2021random} introduces a variation using sinusoidal random Fourier features~\citep{rahimi2007random}  with i.i.d. random vector sampling to construct an unbiased approximation of the softmax kernel, retaining the same $\mathcal{O}(LR)$ complexity and $R \ll L$ requirement. In place of \Performers' orthogonal sampling, \texttt{RFA} instead reduces approximation variance via normalization of queries and keys.

\Performers~\citep{choromanski2021performers} address this by instead using positive orthogonal random features specifically designed to yield an unbiased and lower-variance approximation of the softmax kernel, achieving $\mathcal{O}(LR)$ complexity where $L$ is the context length and $R$ is the number of random features. While this scales linearly in $L$ when $R$ is fixed, $R$ introduces an additional hyperparameter and linear scaling only holds when $R \ll L$. 

\cosFormer~\citep{qin2022cosformer} instead replaces  softmax with a \texttt{ReLU} activation to ensure non-negativity, combined with a cosine-based re-weighting mechanism to recover the locality bias of softmax, achieving true 
$\mathcal{O}(L)$ complexity without additional hyperparameters. However, adapting 
\cosFormer\ for cross-channel attention in multivariate forecasting is not well-motivated: the cosine re-weighting encodes a notion of token proximity that is meaningful along the time dimension but meaningless across channels, which have no inherent ordering. 

\texttt{RetNet}~\citep{sun2023retnet} processes tokens sequentially via a recurrent state update with an exponential decay factor that causes earlier tokens to contribute less to later ones, however this efficiency of $\mathcal{O}(1)$  constant complexity with respect to sequence length $L$ is exclusive to inference as training retains the same $\mathcal{O}(L^2)$ quadratic complexity as standard attention. Similar to \cosFormer, adapting this mechanism for cross-channel aggregation in multivariate forecasting is not well-motivated: channels have no inherent ordering, yet would be processed arbitrarily and systematically downweighted by the decay factor, imposing spurious structure on channels that have no inherent ordering. 

In contrast, Infini-Attention~\citep{munkhdalai2024leave} aggregates tokens via summation, which is order-invariant and achieves true $\mathcal{O}(L)$ complexity without approximation error or additional hyperparameters beyond the gate parameter $\beta$. These properties make Infini-Attention a natural candidate for cross-channel aggregation in multivariate time series, motivating our adaptation of this architectural design for cross-channel mixing and compression.

Additionally, dedicated channel-independent (CI) forecasting models have been developed to address the quadratic complexity of attention mechanisms, including 
\Informer~\citep{zhou2021informer}, which achieves $\mathcal{O}(L \log L)$ complexity 
with input length $L$; \Autoformer~\citep{wu2021autoformer}, which achieves $\mathcal{O}(L \log L)$ complexity; \Fedformer~\citep{zhou2022fedformer}, which achieves $\mathcal{O}(L)$ complexity; and \PatchTST~\citep{nie2023patchtst}, which achieves $\mathcal{O}(P^2)$ complexity by segmenting the input into fixed-length patches of length $P$. While the \MICA\ design is compatible with any channel-independent Transformer architectures, we select \PatchTST\ as our base model given it outperforms \Informer, \Autoformer, and \Fedformer\ on standard benchmarks~\citep{nie2023patchtst} and patch-based tokenization has become the dominant paradigm in modern Time Series Foundation Models, including \Chronos\texttt{-2}, \TimesFM, \Moirai, \TimerXL, and \Moment.

\subsection{Infini-Attention}\label{apd:infini_attn_background}
Infini-Attention was originally proposed to address the long-context bottleneck for transformers with text data, given that processing is quadratic in sequence length~\citep{munkhdalai2024leave}. This attention mechanism extends the standard transformer architecture by incorporating a compressive memory to capture information from infinitely long contexts in an approach that is bounded in memory and computation. The key innovation is maintaining a separate compressive memory state that accumulates information from tokens in past context windows in addition to standard attention computed for the current context window. For each Transformer layer, Infini-Attention computes scaled dot-product attention for the current context window, $i$,
\begin{align}
\textbf{A}_{\text{local}} = \text{softmax}\left(\frac{\textbf{Q}\textbf{K}^{\top}}{\sqrt{d_k}}\right)\textbf{V}.
\end{align}
It also maintains a memory matrix $\mathbf{M} \in \mathbb{R}^{B \times N \times d_k \times d_v}$ and a normalization vector $\mathbf{z}_{i} \in \mathbb{R}^{B \times N \times d_k \times 1}$, where $B$ is the window batch size, $N$ is the number of attention heads, $d_k$ is the per-head hidden dimension of the key vector, and $d_v$ is the per-head hidden dimension of the value vector. Initially, $\mathbf{M}$ is initialized with zeros. Given query, key, and value matrices $\mathbf{Q}, \mathbf{K}, \mathbf{V}$ for the current context window, $i$, the memory is updated using a linear attention mechanism:
\begin{align}
    \mathbf{M}_i &\leftarrow \mathbf{M}_{i-1} + \phi(\mathbf{K})^{\top}\mathbf{V},\\
    \mathbf{z}_{i} &\leftarrow \mathbf{z}_{i-1} + \sum^L_{l=1} \phi(\mathbf{K}_l),
\end{align}
where the outer product $\mathbf{K}^{\top}\mathbf{V}$ compresses key-value associations into the memory matrix, and $\mathbf{z}$ tracks the normalization term for proper retrieval. Here, $\phi$ corresponds to the nonlinear activation $\text{ELU}(\mathbf{X}) + 1$.
To retrieve information from the compressive memory, the query vectors attend to the memory matrix:
\begin{equation}
    \mathbf{A}_{\text{global}} = \frac{\phi(\mathbf{Q})\mathbf{M}_{i-1}}{\phi(\mathbf{Q})\mathbf{z}_{i-1}},
\end{equation}
where element-wise division by $\phi(\mathbf{Q})\mathbf{z}_{i-1}$ normalizes the retrieved values. This linear attention mechanism enables efficient retrieval with $\mathcal{O}(Pd_k d_v)$ complexity.
A learned linear gating mechanism is used to control the influence of both local attention output and global attention output,
\begin{equation}
    \mathbf{A}_{\text{mixed}} = \sigma(\beta) \odot \mathbf{A}_{\text{global}} + (1 - \sigma(\beta)) \odot \mathbf{A}_{\text{local}},
\end{equation}
where $\beta$ is a learned scalar parameter, $\sigma$ is the sigmoid function, and $\odot$ denotes element-wise multiplication. This gating allows the model to dynamically balance between attending to recent local context and distant global context stored in memory.

\newpage
\section{Dataset Details}
\label{section:apd_data}
\begin{table}[ht!]
\centering
\caption{Summary of forecasting datasets used in our empirical study.}
\label{tab:datasets}
\resizebox{1.0\textwidth}{!}{
\begin{tabular}{l l c c c c c c c c}
\toprule
\textbf{Dataset} & \textbf{Domain} & \textbf{Frequency} & \textbf{\# Unique Series IDs} & \textbf{\# Targets} & \textbf{Min Length} & \textbf{Max Length} & \textbf{Horizon} & \textbf{Val Size} & \textbf{Test Size} \\
\midrule
COVID Deaths~\citep{godahewa2021monash} & Healthcare & Daily & 266 & 1 & 212 & 212 & 30 & 30 & 30 \\
\hline
Simglucose \citep{simglucose_dataset} & Healthcare & 5min & 30 & 1 & 25920 & 25920 & 6 & 2592 & 2592 \\
\hline
Iowa IHOP SMEX02 Windspeed \citep{ISU_IHOP2002_AWOS_2008, ISU_SMEX02_AWOS_2005} & Weather & 5min & 34 & 1 & 23040 & 23040 & 24 & 2304 & 2304 \\
Iowa PLOWS Windspeed \citep{ISU_PLOWS_AWOS_2010} & Weather & 5min & 41 & 1 & 37440 & 37440 & 24 & 3744 & 3744 \\
\hline
\multirow{2}{*}{Jena Weather \citep{aksu2024giftevalbenchmark, wu2021autoformer}} & \multirow{2}{*}{Weather} & Hourly & 1 & 21 & 8784 & 8784 & 48 & 48 & 912 \\
{} & {} & Daily & 1 & 21 & 366 & 366 & 30 & 30 & 60 \\
\hline
\multirow{2}{*}{M-DENSE \citep{aksu2024giftevalbenchmark, medrano2020mdense}} & \multirow{2}{*}{Transportation} & Hourly & 30 & 1 & 17520 & 17520 & 48 & 48 & 960 \\
{} & {} & Daily & 30 & 1 & 730 & 730 & 30 & 30 & 90 \\
\hline
Loop-Seattle~\citep{cui2018loopseattle, cui2019loopseattle} & Transportation & Daily & 323 & 1 & 365 & 365 & 30 & 30 & 60 \\
\hline
\multirow{3}{*}{ETT1 \citep{aksu2024giftevalbenchmark, zhou2021informer}} & \multirow{3}{*}{Energy} & Hourly & 1 & 7 & 17420 & 17420 & 48 & 48 & 960 \\
{}  & {} & Daily & 1 & 7 & 725 & 725 & 30 & 30 & 90 \\
{} & {} & Weekly & 1 & 7 & 103 & 103 & 8 & 8 & 16 \\
\hline
\multirow{3}{*}{ETT2 \citep{aksu2024giftevalbenchmark, zhou2021informer}} & \multirow{3}{*}{Energy} & Hourly & 1 & 7 & 17420 & 17420 & 48 & 48 & 960 \\
{} & {} & Daily & 1 & 7 & 725 & 725 & 30 & 30 & 90 \\
{} & {} & Weekly & 1 & 7 & 103 & 103 & 8 & 8 & 16 \\
\hline
\multirow{3}{*}{Solar~\citep{lai2018solar}} & \multirow{3}{*}{Weather} & Hourly & 137 & 1 & 8760 & 8760 & 48 & 48 & 912 \\
{}  & {} & Daily & 137 & 1 & 365 & 365 & 30 & 30 & 60 \\
{} & {} & Weekly & 137 & 1 & 52 & 52 & 8 & 8 & 8 \\
\bottomrule
\end{tabular}
}
\end{table}

\paragraph{Simglucose.} The Simglucose dataset is from an open-source Python implementation of the FDA-approved UVa/Padova Simulator (2008 version), which is contained in the `simglucose' GitHub repository \citep{simglucose_dataset, visentin2016towards_single_day_simulator}. It provides clinical and biological parameters for 30 patients (10 adults, 10 adolescents, and 10 children) with Type-I diabetes. Clinical parameters include information on age and weight, and biological parameters include information on insulin-glucose kinetics, such as absorption constants. We generate 90 days of data for each patient. In a similar approach to \citep{potosnak2025pkforecast}, the meal schedule used to generate simulated data was based on the Harrison-Benedict equation \citep{harris_benedict_basal_metabolism} with approximately 3 meals per day and no additional snacks. Height data was not provided in the patient parameters, and so was estimated at 140cm, 170cm, and 175cm for adults, adolescents, and children, respectively \citep{potosnak2025pkforecast}. The `Dexcom' continuous glucose monitor (CGM) option was used to obtain glucose readings at 5-minute intervals. We hypothesize that the shared system of ODEs governing insulin-glucose dynamics across patients may enable cross-patient multivariate modeling to capture common physiological patterns, potentially improving forecasting performance across patients.

\paragraph{Iowa Windspeed.} The IHOP and SMEX02 datasets contain 1-minute resolution surface meteorological data from 34 Automated Weather Observing System (AWOS) stations in Iowa from May 13, 2002 through June 25, 2002 and June 1, 2002 through July 31, 2002, respectively \citep{ISU_IHOP2002_AWOS_2008, ISU_SMEX02_AWOS_2005}. We combine these datasets given their overlapping stations and time periods. The Iowa PLOWS dataset contains 1-minute resolution data from 41 AWOS stations from November 1, 2009 to March 10, 2010 \citep{ISU_PLOWS_AWOS_2010} and is treated separately due to different stations and time period. Both datasets are downsampled to 5-minute resolution for wind speed forecasting. These open-source datasets provide well-suited test cases for multivariate modeling across geographically distributed stations, where accurate wind forecasting requires capturing cross-channel spatial dependencies from upwind measurement stations.

\paragraph{Gift-Eval.} 
We include several datasets from the Gift-Eval repository including: ETT1, ETT2, Jena Weather, COVID Deaths, Loop-Seattle, Solar, and M-Dense across different temporal resolutions \citep{aksu2024giftevalbenchmark}. We exclude Loop-Seattle (hourly) due to memory issues for baselines, such as \TimerXL, given the 80GB memory constraint of the A100 GPU. Our dataset selection is guided by the goal of representing several datasets across diverse domains of energy, healthcare, weather, and transportation while satisfying one of two key multivariate settings: (1) modeling multiple variates for a single entity (e.g., air temperature, atmospheric pressure, humidity, and wind direction at a single weather station such as in the case with the Jena Weather dataset), or (2) modeling a single variate across multiple entities (e.g., wind speed measurements across different weather stations such as in the case with the Iowa datasets). Following the GIFT-Eval protocol, we construct our test sets based on the product of the forecast horizon and the number of evaluation windows specified by the benchmark. To ensure efficient hyperparameter tuning, we set the validation set size equal to the forecast horizon for each dataset.

\newpage
\section{Training Methodology and Hyperparameters}
\label{section:apd_training}
In this Appendix section, we expand on the training methodology outlined in Section~\ref{section:experiments}. We train both \MLP-based and Transformer-based models, maintaining consistent hyperparameters where applicable to enable controlled comparison of channel mixing strategies across architectures. Table~\ref{table:hyperparameters_shared} details the hyperparameters used for training configuration, common Transformer-based model hyperparameters, and common \MLP-based model hyperparameters. Parameters not specified in the table are set to the defaults of the original implementations in the \texttt{NeuralForecast} library~\citep{olivares2022library_neuralforecast}.
For \MICA~models incorporating an \MLP~mixing gate (\Moment\ and \PatchTST), we perform hyperparameter tuning using Optuna with 5 trials per model. The optimal hyperparameters are selected based on minimum validation loss, computed using the mean absolute error (MAE) loss function defined in Equation~\ref{eqn:mae}.

\begin{table}[!ht]
    \centering
    \caption{Hyperparameters}
    \label{table:hyperparameters_shared}
    \footnotesize
    \begin{tabular}{lc}
        \toprule
        \textsc{Hyperparameter} & \thead{\textsc{Values}} \\ 
        \midrule
        \multicolumn{2}{c}{\textit{Training configuration}} \\
        \midrule
        Single GPU SGD Batch Size\textsuperscript{*} & $N$ channels \\
        Windows batch size & 64 \\
        Maximum training steps $S_{max}$ & 12,000 \\ 
        Validation check steps & 500 \\
        Initial learning rate & 1e-3 \\
        Learning rate decay & 0.5 \\
        Learning rate step size & 4,000 \\
        Early stop patience steps & 20 \\
        Random seed & \{1, 2, 3, 4, 5\} \\
        \midrule
        \multicolumn{2}{c}{\textit{Common Transformer-based model settings}} \\
        \midrule
        Input size & $2*H$ \\
        Number of encoder layers & 4 \\
        Hidden size & 256 \\
        Number of Heads & 4 \\
        Feed Forward hidden size & 1024 \\
        Dropout & 0.0 \\
        Projection/Head dropout & 0.0 \\
        $d_k$ & 32 \\
        $d_v$ & 32 \\
        Scaler type & Standard \\ 
        RevIN & yes, without affine (standardization) \\
        Patch length (if applicable) & 8                      \\
        Stride (if applicable) & 8                       \\
        Positional Encoding (if applicable) & sincos \\
        \midrule
        \multicolumn{2}{c}{\textit{Common \MLP-based model settings}} \\
        \midrule
        Input size & $2*H$ \\
        Number of blocks & 4 \\
        Hidden size & 256 \\
        Dropout & 0.0 \\
        Scaler type & Standard \\ 
        RevIN & yes, without affine (standardization)  \\
        \midrule
        \multicolumn{2}{c}{\textit{\MICA~\MLP~Gate}} \\
        \midrule
        Input size & $2*H$ \\
        Hidden size & \{128, 256, 384, 512\} \\
        Number of layers & \{2, 3, 4\} \\
        Dropout rate & \{0.0, 0.1, 0.2\} \\
        \bottomrule
    \end{tabular}
\end{table}

\newpage
\section{Supplemental Methods}
\label{section:apd_methods}
\subsection{\MICA~Mixing Gate Variants}\label{apd:infini_gate_variants}

\begin{figure*}[ht!]
    \centering
    \includegraphics[width=0.8\textwidth, trim=2 7 2 210, clip]{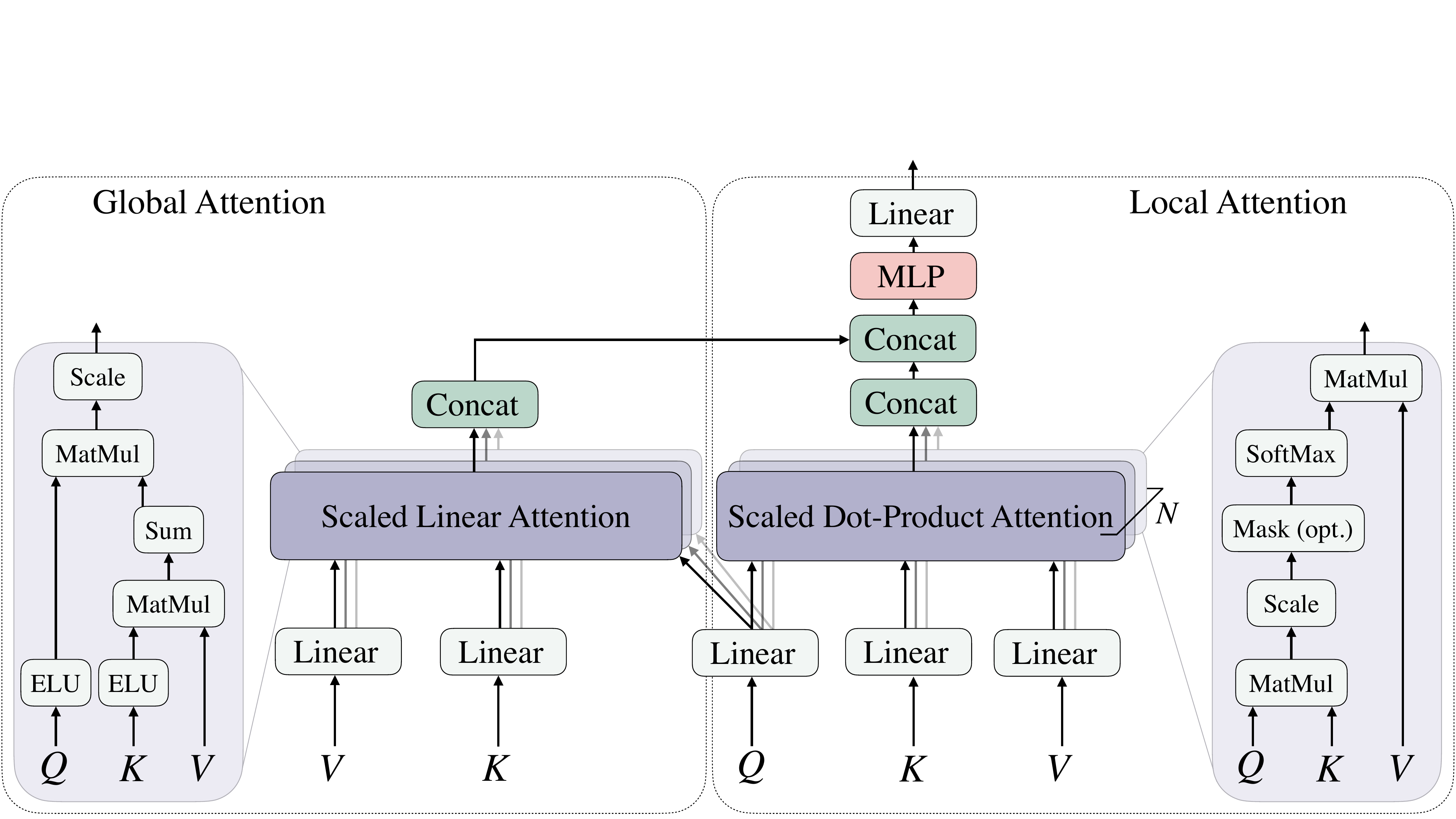}
    
    \caption{\MICA~is a attention-based architectural design proposed to model both local patterns and global cross-channel interactions. $\textbf{Q}$, $\textbf{K}$, and $\textbf{V}$ denote the query, key and value matrices and $N$ denotes the number of attention heads. \MICA~consists of three complementary components: (1) a quadratic attention module that models detailed temporal relationships locally within individual time series, (2) a linear attention module that efficiently globally aggregates information across channels, and (3) a learnable attention mixing gate that adaptively balances local and global information in a computationally efficient manner. This \MICA~variant demonstrates replacing the linear $\beta$ gate with a non-linear \MLP\ gate. An alternative variant concatenates the local projected query information $W_Q(\textbf{Q})$ with the local attention $\textbf{A}_{\text{local}}$ and global attention $\textbf{A}_{\text{global}}$ before passing through the \MLP~mixing gate.}
    \label{fig:method_mica_mlpgate}
    
\end{figure*}

\textbf{Shared $\beta$:} In this configuration, $\beta \in \mathbb{R}^{1 \times 1 \times N \times 1 \times 1}$ is initialized once in the encoder and shared across all Transformer layers. Each attention head $n \in \{1, \hdots, N\}$ has its own learnable parameter, but these parameters remain constant across layers and channels, providing a global baseline for attention mixing throughout the network. This approach adds only $N$ parameters, making it the most parameter-efficient option.

\textbf{Layerwise $\beta$:} Here, $\beta \in \mathbb{R}^{1 \times 1 \times N \times 1 \times 1}$ is independently initialized in each Transformer layer. This allows different layers to learn distinct mixing strategies, enabling the model to adaptively balance local and global attention at different levels of the hierarchical representation. This approach adds $NL$ parameters, where $L$ is the number of Transformer layers.

\textbf{Channelwise $\beta$:} In this approach, $\beta \in \mathbb{R}^{1 \times C \times N \times 1 \times 1}$ is initialized in the encoder and passed to each Transformer layer, where $C$ denotes the number of channels. This configuration enables channel-specific attention mixing while maintaining consistency across layers, allowing the model to capture channel-specific temporal patterns. This approach adds $NC$ parameters.

\textbf{Layerwise Channelwise $\beta$:} This configuration uses $\beta \in \mathbb{R}^{1 \times C \times N \times 1 \times 1}$ initialized independently in each Transformer layer. It provides the finest-grained control by allowing each layer to learn channel-specific and head-specific mixing weights, offering maximum flexibility at the cost of additional parameters. This approach adds $NLC$ parameters.

\textbf{MLP:} The multi-layer perceptron mixing gate replaces the learnable $\beta$ parameter with a neural network that dynamically computes mixing weights. The \MLP\ takes both $\mathbf{A}_{\text{global}}$ and $\mathbf{A}_{\text{local}}$ as input and produces channel-specific mixing weights, enabling adaptive attention mixing based on the computed attention outputs themselves. The \MLP~gate is shown in Fig.~\ref{fig:method_mica_mlpgate}. The \MLP\ concatenates attention outputs and applies a feedforward network. Parameter count scales with the \MLP\ hidden layer size.

\textbf{MLP with Query:} This variant extends the MLP mixing gate shown in Fig.~\ref{fig:method_mica_mlpgate} by additionally incorporating the query matrix~$\mathbf{Q}$ as input to the \MLP~gate. By conditioning on query information, the gate can adapt its mixing strategy based on the semantic content of the queries, enabling context-aware attention mixing that responds to the specific characteristics of the input sequence. Parameter count scales with the \MLP\ hidden layer size.

\newpage
\subsection{\MICA~Channel Exclusion Variant}\label{apd:infini_ciexcl_variants}
\begin{figure*}[ht!]
    \centering
    \includegraphics[width=0.8\textwidth, trim=2 7 2 210, clip]{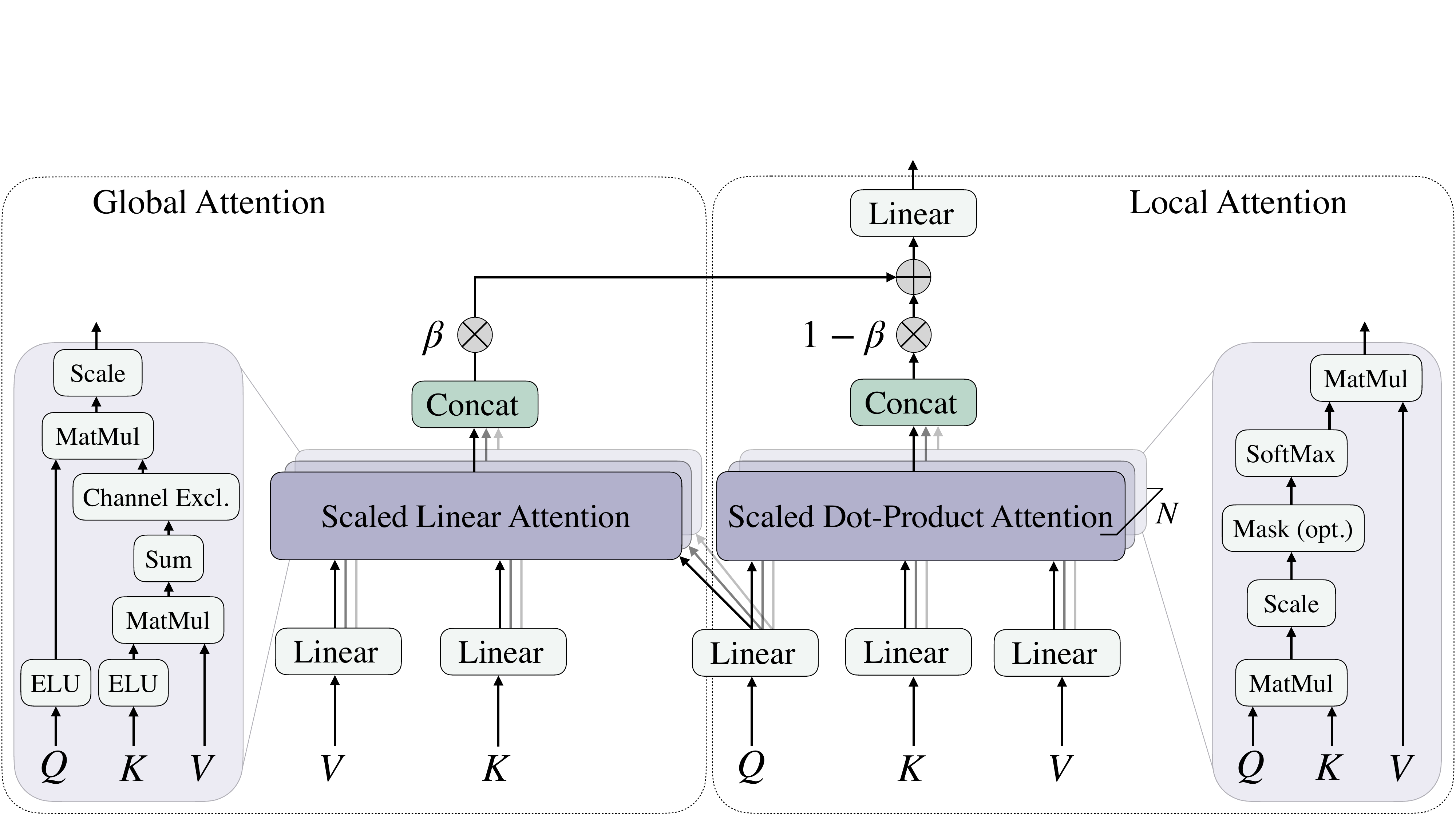}
    
    \caption{\MICA~is a attention-based architectural design proposed to model both local patterns and global cross-channel interactions. $\textbf{Q}$, $\textbf{K}$, and $\textbf{V}$ denote the query, key and value matrices and $N$ denotes the number of attention heads. \MICA~consists of three complementary components: (1) a quadratic attention module that models detailed temporal relationships locally within individual time series, (2) a linear attention module that efficiently globally aggregates information across channels, and (3) a learnable attention mixing gate that adaptively balances local and global information in a computationally efficient manner. This \MICA~variant demonstrates the $\beta$ gate with channel exclusion performed for scaled linear attention. Channel exclusion, denoted by the Channel Excl. block in global attention, removes information from the current channel $i$ within the global attention matrix to avoid redundancy with information already captured through local attention.}
    \label{fig:method_mica_ciexcl}
\end{figure*}

The channel exclusion variant avoids redundancy with information already captured through local attention. Channel exclusion constructs channel-specific $\mathbf{M}$ and $\mathbf{z}$ by excluding each channel's information from the global context. Unlike the base variant where $\mathbf{M} \in \mathbb{R}^{B \times 1 \times N \times d_k \times d_v}$ and $\mathbf{z} \in \mathbb{R}^{B \times 1 \times N \times d_k \times 1}$ are shared across all channels, channel exclusion maintains distinct values for each channel with $\mathbf{M} \in \mathbb{R}^{B \times C \times N \times d_k \times d_v}$ and $\mathbf{z} \in \mathbb{R}^{B \times C \times N \times d_k \times 1}$:
\begin{align} 
\label{eq:M_z_update_exclusion}
    \mathbf{M}_{\text{all}} &= \sum_{c=1}^C \phi(\mathbf{K}^{(c)})^\top \mathbf{V}^{(c)}, \\
    \mathbf{M} &= \mathbf{M}_{\text{all}}^{\oplus C} - \phi(\mathbf{K})^\top \mathbf{V}, \\
    \mathbf{z}_{\text{all}} &= \sum_{c=1}^C \sum_{p=1}^{P} \phi(\mathbf{K}^{(c)}_{p}), \\
    \mathbf{z} &= \mathbf{z}_{\text{all}}^{\oplus C} - \sum_{p=1}^{P}\phi(\mathbf{K}_{p}), \\
    \mathbf{A}_{\text{global}} &= \frac{\phi(\mathbf{Q})\mathbf{M}}{\phi(\mathbf{Q})\mathbf{z} + \epsilon},
\end{align}
where $\mathbf{M}_{\text{all}}$ and $\mathbf{z}_{\text{all}}$ aggregate information across all channels, and $\mathbf{M}_{\text{all}}^{\oplus C}$ and $\mathbf{z}_{\text{all}}^{\oplus C}$ denote these values broadcasted along the channel dimension. The subtraction operations are performed channel-wise, creating channel-specific global contexts where each channel's memory matrix excludes its own information. The global attention has time and memory complexity of $\mathcal{O}(PC)$. Combined with local quadratic attention $\mathcal{O}(P^2)$, the overall complexities become $\mathcal{O}(P^2C + PC)$. Alternatively, the global attention memory complexity can be reduced to $\mathcal{O}(P^2C + P)$ with a time-memory tradeoff by processing channels sequentially. In the sequential case, the time complexity is still in $\mathcal{O}(P^2C + PC)$ but in practice is significantly slower due to the loss of GPU parallelization. The channel exclusion variant is shown in Fig.~\ref{fig:method_mica_ciexcl}. The channel inclusion and exclusion variants can be used with any \MICA\ gate variant discussed in Appendix~\ref{apd:infini_gate_variants}. We study the effect of each channel exclusion in dedicated ablation studies, with full results reported in Appendix~\ref{apd:mica_ciexc_ablation}.



\newpage
\subsection{\MICA~Weighted Channel Aggregation Variants}\label{apd:infini_weighted_channels_variants}

\begin{figure*}[ht!]
    \centering
    \includegraphics[width=0.8\textwidth, trim=2 7 2 210, clip]{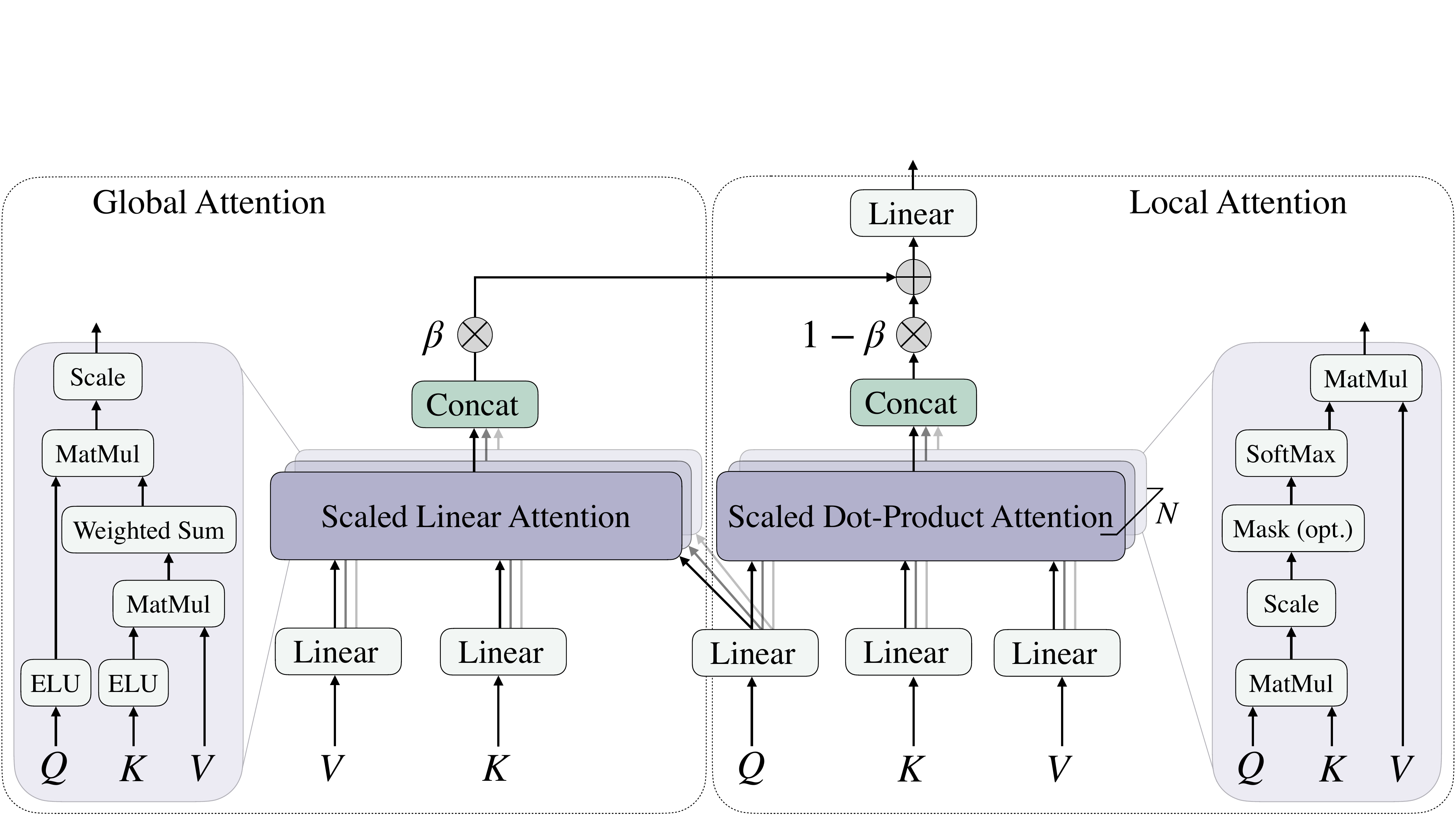}
    
    \caption{\MICA~is a attention-based architectural design proposed to model both local patterns and global cross-channel interactions. $\textbf{Q}$, $\textbf{K}$, and $\textbf{V}$ denote the query, key and value matrices and $N$ denotes the number of attention heads. \MICA~consists of three complementary components: (1) a quadratic attention module that models detailed temporal relationships locally within individual time series, (2) a linear attention module that efficiently globally aggregates information across channels, and (3) a learnable attention mixing gate that adaptively balances local and global information in a computationally efficient manner. This \MICA\ variant applies the $\beta$ gate with weighted channel aggregation for scaled linear attention. We propose two options for computing aggregation weights $\mathbf{w} \in \mathbb{R}^{C}$: learnable static weights, and dynamic weights derived from the query $\mathbf{Q}$.}
    \label{fig:method_mica_weighted_channels}
\end{figure*}

In addition to the uniform summation across channels as proposed in the original linear attention \citep{pmlr-v119-katharopoulos20a}, we propose novel extensions of weighted channel aggregation. We propose two options including learnable static weights $\mathbf{w}\in \mathbb{R}^{C}$ and dynamic weights derived from the query $\mathbf{Q}$. Channel weights could also be user-defined, such as based on domain-specific knowledge. The channel weight variants can be used with any \MICA\ gate variant and in conjunction with either the channel inclusion or channel exclusion \MICA\ variants discussed in Appendices~\ref{apd:infini_gate_variants} and ~\ref{apd:infini_ciexcl_variants}.

Uniform weights entail setting $\mathbf{w} \in \{1\}$ throughout training, whereas learnable static weights use this initial starting point, but set the weight as a network parameter that can be updated throughout training, where higher channel weights determine higher attention output contributions from those channels:
\begin{align}
\label{eqn:channel_weighted_global_attn}
    \mathbf{M} &= \sum_{c=1}^C \mathbf{w}\left( \phi(\mathbf{K}^{(c)})^\top \mathbf{V}^{(c)} \right), \\
    \mathbf{z} &= \sum_{c=1}^C \mathbf{w} \sum_{p=1}^{P} \phi(\mathbf{K}^{(c)}_{p}), \\
    \mathbf{A}_{\text{global}} &= \frac{\phi(\mathbf{Q})\mathbf{M}}{\phi(\mathbf{Q})\mathbf{z} + \epsilon}.
\end{align}
The static channel weight strategy requires the channel count as input for initialization of the weight vector and the final weights are fixed post-training. These limitations prevent this method from being dynamic to different channel counts and new contexts. To address this, we propose an alternative strategy that uses the query $\mathbf{Q} \in \mathbb{R}^{B \times C \times N \times P \times d_q}$ to derive channel weights, supporting fully dynamic weights that can adapt on the fly to various channel counts and contexts. Query-based weights are computed as
\begin{align}
    \mathbf{w} = W_{d_q}\!\left( \sum_{p=1}^{P} \mathbf{Q} \right)
\end{align}
where $W_{d_q}$ is a linear projection function and $d_q$ corresponds to the per-head query hidden dimension, producing $\mathbf{w} \in \mathbb{R}^{B \times C \times H \times 1 \times 1}$. The query-based weights are first computed and then applied to generate $\mathbf{M}$ and $\mathbf{z}$ as shown in Equation~\ref{eqn:channel_weighted_global_attn}. The channel exclusion variant is shown in Fig.~\ref{fig:method_mica_weighted_channels}. We study the effect of each channel weighting strategy in dedicated ablation studies, with full results reported in Appendix~\ref{apd:mica_channel_weight_ablation}.

\newpage
\section{A Memory-Efficient Enhancement to MICA: Algorithm Sketch}
\label{section:apd_flashatt}
\noindent We present an algorithmic sketch for a hardware-aware version of \MICA\ designed to handle high-dimensional multivariate settings and long temporal contexts. The main idea lies in the fusion of local and global attention mechanisms into a specialized two-pass kernel. In the first pass (\textit{Global Reduction}), a global compressive memory state $\mathbf{M}_{\text{all}}$ is computed by aggregating keys and values across all channels in a single linear sweep with complexity $\mathcal{O}(PC)$. In the second pass (\textit{Fused Channel-Wise Attention}), each channel is processed via a theoretical IO-aware kernel. Within this stage, local temporal patterns are computed through a tiled FlashAttention mechanism, while a local linear correction term for strict channel exclusion is simultaneously determined. By reusing the loaded keys and values from SRAM for both operations, redundant HBM access is eliminated, and the linear correction is computed with negligible overhead. Finally, an on-chip convex combination of the local and global outputs is performed using the learned mixing gate $\beta$, whereby only the final representations are written to memory. Through this sketched approach, spatial and temporal complexity are decoupled, providing a path for linear scaling with channel count $C$ and optimized quadratic scaling with sequence length $L$. The following algorithm pertain to the channel inclusion \MICA\ variant.

\begin{algorithm}[H]
\caption{Optimized Fused MICA (Flash-Local + Linear-Global)}
\label{alg:mica_fused}
\begin{algorithmic}[1]
\REQUIRE $\mathbf{Q, K, V} \in \mathbb{R}^{B \times C \times N \times P \times d}$, mixing gate $\beta$
\ENSURE $\mathbf{Y} \in \mathbb{R}^{B \times C \times N \times P \times d}$

\STATE \textbf{Hyperparameters:} Block sizes $B_r, B_c$ for tiling; $\phi(\mathbf{X}) = \text{ELU}(\mathbf{X}) + 1$.
\vspace{0.1cm}
\hrule
\vspace{0.1cm}
\STATE \textbf{Pass 1: Global Memory Reduction (Kernel 1)}
\vspace{0.1cm}
\hrule
\vspace{0.1cm}

\FOR{batch $b$, head $n$ \textbf{in parallel}}
    \STATE Initialize $\mathbf{M}_{\text{all}} \leftarrow \mathbf{0}^{d \times d}, \mathbf{z}_{\text{all}} \leftarrow \mathbf{0}^{d}$
    \FOR{channel $c = 1$ \textbf{to} $C$}
        \STATE Load $\mathbf{K}^{(c)}, \mathbf{V}^{(c)}$ from HBM
        \STATE $\mathbf{K}' \leftarrow \phi(\mathbf{K}^{(c)})$
        \STATE $\mathbf{M}_{\text{all}} \leftarrow \mathbf{M}_{\text{all}} + (\mathbf{K}')^\top \mathbf{V}^{(c)}$ \COMMENT{Global State Accumulation}
        \STATE $\mathbf{z}_{\text{all}} \leftarrow \mathbf{z}_{\text{all}} + \sum_p \mathbf{K}'_p$
    \ENDFOR
    \STATE Store $\mathbf{M}_{\text{all}}, \mathbf{z}_{\text{all}}$ to HBM
\ENDFOR

\vspace{0.2cm}
\hrule
\vspace{0.1cm}
\STATE \textbf{Pass 2: Fused Channel-Wise Attention (Kernel 2)}
\vspace{0.1cm}
\hrule
\vspace{0.1cm}

\FOR{channel $c = 1$ \textbf{to} $C$ \textbf{in parallel}}
    \STATE Load $\mathbf{M}_{\text{all}}, \mathbf{z}_{\text{all}}$ into \textbf{SRAM}
    \FOR{query block $i = 1$ \textbf{to} $P/B_r$}
        \STATE Initialize Local Acc: $\mathbf{O}_{\text{quad}}^{(i)} \leftarrow \mathbf{0}, \ell^{(i)} \leftarrow \mathbf{0}, m^{(i)} \leftarrow -\infty$
        
        \FOR{key-value block $j = 1$ \textbf{to} $P/B_c$}
            \STATE Load $\mathbf{K}^{(c)}_{[j]}, \mathbf{V}^{(c)}_{[j]}$ from channel $c$ to \textbf{SRAM}
            
            \vspace{0.1cm}
            \STATE \textit{// Quadratic Local Attention (FlashAttention Logic)}
            \STATE Compute $\mathbf{S}_{ij} = \mathbf{Q}^{(c)}_{[i]} (\mathbf{K}^{(c)}_{[j]})^\top \cdot \text{scale}$
            \STATE Update softmax stats $m^{(i)}, \ell^{(i)}$ and $\mathbf{O}_{\text{quad}}^{(i)}$ using online softmax reduction
        \ENDFOR
        
        \STATE $\mathbf{O}_{\text{global}}^{(i)} = (\phi(\mathbf{Q}^{(c)}_{[i]}) \mathbf{M}_{\text{all}}) / (\phi(\mathbf{Q}^{(c)}_{[i]}) \mathbf{z}_{\text{all}} + \epsilon)$
        
        \STATE \textbf{Mixing:} $\mathbf{Y}^{(c)}_{[i]} = \sigma(\beta) \cdot \mathbf{O}_{\text{global}}^{(i)} + (1 - \sigma(\beta)) \cdot \mathbf{O}_{\text{quad}}^{(i)}$
        \STATE Write $\mathbf{Y}^{(c)}_{[i]}$ to HBM
    \ENDFOR
\ENDFOR
\end{algorithmic}
\end{algorithm}

\noindent\textbf{Online Softmax Reduction:} The update in line X maintains running statistics for numerically stable softmax computation across blocks. Specifically, for each block $j$,  update: $m^{\text{new}} = \max(m^{\text{old}}, \max(\mathbf{S}_{ij}))$ and $\ell^{\text{new}} = \exp(m^{\text{old}} - m^{\text{new}}) \cdot \ell^{\text{old}} + \sum \exp(\mathbf{S}_{ij} - m^{\text{new}})$, following the FlashAttention approach~\citep{dao2022flashattention}.

\newpage
\section{Computational Parameter Study}
\label{section:apd_computation}
We measure computational efficiency using three metrics: (1) GFLOPs (giga floating-point operations) computed using PyTorch's FlopCounterMode during a single forward pass, (2) trainable parameters (millions) counted by summing all learnable model weights, and (3) inference speed (milliseconds) measured as the average latency over 100 runs with 10 warmup iterations using CUDA events for precise GPU timing. 

\subsection{Channel Count Efficiency Analysis}\label{apd:channel_count_efficiency_analysis}

We evaluate computational efficiency across two representative settings: a low-dimensional ($C=7$, $H=48$, $L=96$) and a high-dimensional ($C=600$, $H=48$, $L=96$) setting. Tables~\ref{tab:mica_variant_flops_n7} and ~\ref{tab:mica_variant_flops_n600} presents the computational cost of \MICA~variants relative to their univariate baselines for the low- and high-dimensional settings, respectively. Tables~\ref{tab:baseline_flops_n7} and ~\ref{tab:baseline_flops_n600} presents the computational cost of \MICA~(\MLP-Query gate) relative to their multivariate baselines for the low- and high-dimensional settings, respectively. 

\MICA\ efficiency gains are most pronounced in the high-dimensional setting ($C=600$) as shown in Table~\ref{tab:mica_variant_flops_n600}. The Multivariate Head increases parameters by $35.2\times$ and inference latency by $11.2\times$ compared to univariate \PatchTST. In contrast, \PatchTST-\MICA\ (layerwise $\beta$) increases parameters by $1.0\times$ (no substantial change) and inference by only $1.15\times$. Similarly, for \Moment, the Multivariate Head inflates parameters by $9.3\times$ compared to $1.0\times$ for all $\beta$-based gate variants. The \MLP-based gating variants incur higher GFLOPs due to the additional 2-layer model gate (hidden size 128), but remain more parameter-efficient than the Multivariate Head across both architectures. This demonstrates that as the number of channels grows, \MICA\ efficiency gains become crucial.

In the high-dimensional setting ($
C=600$), \MICA\ remains computationally competitive with multivariate baselines as shown in Table~\ref{tab:baseline_flops_n600}. \PatchTST-\MICA\ (\MLP-Query) requires 45.934 GFLOPs and 2.861M parameters, compared to \Crossformer\ ($4.5\times$ the GFLOPs, $5.0\times$ the parameters), \TimerXL\ ($3.6\times$ the GFLOPs),  \Chronos\texttt{-2} ($33.4\times$ the GFLOPs, $41.9\times$ the parameters). In terms of inference speed, \PatchTST-\MICA\ runs in 7.939ms, $2.9\times$ faster than \Crossformer, $9.7\times$ faster than \TimerXL, and $17.3\times$ faster than \Chronos\texttt{-2}, though slower than \iTransformer\ and \MLP-based models.

The efficiency advantages of \MICA\ also extend to low-dimensional settings ($C=7$) as shown in Table~\ref{tab:mica_variant_flops_n7}. The $\beta$-based variants introduce negligible overhead, increasing GFLOPs by only $1.01\times$ for \PatchTST~and $1.04\times$ for \Moment~with no increase in parameter count. In contrast, the Multivariate Head baseline increases parameters by $1.34\times$ for \PatchTST~and $1.08 \times$ for \Moment. 

In the low-dimensional setting ($C=7$), \MICA\ maintains competitive efficiency with multivariate baselines as shown in Table~\ref{tab:baseline_flops_n7}. \PatchTST-\MICA\ (\MLP-Query) requires 0.536 GFLOPs and 2.861M parameters, comparable to \TimerXL, though slightly less efficient ($1.08\times$ the GFLOPs, $1.02\times$ the parameters). \PatchTST-\MICA\ is substantially lighter than \Crossformer\ ($4.5\times$ the GFLOPs, $5.0\times$ the parameters) and \Chronos\texttt{-2} ($30.5\times$ the GFLOPs, $41.9\times$ the parameters). Inference speed for \PatchTST-\MICA\ is $3.8\times$ faster than \Crossformer\ and $4.6\times$ faster than \Chronos\texttt{-2}, though slower than \TimerXL, \iTransformer, and \MLP-based models.

Note that \Chronos\texttt{-2} GFLOPs are computed using the same 
\texttt{FlopCounterMode} methodology as all other models, but may underestimate 
true computation as the model is imported from HuggingFace \citep{wolf2020transformers} rather than natively implemented in \texttt{NeuralForecast} \citep{olivares2022library_neuralforecast}, and processing operations outside the core forward pass may not be fully captured.

\begin{table}[ht]
\newcommand{\lastmodel}{}
\centering
\caption{Model complexity comparison among \MICA~variants for the low-dimensional setting ($C=7$, $H=48$, $L=96$). Percentage increases ($\uparrow \text{x.x}\%$) are computed relative to the original univariate implementation.}
\label{tab:mica_variant_flops_n7}
\resizebox{0.83\textwidth}{!}{
\begin{tabular}{ll|llll}
\toprule
\textbf{Model} & \textbf{Variant} &\textbf{GFLOPs} & \textbf{Trainable Parameters (M)} & \textbf{Inference Speed (ms)} \\
\midrule
\begin{filecontents*}{tables/flops_mica_table_n7.csv}
model,variant,gflops,trainable_params,inference_speed
PatchTST,Univariate,0.482,2.795,2.978
,Multivariate Head,0.482 ($\uparrow 0.0\%$),3.754 ($\uparrow 34.3\%$),3.502 ($\uparrow 17.6\%$)
,MICA (Shared $\beta$),0.488 ($\uparrow 1.2\%$),2.795 ($\uparrow 0.0\%$),4.496 ($\uparrow 51.0\%$)
,MICA (Channelwise $\beta$),0.488 ($\uparrow 1.2\%$),2.795 ($\uparrow 0.0\%$),4.456 ($\uparrow 49.6\%$)
,MICA (Layerwise $\beta$),0.488 ($\uparrow 1.2\%$),2.795 ($\uparrow 0.0\%$),4.624 ($\uparrow 55.3\%$)
,MICA (Layerwise Channelwise $\beta$),0.488 ($\uparrow 1.2\%$),2.795 ($\uparrow 0.0\%$),4.498 ($\uparrow 51.0\%$)
,MICA (MLP),0.524 ($\uparrow 8.7\%$),2.845 ($\uparrow 1.8\%$),4.725 ($\uparrow 58.7\%$)
,MICA (MLP w/ Query),0.536 ($\uparrow 11.2\%$),2.861 ($\uparrow 2.4\%$),4.766 ($\uparrow 60.0\%$)
Moment,Univariate,0.580,11.535,3.396
,Multivariate Head,0.580 ($\uparrow 0.0\%$),12.494 ($\uparrow 8.3\%$),3.878 ($\uparrow 14.2\%$)
,MICA (Shared $\beta$),0.604 ($\uparrow 4.1\%$),11.535 ($\uparrow 0.0\%$),5.432 ($\uparrow 60.0\%$)
,MICA (Channelwise $\beta$),0.604 ($\uparrow 4.1\%$),11.535 ($\uparrow 0.0\%$),5.083 ($\uparrow 49.7\%$)
,MICA (Layerwise $\beta$),0.604 ($\uparrow 4.1\%$),11.535 ($\uparrow 0.0\%$),5.183 ($\uparrow 52.6\%$)
,MICA (Layerwise Channelwise $\beta$),0.604 ($\uparrow 4.1\%$),11.535 ($\uparrow 0.0\%$),5.407 ($\uparrow 59.2\%$)
,MICA (MLP),0.676 ($\uparrow 16.6\%$),11.634 ($\uparrow 0.9\%$),5.494 ($\uparrow 61.8\%$)
,MICA (MLP w/ Query),0.699 ($\uparrow 20.5\%$),11.667 ($\uparrow 1.1\%$),5.412 ($\uparrow 59.4\%$)
\end{filecontents*}
\csvreader[
    late after line=\\,
    late after last line=\\\bottomrule
]{tables/flops_mica_table_n7.csv}
{model=\model,
 variant=\variant,
 gflops=\gflops,
 trainable_params=\trainable,
 inference_speed=\infspeed
 }
{
\ifthenelse{\equal{\model}{Moment}}%
    {\ifthenelse{\equal{\lastmodel}{Moment}}%
        {}
        {\\[-8pt]\hline\\[-8pt]}
    }%
    {}
\texttt{\model} & \variant & \gflops & \trainable & \infspeed
}
\end{tabular}
}
\end{table}

\begin{table}[ht]
\newcommand{\lastmodel}{}
\centering
\caption{Model complexity comparison among \MICA~variants for the high-dimensional setting ($C=600$). Percentage increases ($\uparrow \text{x.x}\%$) are computed relative to the original univariate implementation.}
\label{tab:mica_variant_flops_n600}
\resizebox{0.83\textwidth}{!}{
\begin{tabular}{ll|llll}
\toprule
\textbf{Model} & \textbf{Variant} &\textbf{GFLOPs} & \textbf{Trainable Parameters (M)} & \textbf{Inference Speed (ms)} \\
\midrule
\begin{filecontents*}{tables/flops_mica_table_n600_is96.csv}
model,variant,gflops,trainable_params,inference_speed
PatchTST,Univariate,41.326,2.795,5.499
,Multivariate Head,41.326 ($\uparrow 0.0\%$),98.511 ($\uparrow 3424.5\%$),61.638 ($\uparrow 1020.9\%$)
,MICA (Shared $\beta$),41.845 ($\uparrow 1.3\%$),2.795 ($\uparrow 0.0\%$),9.329 ($\uparrow 69.6\%$)
,MICA (Channelwise $\beta$),41.845 ($\uparrow 1.3\%$),2.798 ($\uparrow 0.1\%$),6.352 ($\uparrow 15.5\%$)
,MICA (Layerwise $\beta$),41.845 ($\uparrow 1.3\%$),2.795 ($\uparrow 0.0\%$),6.326 ($\uparrow 15.0\%$)
,MICA (Layerwise Channelwise $\beta$),41.845 ($\uparrow 1.3\%$),2.805 ($\uparrow 0.4\%$),7.559 ($\uparrow 37.5\%$)
,MICA (MLP),44.912 ($\uparrow 8.7\%$),2.845 ($\uparrow 1.8\%$),12.991 ($\uparrow 136.2\%$)
,MICA (MLP w/ Query),45.934 ($\uparrow 11.2\%$),2.861 ($\uparrow 2.4\%$),10.390 ($\uparrow 88.9\%$)
Moment,Univariate,49.712,11.535,5.019
,Multivariate Head,49.712 ($\uparrow 0.0\%$),107.250 ($\uparrow 829.8\%$),49.499 ($\uparrow 886.2\%$)
,MICA (Shared $\beta$),51.773 ($\uparrow 4.1\%$),11.535 ($\uparrow 0.0\%$),7.162 ($\uparrow 42.7\%$)
,MICA (Channelwise $\beta$),51.773 ($\uparrow 4.1\%$),11.537 ($\uparrow 0.0\%$),7.250 ($\uparrow 44.5\%$)
,MICA (Layerwise $\beta$),51.773 ($\uparrow 4.1\%$),11.535 ($\uparrow 0.0\%$),8.070 ($\uparrow 60.8\%$)
,MICA (Layerwise Channelwise $\beta$),51.773 ($\uparrow 4.1\%$),11.545 ($\uparrow 0.1\%$),6.918 ($\uparrow 37.8\%$)
,MICA (MLP),57.907 ($\uparrow 16.5\%$),11.634 ($\uparrow 0.9\%$),8.629 ($\uparrow 71.9\%$)
,MICA (MLP w/ Query),59.952 ($\uparrow 20.6\%$),11.667 ($\uparrow 1.1\%$),13.803 ($\uparrow 175.0\%$)
\end{filecontents*}
\csvreader[
    late after line=\\,
    late after last line=\\\bottomrule
]{tables/flops_mica_table_n600_is96.csv}
{model=\model,
 variant=\variant,
 gflops=\gflops,
 trainable_params=\trainable,
 inference_speed=\infspeed
 }
{
\ifthenelse{\equal{\model}{Moment}}%
    {\ifthenelse{\equal{\lastmodel}{Moment}}%
        {}
        {\\[-8pt]\hline\\[-8pt]}
    }%
    {}
\texttt{\model} & \variant & \gflops & \trainable & \infspeed
}
\end{tabular}
}
\end{table}

\begin{table}[ht]
\newcommand{\lastmodel}{}
\centering
\caption{Model complexity comparison among multivariate models for the low-dimensional setting ($C = 7$).}
\label{tab:baseline_flops_n7}
\resizebox{0.83\textwidth}{!}{
\begin{tabular}{ll|llll}
\toprule
\textbf{Model} & \textbf{Variant} &\textbf{GFLOPs} & \textbf{Trainable Parameters (M)} & \textbf{Inference Speed (ms)} \\
\midrule
\begin{filecontents*}{tables/flops_baseline_table_n7.csv}
model,variant,gflops,trainable_params,inference_speed
PatchTST,MICA (MLP w/ Query),0.536,2.861,4.713
Moment,MICA (MLP w/ Query),0.699,11.667,5.311
iTransformer,,0.037,2.671,2.157
iTransformer-T5,,0.045,11.41,3.241
Crossformer,,2.428,14.399,18.033
Timer-XL,,0.497,2.796,4.487
TSMixer,,0.003,0.068,1.634
TimeMixer,,1.256,4.324,4.994
MLP,,0.001,0.456,0.186
Chronos-2,,16.372,120.0,21.635
\end{filecontents*}
\csvreader[
    late after line=\\,
    late after last line=\\\bottomrule
]{tables/flops_baseline_table_n7.csv}
{model=\model,
 variant=\variant,
 gflops=\gflops,
 trainable_params=\trainable,
 inference_speed=\infspeed
 }
{
\texttt{\model} & \variant & \gflops & \trainable & \infspeed
}
\end{tabular}
}
\end{table}

\begin{table}[ht]
\newcommand{\lastmodel}{}
\centering
\caption{Model complexity comparison among multivariate models for the high-dimensional setting ($C = 600$).}
\label{tab:baseline_flops_n600}
\resizebox{0.83\textwidth}{!}{
\begin{tabular}{ll|llll}
\toprule
\textbf{Model} & \textbf{Variant} &\textbf{GFLOPs} & \textbf{Trainable Parameters (M)} & \textbf{Inference Speed (ms)} \\
\midrule
\begin{filecontents*}{tables/flops_baseline_table_n600_is96.csv}
model,variant,gflops,trainable_params,inference_speed
PatchTST,MICA (MLP w/ Query),45.934,2.861,7.939
Moment,MICA (MLP w/ Query),59.952,11.667,8.143
iTransformer,,3.927,2.671,2.137
iTransformer-T5,,5.294,11.41,3.369
Crossformer,,205.475,14.399,23.145
Timer-XL,,165.718,2.796,77.208
TSMixer,,0.286,2.196,1.664
TimeMixer,,1.438,4.931,4.944
MLP,,0.045,22.345,0.187
Chronos-2,,1534.486,120.0,137.048
\end{filecontents*}
\csvreader[
    late after line=\\,
    late after last line=\\\bottomrule
]{tables/flops_baseline_table_n600_is96.csv}
{model=\model,
 variant=\variant,
 gflops=\gflops,
 trainable_params=\trainable,
 inference_speed=\infspeed
 }
{
\texttt{\model} & \variant & \gflops & \trainable & \infspeed
}
\end{tabular}
}
\end{table}

\begin{figure*}[ht!]
    \centering
    \includegraphics[width=0.85\textwidth, trim=0 10 0 0, clip]{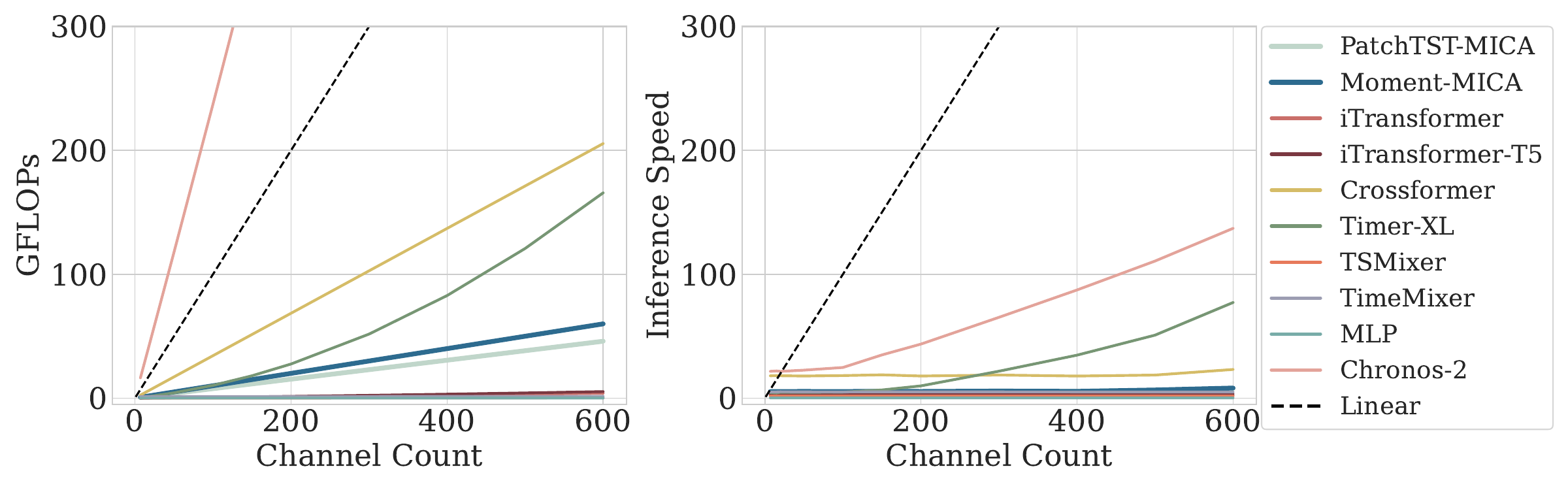}
    \caption{GFLOPs and inference speed (ms) vs. channel count $C \in [7, 600]$ for multivariate models, with context length $L=96$ and horizon $H=48$ held constant. \MLP-based methods (\TSMixer, \MLP) are the most computationally efficient across all channel counts. Among Transformer-based models, \MICA\ variants scale the best with the exception of \iTransformer while substantially outperforming \Crossformer, \TimerXL, and \Chronos-2 in both GFLOPs and inference speed. The dashed line indicates linear scaling for reference.}    \label{fig:scaling_channels_comparison}
\end{figure*}

\subsection{Context Length Efficiency Analysis}\label{apd:seqlen_efficiency_analysis}

\begin{figure*}[ht!]
    \centering
    \includegraphics[width=0.85\textwidth, trim=0 10 0 0, clip]{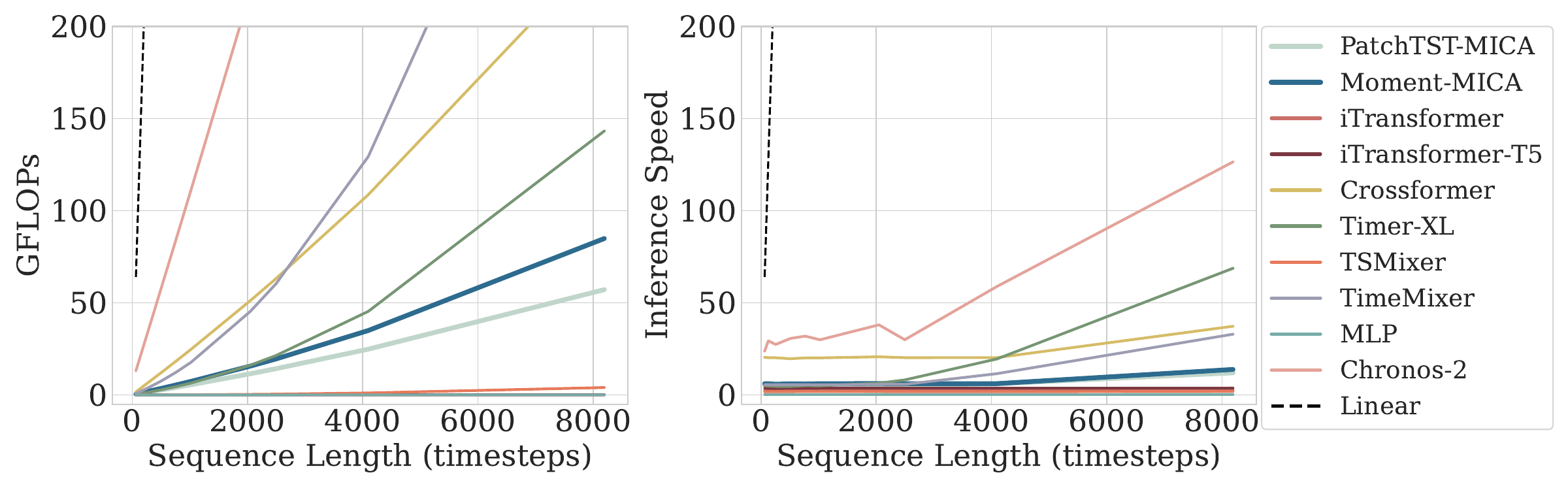}
    \caption{GFLOPs and inference time (ms) vs. context length $L \in [64,8192]$ for multivariate models, with channel count ($C=7$) and horizon ($H=48$) held constant. \MLP-based methods (\TSMixer, \MLP) are the most computationally efficient across all channel counts. One exception is \TimeMixer, which is outperformed by \MICA\ models in both GFLOPs and inference speed. Among Transformer-based models, \MICA\ variants scale the best with the exception of \iTransformer while substantially outperforming \Crossformer, \TimerXL, and \Chronos-2 in both GFLOPs and inference speed. The dashed line indicates linear scaling for reference.}    \label{fig:scaling_seqlen_comparison}
\end{figure*}

We evaluate computational efficiency in context length ($L$) across multivariate models with various context lengths: $L=[64, 128, 256, 512, 768, 1024, 2048, 2496, 4096, 8192]$ holding channel count ($C=7$) and horizon ($H=48$) constant. We find that \MICA\ models scale better than all Transformer-based models regarding longer context length with the exception of \iTransformer as shown in Figs.~\ref{fig:scaling_seqlen_comparison} and~\ref{fig:scaling_combined_log}. At $L=8192$, \PatchTST-\MICA\ demonstrates $4.3\times$ less GFLOPs \Crossformer, $2.5\times$ less than \TimerXL, $15.9\times$ less than \Chronos\texttt{-2} as shown in Table~\ref{tab:baseline_flops_n7_is8192}. Notably, \PatchTST-\MICA\ scales more efficiently with sequence length than \TimeMixer\ (\MLP-based model) with $7.2\times$ less GFLOPs, due to the univariate backbone’s use of patch-based tokenization. \PatchTST-\MICA\ also scales more efficiently in terms of inference speed compared to \Crossformer\ ($3.1\times$), \TimerXL\ ($5.7\times$), \Chronos\texttt{-2} ($10.6\times$), and TimeMixer ($2.7\times$).

\begin{table}[ht!]
\newcommand{\lastmodel}{}
\centering
\caption{Model complexity comparison among multivariate models for a large context length setting ($L = 8192$).}
\label{tab:baseline_flops_n7_is8192}
\resizebox{0.83\textwidth}{!}{
\begin{tabular}{ll|llll}
\toprule
\textbf{Model} & \textbf{Variant} &\textbf{GFLOPs} & \textbf{Trainable Parameters (M)} & \textbf{Inference Speed (ms)} \\
\midrule
\begin{filecontents*}{tables/flops_baseline_table_n7_is8192.csv}
model,variant,gflops,trainable_params,inference_speed
PatchTST,MICA (MLP w/ Query),57.124,15.297,11.968
Moment,MICA (MLP w/ Query),84.89,24.102,13.825
iTransformer,,0.066,4.743,2.476
iTransformer-T5,,0.074,13.483,3.676
Crossformer,,243.449,16.731,37.266
Timer-XL,,143.256,15.231,68.732
TSMixer,,3.998,269.794,1.888
TimeMixer,,413.934,693.37,32.957
MLP,,0.03,14.964,0.227
Chronos-2,,906.223,120.0,126.501
\end{filecontents*}
\csvreader[
    late after line=\\,
    late after last line=\\\bottomrule
]{tables/flops_baseline_table_n7_is8192.csv}
{model=\model,
 variant=\variant,
 gflops=\gflops,
 trainable_params=\trainable,
 inference_speed=\infspeed
 }
{
\texttt{\model} & \variant & \gflops & \trainable & \infspeed
}
\end{tabular}
}
\end{table}

\newpage
\section{Supplemental Results}
\label{section:apd_results}
\subsection{Additional Point Forecasting Results}
\vspace{0.3em}
\newcommand{\prevdatasettwo}{}
\begin{table*}[ht!]
\centering
\caption{Forecasting RMSE averaged over 5 random seeds with standard deviation in parentheses. Methods without standard deviation have deterministic solutions. \MICA~results correspond to the MLP-Query Gate. Best results are shown in \textbf{bold}. Second best results are \underline{underlined}. \textcolor{blue}{Blue} results indicate lower forecast error of \MICA~compared with the univariate model counterpart. The average rank across datasets for deep learning models is presented at the bottom of the table, with the best result in \textbf{bold} and second best \underline{underlined}.}
\scriptsize
\resizebox{1.0\textwidth}{!}{
\begin{tabular}{ll|ccccccccccccc}
\toprule
\textbf{Dataset} & \textbf{Freq.} & \textbf{\Moment} & \textbf{\Moment-\MICA} & \textbf{\PatchTST} & \textbf{\PatchTST-\MICA} & \textbf{\iTransformer} & \textbf{\iTransformer-T5} & \textbf{\Crossformer} & \textbf{\TimerXL} &\textbf{\TSMixer} & \textbf{\TimeMixer} & \textbf{\MLP} & \textbf{\texttt{\Chronos-2}} & \textbf{\AutoETS} \\
\midrule
\addlinespace[6pt]
\begin{filecontents*}{tables/rmse_main.csv}
Dataset,Frequency,moment_u,moment_u_se,moment_m,moment_m_se,patchtst_u,patchtst_u_se,patchtst_m,patchtst_m_se,itransformer_m,itransformer_m_se,itransformer_t5_m,itransformer_t5_m_se,crossformer_m,crossformer_m_se,timerxl_m,timerxl_m_se,tsmixer_m,tsmixer_m_se,timemixer_m,timemixer_m_se,mlp_m,mlp_m_se,chronos_m,chronos_m_se,autoets,autoets_se
Simglucose,5min,9.488,0.064,\textcolor{blue}{8.254},0.042,9.838,0.604,\textcolor{blue}{\underline{8.050}},0.090,8.176,0.098,8.365,0.052,\textbf{7.993},0.263,8.819,0.064,10.779,0.068,11.429,0.174,12.981,0.047,13.857,-,14.545,-
COVID Deaths,D,975.610,188.437,\textcolor{blue}{604.497},133.317,1075.895,653.409,\textcolor{blue}{807.538},272.668,2065.562,1652.060,1005.541,366.237,927.411,381.748,986.290,345.592,3718.205,2420.913,\textbf{516.775},8.571,2633.906,46.835,568.228,-,\underline{538.832},-
Iowa IHOP SMEX02,5min,2.438,0.011,\textcolor{blue}{2.351},0.007,2.488,0.027,\textcolor{blue}{\textbf{2.339}},0.011,2.757,0.002,2.398,0.014,2.474,0.089,2.729,0.014,\underline{2.341},0.003,2.410,0.017,2.511,0.014,2.463,-,2.530,-
Iowa PLOWS,5min,1.936,0.004,\textcolor{blue}{1.878},0.008,1.944,0.007,\textcolor{blue}{\underline{1.869}},0.010,2.131,0.003,1.897,0.008,1.917,0.015,2.072,0.015,\textbf{1.869},0.002,1.887,0.002,2.001,0.003,2.006,-,2.007,-
Jena Weather,H,34.582,0.458,\textcolor{blue}{\textbf{33.999}},0.482,34.955,0.502,\textcolor{blue}{\underline{34.131}},0.485,36.421,0.362,35.115,0.379,34.932,0.214,35.134,0.256,40.629,0.376,38.930,0.435,41.965,0.688,44.152,-,50.963,-
,D,\underline{25.762},1.145,28.463,0.616,27.659,0.399,\textcolor{blue}{27.426},0.832,29.887,1.066,27.117,1.091,28.666,1.056,27.204,0.663,33.499,1.565,30.687,2.123,38.271,1.948,29.543,-,\textbf{22.570},-
M-DENSE,H,212.781,2.432,\textcolor{blue}{\underline{199.976}},5.530,213.261,4.906,\textcolor{blue}{\textbf{197.044}},3.055,211.542,0.788,215.811,1.433,338.726,121.596,212.373,0.953,220.016,1.618,230.314,1.282,246.623,1.467,265.483,-,316.829,-
,D,106.914,1.256,\textcolor{blue}{104.616},0.538,103.693,0.452,\textcolor{blue}{101.776},0.705,102.436,0.589,\underline{97.383},0.749,102.145,3.502,100.070,0.529,97.713,0.518,106.955,2.703,108.179,1.006,\textbf{93.433},-,105.462,-
Loop-Seattle,D,4.534,0.014,\textcolor{blue}{4.423},0.010,4.705,0.015,\textcolor{blue}{4.458},0.007,4.964,0.041,4.573,0.041,4.607,0.011,4.612,0.030,4.503,0.049,\underline{4.308},0.079,4.449,0.022,\textbf{4.284},-,4.478,-
ETT1,H,11.294,0.177,11.778,0.354,10.444,0.026,\textcolor{blue}{\underline{10.424}},0.292,11.593,0.154,11.419,0.119,11.343,0.228,10.982,0.102,\textbf{10.359},0.077,11.606,0.262,10.868,0.090,10.642,-,23.454,-
,D,230.474,3.187,234.068,3.012,228.325,2.322,231.300,3.275,240.348,2.754,250.081,3.091,317.775,43.342,\underline{223.024},2.071,279.778,9.384,236.715,8.341,273.694,9.096,\textbf{213.547},-,248.501,-
,W,1360.145,63.914,1375.500,53.049,1404.786,37.245,\textcolor{blue}{\underline{1342.545}},64.424,1397.429,73.213,1431.676,75.819,1795.709,190.657,1407.657,49.443,1572.330,141.177,1372.974,60.450,1745.736,184.757,1387.333,-,\textbf{1196.824},-
ETT2,H,12.144,0.175,12.504,0.193,11.804,0.247,\textcolor{blue}{11.670},0.032,12.273,0.112,12.340,0.174,12.036,0.596,12.372,0.081,\textbf{11.523},0.097,12.872,0.460,12.009,0.153,\underline{11.540},-,15.357,-
,D,\textbf{362.096},17.156,417.034,24.886,401.893,19.439,\textcolor{blue}{383.881},7.851,453.978,52.366,508.237,106.207,684.881,101.191,383.199,3.924,712.710,58.727,423.639,18.180,1112.936,235.365,\underline{377.405},-,407.549,-
,W,5006.471,654.009,\textcolor{blue}{3304.356},387.108,5901.160,3409.781,\textcolor{blue}{\underline{3212.982}},288.743,4303.516,846.771,3651.227,423.191,5021.404,827.990,3935.100,402.571,3837.760,542.886,3549.605,126.222,4455.781,240.462,3509.154,-,\textbf{2310.677},-
Solar,H,28.555,0.312,30.381,1.783,26.320,1.245,29.658,0.959,27.560,0.621,30.057,0.153,51.034,11.420,\textbf{24.513},0.424,\underline{24.551},0.290,26.816,0.759,28.542,0.713,26.398,-,47.962,-
,D,343.408,1.308,\textcolor{blue}{342.889},1.326,339.974,1.854,\textcolor{blue}{334.769},1.371,355.707,2.003,386.462,1.913,370.623,17.872,341.111,1.474,386.142,6.153,413.068,18.782,343.042,3.532,\underline{324.109},-,\textbf{319.640},-
,W,1223.832,21.946,\textcolor{blue}{1196.993},75.591,1414.840,22.908,\textcolor{blue}{1325.378},50.085,1195.734,57.949,1080.363,39.955,1258.900,40.770,\textbf{1038.434},34.269,1048.286,103.252,1543.885,230.390,\underline{1040.493},25.294,1599.687,-,1207.444,-
\end{filecontents*}
\csvreader[before reading=\gdef\prevdatasettwo{},late after line=\\,late after last line=\vspace{3pt}\\\bottomrule]{tables/rmse_main.csv}
{Dataset=\dataset,
Frequency=\freq,
moment_u=\momentU,
moment_u_se=\momentUSE,
moment_m=\momentM,
moment_m_se=\momentMSE,
patchtst_u=\patchtstU,
patchtst_u_se=\patchtstUSE,
patchtst_m=\patchtstM,
patchtst_m_se=\patchtstMSE,
itransformer_m=\itransformerM,
itransformer_m_se=\itransformerMSE,
itransformer_t5_m=\itransformerTfiveM,
itransformer_t5_m_se=\itransformerTfiveMSE,
crossformer_m=\crossformerM,
crossformer_m_se=\crossformerMSE,
chronos_m=\chronosM,
chronos_m_se=\chronosMSE,
timerxl_m=\timerxlM,
timerxl_m_se=\timerxlMSE,
tsmixer_m=\tsmixerM,
tsmixer_m_se=\tsmixerMSE,
timemixer_m=\timemixerM,
timemixer_m_se=\timemixerMSE,
mlp_m=\mlpM,
mlp_m_se=\mlpMSE,
autoets=\autoets,
autoets_se=\autoetsSE
}
{%
\IfStrEq{\dataset}{}{}{%
  \IfStrEq{\prevdatasettwo}{}{}{\\\addlinespace[-3pt]\hline\addlinespace[6pt]}%
  \gdef\prevdatasettwo{\dataset}%
}%
\multirow{2}{*}{\dataset} & \multirow{2}{*}{\freq} & \momentU & \momentM & \patchtstU & \patchtstM & \itransformerM & \itransformerTfiveM & \crossformerM & \timerxlM & \tsmixerM & \timemixerM & \mlpM & \chronosM & \autoets  \\
{} & {} & \SE{\momentUSE} & \SE{\momentMSE} & \SE{\patchtstUSE} & \SE{\patchtstMSE} & \SE{\itransformerMSE} & \SE{\itransformerTfiveMSE} & \SE{\crossformerMSE} & \SE{\timerxlMSE} & \SE{\tsmixerMSE} & \SE{\timemixerMSE} & \SE{\mlpMSE} & -- & -- }
\multicolumn{2}{l|}{\textbf{Average Rank}} & 5.833 & \underline{5.389} & 6.611 & \textbf{3.444} & 7.944 & 6.778 & 8.111 & 5.833 & 6.500 & 7.389 & 8.722 & 5.444 & -- \\
\bottomrule
\end{tabular}}
\label{tab:rmse_main}
\end{table*}

In addition to MAE, we report Root Mean Squared Error (RMSE) across all experiments, defined as

\begin{equation}
    \text{RMSE}(\mathbf{Y}, \mathbf{\hat{Y}}) = \sqrt{\frac{1}{B C H} \sum_{b=1}^{B} \sum_{c=1}^{C} \sum_{h=1}^{H} \left(\mathbf{Y}_{b, c, t+h} - \mathbf{\hat{Y}}_{b, c, t+h}\right)^2}
\end{equation}
where $\mathbf{Y}$ and $\mathbf{\hat{Y}}$ are the ground truth and predicted values over batch $B$, channels $C$, and horizon $H$, respectively.

\newpage
\subsection{MICA Comparison with Univariate Adapters}
\begin{table*}[ht!]
\newcommand{\prevdatasetthree}{}
\centering
\caption{Comparison of channel-independent models (Baseline), channel-independent models with a multivariate output layer (MOL), channel-independent models with \PCA\ preprocessing, and channel-dependent models with \MICA\ (MLP-Query gate). Forecasting MAE averaged over 5 random seeds with standard deviation in parentheses. Best results are shown in \textbf{bold}. Second-best results are \underline{underlined}.}
\scriptsize
\resizebox{1.0\textwidth}{!}{
\begin{tabular}{ll|cccc|cccc}
\toprule
\multirow{2}{*}{\textbf{Dataset}} & \multirow{2}{*}{\textbf{Freq.}} & \multicolumn{4}{c|}{\textbf{MOMENT}} & \multicolumn{4}{c}{\textbf{PatchTST}} \\
{} & {} & Baseline & Baseline w/ MOL & Baseline w/ PCA & \MICA & Baseline & Baseline w/ MOL & Baseline w/ PCA & \MICA \\
\midrule
\addlinespace[6pt]
\begin{filecontents*}{tables/pca_ablation.csv}
Dataset,Frequency,moment_vanilla,moment_vanilla_se,moment_pca,moment_pca_se,moment_multivariatehead,moment_multivariatehead_se,moment_mica,moment_mica_se,patchtst_vanilla,patchtst_vanilla_se,patchtst_pca,patchtst_pca_se,patchtst_multivariatehead,patchtst_multivariatehead_se,patchtst_mica,patchtst_mica_se
Simglucose,5min,5.136,0.057,5.991,0.024,5.114,0.074,\underline{4.347},0.035,5.593,0.577,6.000,0.181,5.461,0.477,\textbf{4.241},0.097
Iowa IHOP SMEX02,5min,1.733,0.005,1.688,0.001,1.725,0.004,\underline{1.666},0.005,1.765,0.019,1.737,0.024,1.735,0.013,\textbf{1.662},0.008
Iowa PLOWS,5min,1.369,0.002,1.354,0.002,1.370,0.002,\underline{1.332},0.005,1.382,0.009,1.379,0.018,1.374,0.008,\textbf{1.327},0.006
M-DENSE,H,92.637,0.916,104.366,1.590,93.697,1.335,\textbf{87.412},1.880,95.861,2.105,113.495,3.970,93.474,0.539,\underline{88.020},1.220
,D,52.927,0.849,52.486,1.082,53.651,0.228,\textbf{51.740},0.413,53.723,0.236,64.278,0.612,54.864,1.503,\underline{51.959},0.203
Jena Weather,H,9.682,0.131,19.047,0.194,9.659,0.110,\textbf{9.387},0.097,9.911,0.225,14.246,1.991,10.033,0.109,\underline{9.543},0.298
,D,\textbf{13.155},0.438,15.568,0.405,\underline{13.439},0.594,14.907,0.271,14.005,0.185,16.747,0.623,13.468,0.254,13.799,0.328
ETT1,H,5.683,0.055,5.795,0.144,5.528,0.106,5.851,0.132,\underline{5.454},0.035,5.650,0.131,5.475,0.110,\textbf{5.403},0.199
,D,157.189,2.469,162.339,1.953,158.716,3.713,157.090,1.833,\underline{155.534},2.508,161.843,2.345,\textbf{154.765},1.685,157.871,2.663
,W,\underline{982.317},37.048,1108.784,22.397,1012.044,37.694,994.716,34.867,1003.563,29.109,1036.403,72.376,1064.893,55.827,\textbf{959.206},42.454
ETT2,H,7.618,0.071,8.016,0.115,7.489,0.080,7.720,0.100,7.452,0.127,7.392,0.118,\underline{7.339},0.097,\textbf{7.334},0.013
,D,\textbf{228.246},11.983,284.723,5.515,\underline{235.198},6.864,269.353,16.031,260.281,11.639,286.482,7.231,311.022,42.378,254.152,5.445
,W,2868.244,359.457,2526.108,103.823,2980.910,308.796,\textbf{2211.349},203.851,3016.452,686.651,2853.026,260.526,2835.500,370.455,\underline{2249.077},188.625
\end{filecontents*}
\csvreader[before reading=\gdef\prevdatasetthree{},late after line=\\,late after last line=\vspace{3pt}\\\bottomrule]{tables/pca_ablation.csv}
{Dataset=\dataset,
Frequency=\freq,
moment_vanilla=\momentVanilla,
moment_vanilla_se=\momentVanillaSE,
moment_mica=\momentMica,
moment_mica_se=\momentMicaSE,
moment_pca=\momentPca,
moment_pca_se=\momentPcaSE,
moment_multivariatehead=\momentMH,
moment_multivariatehead_se=\momentMHSE,
patchtst_vanilla=\patchtstVanilla,
patchtst_vanilla_se=\patchtstVanillaSE,
patchtst_mica=\patchtstMica,
patchtst_mica_se=\patchtstMicaSE,
patchtst_pca=\patchtstPca,
patchtst_pca_se=\patchtstPcaSE,
patchtst_multivariatehead=\patchtstMH,
patchtst_multivariatehead_se=\patchtstMHSE
}
{%
\IfStrEq{\dataset}{}{}{%
  \IfStrEq{\prevdatasetthree}{}{}{\\\addlinespace[-3pt]\hline\addlinespace[6pt]}%
  \gdef\prevdatasetthree{\dataset}%
}%
\multirow{2}{*}{\dataset} & \multirow{2}{*}{\freq} & \momentVanilla & \momentPca & \momentMH & \momentMica & \patchtstVanilla & \patchtstPca & \patchtstMH & \patchtstMica \\
{} & {} & \SE{\momentVanillaSE} & \SE{\momentPcaSE} & \SE{\momentMHSE}& \SE{\momentMicaSE}  & \SE{\patchtstVanillaSE} & \SE{\patchtstPcaSE} & \SE{\patchtstMHSE} & \SE{\patchtstMicaSE} 
}
\multicolumn{2}{l|}{\textbf{Average Rank}} & 3.846 & 4.308 & 6.000 & \underline{3.231} & 5.231 & 4.769 & 6.615 & \textbf{2.000} \\
\bottomrule
\end{tabular}}
\label{tab:mica_vs_pca}
\end{table*}

\newpage
\subsection{MICA Channel Exclusion Ablation}\label{apd:mica_ciexc_ablation}
\begin{table*}[ht!]
\newcommand{\prevdatasetfour}{}
\centering
\caption{\Moment~MAE values averaged over 5 random seeds with standard deviation in parentheses. Channel inclusion (incl.) and exclusion (excl.) results are shown for each mixing gate variant. Best results are shown in \textbf{bold}. Second best results are \underline{underlined}.}
\scriptsize
\resizebox{0.85\textwidth}{!}{
\begin{tabular}{ll|cc|cc|cc|cc|cc|cc}
\toprule
\multirow{2}{*}{\textbf{Dataset}} & \multirow{2}{*}{\textbf{Freq.}} & \multicolumn{2}{c|}{Shared $\beta$} & \multicolumn{2}{c|}{Channelwise $\beta$} & \multicolumn{2}{c|}{Layerwise $\beta$} & \multicolumn{2}{c|}{Layerwise Channelwise $\beta$} & \multicolumn{2}{c|}{\MLP} & \multicolumn{2}{c}{\MLP-Query} \\
{} & {} & incl. & excl. & incl. & excl. & incl. & excl. & incl. & excl. & incl. & excl. & incl. & excl. \\
\midrule
\addlinespace[6pt]
\begin{filecontents*}{tables/mae_moment_infini_ablation.csv}
Dataset,Frequency,shared_incl,shared_incl_se,shared_excl,shared_excl_se,channelwise_incl,channelwise_incl_se,channelwise_excl,channelwise_excl_se,layerwise_incl,layerwise_incl_se,layerwise_excl,layerwise_excl_se,layerwise_channelwise_incl,layerwise_channelwise_incl_se,layerwise_channelwise_excl,layerwise_channelwise_excl_se,mlp_incl,mlp_incl_se,mlp_excl,mlp_excl_se,mlpquery_incl,mlpquery_incl_se,mlpquery_excl,mlpquery_excl_se
Simglucose,5min,4.450,0.061,4.439,0.018,4.480,0.088,4.459,0.065,4.437,0.050,4.409,0.056,4.427,0.080,4.421,0.069,\underline{4.349},0.026,4.367,0.013,\textbf{4.347},0.035,4.360,0.025
Iowa IHOP SMEX02,5min,1.660,0.008,1.659,0.004,1.656,0.006,\textbf{1.654},0.005,1.663,0.005,1.660,0.009,\underline{1.654},0.005,1.656,0.007,1.659,0.009,1.661,0.003,1.666,0.005,1.664,0.002
Iowa PLOWS,5min,1.329,0.003,1.330,0.004,\textbf{1.324},0.002,\underline{1.326},0.003,1.328,0.003,1.328,0.002,1.327,0.003,1.326,0.003,1.335,0.006,1.335,0.002,1.332,0.005,1.334,0.006
M-DENSE,H,90.566,2.289,90.275,1.503,90.798,1.036,90.935,1.691,89.908,2.602,90.311,1.392,89.595,1.854,89.614,1.074,\textbf{86.631},1.577,87.847,0.756,87.412,1.880,\underline{87.400},1.181
,D,53.763,0.891,53.539,0.589,53.605,0.628,52.896,1.205,53.110,0.896,52.947,0.380,53.173,0.727,52.986,0.658,52.691,1.136,53.192,1.324,\textbf{51.740},0.413,\underline{51.752},0.648
Jena Weather,H,9.357,0.129,9.407,0.068,9.391,0.238,9.335,0.061,9.333,0.067,9.308,0.173,\textbf{9.271},0.115,\underline{9.282},0.194,9.583,0.108,9.507,0.128,9.387,0.097,9.371,0.227
,D,\underline{13.739},0.320,13.914,0.347,13.788,0.253,13.830,0.391,\textbf{13.648},0.441,13.781,0.346,13.798,0.486,13.968,0.281,14.821,0.628,15.161,1.146,14.907,0.271,14.947,0.370
ETT1,H,\underline{5.759},0.139,5.835,0.101,\textbf{5.746},0.117,5.769,0.104,5.808,0.194,5.782,0.152,5.800,0.093,5.837,0.099,5.892,0.121,5.821,0.103,5.851,0.132,5.801,0.102
,D,155.723,3.906,156.782,2.461,156.327,1.587,155.456,2.188,154.678,1.299,154.107,4.226,\underline{154.088},1.352,\textbf{152.694},2.036,155.643,4.374,154.289,2.145,157.090,1.833,156.998,2.109
,W,993.061,33.693,979.715,27.245,976.330,16.314,978.413,31.396,972.070,34.004,\textbf{968.032},26.293,996.186,38.992,1012.946,35.419,\underline{971.038},20.054,984.146,41.702,994.716,34.867,987.018,28.277
ETT2,H,7.637,0.074,7.779,0.085,\textbf{7.579},0.083,\underline{7.637},0.166,7.662,0.066,7.678,0.117,7.642,0.046,7.655,0.070,7.702,0.159,7.721,0.109,7.720,0.100,7.677,0.102
,D,\textbf{255.603},7.739,274.739,6.838,259.421,5.664,\underline{259.211},13.868,261.925,7.355,259.887,10.799,259.361,13.834,268.043,6.749,279.437,13.377,285.798,20.996,269.353,16.031,279.872,17.638
,W,\textbf{2044.221},182.469,2109.747,104.115,\underline{2051.375},109.040,2122.875,82.339,2364.977,265.811,2081.167,124.617,2181.748,250.132,2257.975,70.381,2241.918,260.832,2192.253,178.628,2211.349,203.851,2192.882,121.789
\end{filecontents*}
\csvreader[before reading=\gdef\prevdatasetfour{},late after line=\\,late after last line=\vspace{3pt}\\\bottomrule]{tables/mae_moment_infini_ablation.csv}
{Dataset=\dataset,
Frequency=\freq,
shared_incl=\sharedIncl,
shared_incl_se=\sharedInclSE,
shared_excl=\sharedExcl,
shared_excl_se=\sharedExclSE,
channelwise_incl=\channelwiseIncl,
channelwise_incl_se=\channelwiseInclSE,
channelwise_excl=\channelwiseExcl,
channelwise_excl_se=\channelwiseExclSE,
layerwise_incl=\layerwiseIncl,
layerwise_incl_se=\layerwiseInclSE,
layerwise_excl=\layerwiseExcl,
layerwise_excl_se=\layerwiseExclSE,
layerwise_channelwise_incl=\layerwisechannelwiseIncl,
layerwise_channelwise_incl_se=\layerwisechannelwiseInclSE,
layerwise_channelwise_excl=\layerwisechannelwiseExcl,
layerwise_channelwise_excl_se=\layerwisechannelwiseExclSE,
mlp_incl=\mlpIncl,
mlp_incl_se=\mlpInclSE,
mlp_excl=\mlpExcl,
mlp_excl_se=\mlpExclSE,
mlpquery_incl=\mlpqueryIncl,
mlpquery_incl_se=\mlpqueryInclSE,
mlpquery_excl=\mlpqueryExcl,
mlpquery_excl_se=\mlpqueryExclSE
}
{%
\IfStrEq{\dataset}{}{}{%
  \IfStrEq{\prevdatasetfour}{}{}{\\\addlinespace[-3pt]\hline\addlinespace[6pt]}%
  \gdef\prevdatasetfour{\dataset}%
}%
\multirow{2}{*}{\dataset} & \multirow{2}{*}{\freq} & \sharedIncl & \sharedExcl & \channelwiseIncl & \channelwiseExcl & \layerwiseIncl & \layerwiseExcl & \layerwisechannelwiseIncl & \layerwisechannelwiseExcl & \mlpIncl & \mlpExcl & \mlpqueryIncl & \mlpqueryExcl \\
{} & {} & \SE{\sharedInclSE} & \SE{\sharedExclSE} & \SE{\channelwiseInclSE} & \SE{\channelwiseExclSE} & \SE{\layerwiseInclSE} & \SE{\layerwiseExclSE} & \SE{\layerwisechannelwiseInclSE} & \SE{\layerwisechannelwiseExclSE} & \SE{\mlpInclSE} & \SE{\mlpExclSE} & \SE{\mlpqueryInclSE} & \SE{\mlpqueryExclSE}
}
\end{tabular}}
\label{tab:mae_moment_infini_ablation}
\end{table*}

\begin{figure}[ht!]
    \centering
    \includegraphics[width=0.7\textwidth, trim=0 60 0 0, clip]{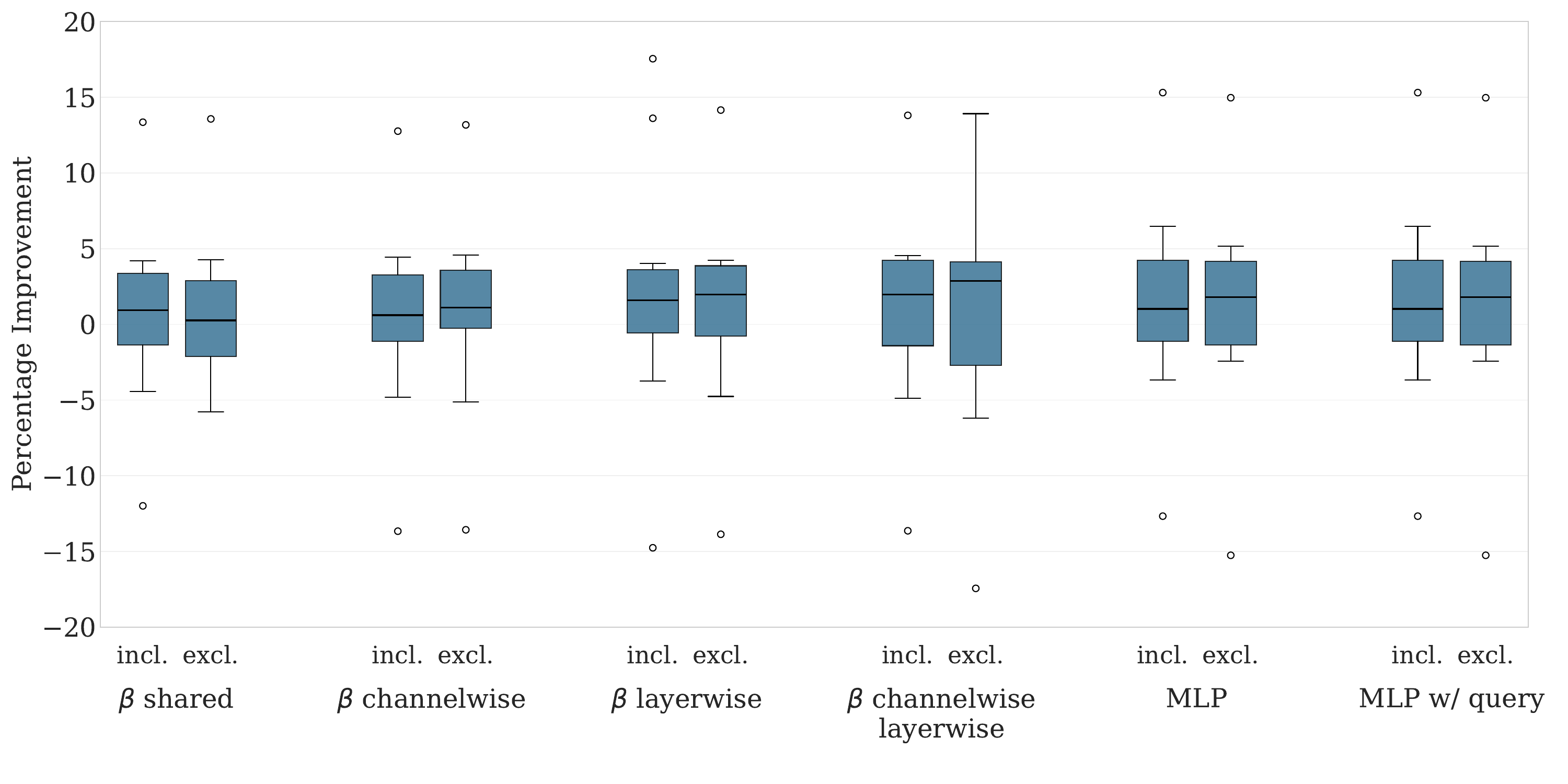}
    \caption{Distribution of percentage error reduction across datasets when augmenting \Moment\ with different \MICA\ gate variants, comparing channel inclusion (incl.) and channel exclusion (excl.) approaches relative to the original univariate implementation. Channel inclusion variants achieve comparable MAE reductions to channel exclusion variants while requiring less computation.}
    \label{fig:boxplot_infini_varient_ablation_moment}
\end{figure}

\begin{figure}[ht!]
    \centering
    \includegraphics[width=0.7\textwidth, trim=0 5 0 0, clip]{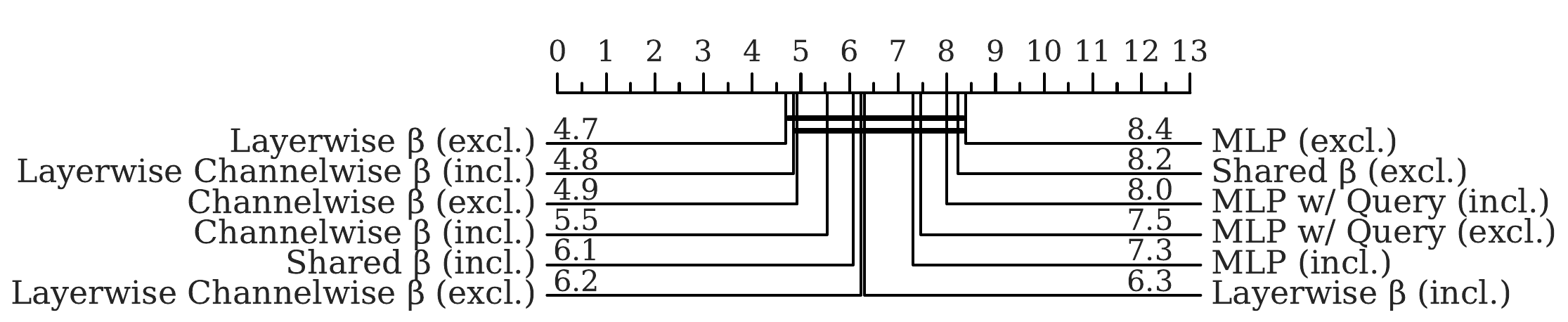}
    \caption{Critical Difference diagrams based on Friedman test ($\alpha = 0.2$) with post-hoc Wilcoxon signed-rank tests (Holm-corrected). Horizontal bars connect methods with no statistically significant performance differences. Lower ranks indicate better performance. For \Moment, $\beta$-based gates achieve lower forecast error across datasets (top ranks 4.7-6.2).}
    \label{fig:mica_variant_cd_moment}
\end{figure}

\newpage
\begin{table*}[ht!]
\newcommand{\prevdatasetfive}{}
\centering
\caption{\PatchTST~MAE values averaged over 5 random seeds with standard deviation in parentheses. Channel inclusion (incl.) and exclusion (excl.) results are shown for each mixing gate variant. Best results are shown in \textbf{bold}. Second best results are \underline{underlined}.}
\scriptsize
\resizebox{0.85\textwidth}{!}{
\begin{tabular}{ll|cc|cc|cc|cc|cc|cc}
\toprule
\multirow{2}{*}{\textbf{Dataset}} & \multirow{2}{*}{\textbf{Freq.}} & \multicolumn{2}{c|}{Shared $\beta$} & \multicolumn{2}{c|}{Channelwise $\beta$} & \multicolumn{2}{c|}{Layerwise $\beta$} & \multicolumn{2}{c|}{Layerwise Channelwise $\beta$} & \multicolumn{2}{c|}{\MLP} & \multicolumn{2}{c}{\MLP-Query} \\
{} & {} & incl. & excl. & incl. & excl. & incl. & excl. & incl. & excl. & incl. & excl. & incl. & excl. \\
\midrule
\addlinespace[6pt]
\begin{filecontents*}{tables/mae_patchtst_infini_ablation.csv}
Dataset,Frequency,shared_incl,shared_incl_se,shared_excl,shared_excl_se,channelwise_incl,channelwise_incl_se,channelwise_excl,channelwise_excl_se,layerwise_incl,layerwise_incl_se,layerwise_excl,layerwise_excl_se,layerwise_channelwise_incl,layerwise_channelwise_incl_se,layerwise_channelwise_excl,layerwise_channelwise_excl_se,mlp_incl,mlp_incl_se,mlp_excl,mlp_excl_se,mlpquery_incl,mlpquery_incl_se,mlpquery_excl,mlpquery_excl_se
Simglucose,5min,4.621,0.270,4.665,0.353,4.515,0.324,4.484,0.339,4.502,0.143,4.495,0.322,4.447,0.188,4.574,0.395,4.383,0.193,\underline{4.308},0.056,\textbf{4.241},0.097,4.334,0.131
Iowa IHOP SMEX02,5min,1.708,0.010,1.703,0.010,1.676,0.012,1.674,0.012,1.707,0.018,1.713,0.013,1.672,0.008,1.670,0.006,1.663,0.001,1.662,0.005,\underline{1.662},0.008,\textbf{1.662},0.003
Iowa PLOWS,5min,1.346,0.005,1.345,0.010,1.332,0.005,1.332,0.003,1.339,0.004,1.340,0.009,1.348,0.017,1.334,0.007,1.331,0.007,1.332,0.006,\textbf{1.327},0.006,\underline{1.330},0.004
M-DENSE,H,90.092,1.925,\underline{87.559},1.861,88.762,2.462,89.122,1.644,87.797,1.126,88.380,1.170,87.875,1.728,88.052,1.574,87.916,1.963,88.410,1.370,88.020,1.220,\textbf{87.089},1.499
,D,53.236,0.376,53.286,0.242,53.654,0.451,53.658,0.336,53.646,0.187,53.473,0.345,53.585,0.310,53.744,0.468,52.287,0.380,52.506,0.638,\underline{51.959},0.203,\textbf{51.594},0.598
Jena Weather,H,10.039,0.161,10.021,0.157,9.918,0.145,9.937,0.108,9.894,0.106,9.932,0.069,9.984,0.192,10.003,0.103,9.762,0.398,9.708,0.358,\underline{9.543},0.298,\textbf{9.469},0.211
,D,14.205,0.310,14.115,0.273,14.289,0.370,14.343,0.603,14.203,0.284,14.232,0.375,13.938,0.371,14.043,0.244,\underline{13.770},0.225,13.858,0.148,13.799,0.328,\textbf{13.752},0.061
ETT1,H,5.767,0.081,5.696,0.179,5.703,0.135,5.628,0.071,5.774,0.145,5.763,0.186,5.699,0.186,5.713,0.070,5.551,0.152,5.503,0.140,\textbf{5.403},0.199,\underline{5.436},0.093
,D,156.026,1.922,157.169,1.618,155.099,1.273,157.402,0.802,157.989,0.814,156.879,2.640,\textbf{154.775},1.697,158.056,3.302,157.203,0.771,\underline{154.993},1.370,157.871,2.663,158.141,1.355
,W,999.524,68.341,987.581,39.631,993.531,28.579,994.968,26.592,990.500,34.715,982.253,25.797,970.395,22.805,\textbf{935.564},17.231,958.801,23.069,985.183,36.963,959.206,42.454,\underline{944.441},28.565
ETT2,H,8.205,0.157,8.237,0.147,8.153,0.199,8.119,0.054,8.118,0.232,8.007,0.221,8.158,0.081,8.035,0.372,7.528,0.239,7.660,0.316,\textbf{7.334},0.013,\underline{7.473},0.366
,D,247.162,7.129,248.721,8.163,\underline{241.872},9.095,251.775,8.836,\textbf{241.521},4.358,252.765,12.941,247.694,7.569,248.600,7.576,256.846,5.057,264.841,6.113,254.152,5.445,269.223,11.652
,W,2798.743,230.981,2545.244,189.036,2488.112,224.703,2489.179,271.921,2618.755,189.615,2554.740,283.385,2578.481,207.308,2728.587,374.751,2465.600,128.406,2453.902,312.098,\underline{2249.077},188.625,\textbf{2169.814},112.729
\end{filecontents*}
\csvreader[before reading=\gdef\prevdatasetfive{},late after line=\\,late after last line=\vspace{3pt}\\\bottomrule]{tables/mae_patchtst_infini_ablation.csv}
{Dataset=\dataset,
Frequency=\freq,
shared_incl=\sharedIncl,
shared_incl_se=\sharedInclSE,
shared_excl=\sharedExcl,
shared_excl_se=\sharedExclSE,
channelwise_incl=\channelwiseIncl,
channelwise_incl_se=\channelwiseInclSE,
channelwise_excl=\channelwiseExcl,
channelwise_excl_se=\channelwiseExclSE,
layerwise_incl=\layerwiseIncl,
layerwise_incl_se=\layerwiseInclSE,
layerwise_excl=\layerwiseExcl,
layerwise_excl_se=\layerwiseExclSE,
layerwise_channelwise_incl=\layerwisechannelwiseIncl,
layerwise_channelwise_incl_se=\layerwisechannelwiseInclSE,
layerwise_channelwise_excl=\layerwisechannelwiseExcl,
layerwise_channelwise_excl_se=\layerwisechannelwiseExclSE,
mlp_incl=\mlpIncl,
mlp_incl_se=\mlpInclSE,
mlp_excl=\mlpExcl,
mlp_excl_se=\mlpExclSE,
mlpquery_incl=\mlpqueryIncl,
mlpquery_incl_se=\mlpqueryInclSE,
mlpquery_excl=\mlpqueryExcl,
mlpquery_excl_se=\mlpqueryExclSE
}
{%
\IfStrEq{\dataset}{}{}{%
  \IfStrEq{\prevdatasetfive}{}{}{\\\addlinespace[-3pt]\hline\addlinespace[6pt]}%
  \gdef\prevdatasetfive{\dataset}%
}%
\multirow{2}{*}{\dataset} & \multirow{2}{*}{\freq} & \sharedIncl & \sharedExcl & \channelwiseIncl & \channelwiseExcl & \layerwiseIncl & \layerwiseExcl & \layerwisechannelwiseIncl & \layerwisechannelwiseExcl & \mlpIncl & \mlpExcl & \mlpqueryIncl & \mlpqueryExcl \\
{} & {} & \SE{\sharedInclSE} & \SE{\sharedExclSE} & \SE{\channelwiseInclSE} & \SE{\channelwiseExclSE} & \SE{\layerwiseInclSE} & \SE{\layerwiseExclSE} & \SE{\layerwisechannelwiseInclSE} & \SE{\layerwisechannelwiseExclSE} & \SE{\mlpInclSE} & \SE{\mlpExclSE} & \SE{\mlpqueryInclSE} & \SE{\mlpqueryExclSE}
}
\end{tabular}}
\label{tab:mae_patchtst_infini_ablation}
\end{table*}

\begin{figure}[ht!]
    \centering
    \includegraphics[width=0.7\textwidth, trim=0 60 0 0, clip]{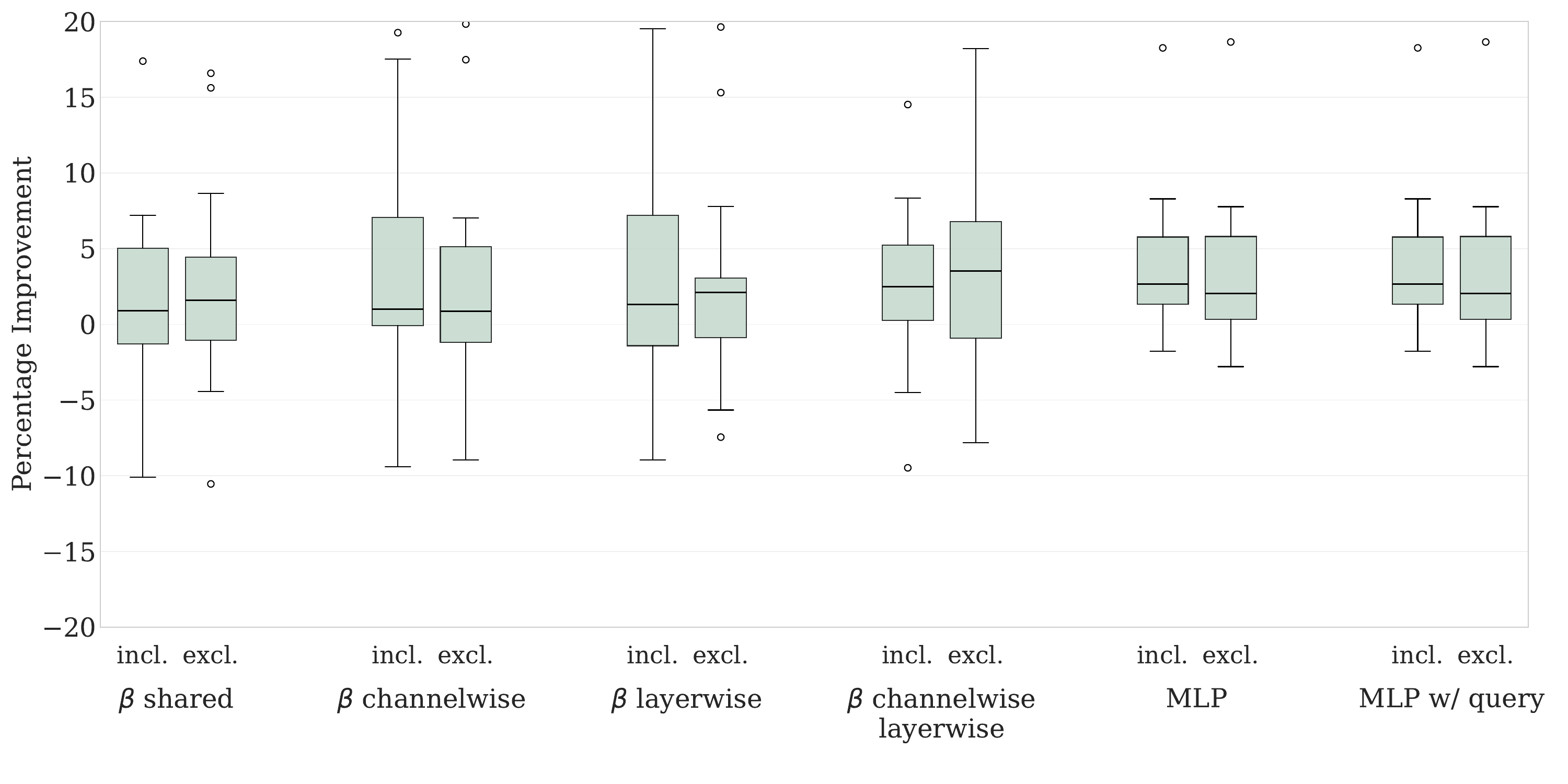}
    \caption{Distribution of percentage error reduction across datasets when augmenting \PatchTST\ with different \MICA\ gate variants, comparing channel inclusion (incl.) and channel exclusion (excl.) approaches relative to the original univariate implementation. Channel inclusion variants achieve comparable MAE reductions to channel exclusion variants while requiring less computation.}
    \label{fig:boxplot_infini_varient_ablation_patchtst}
\end{figure}

\begin{figure}[ht!]
    \centering
    \includegraphics[width=0.7\textwidth, trim=0 5 0 0, clip]{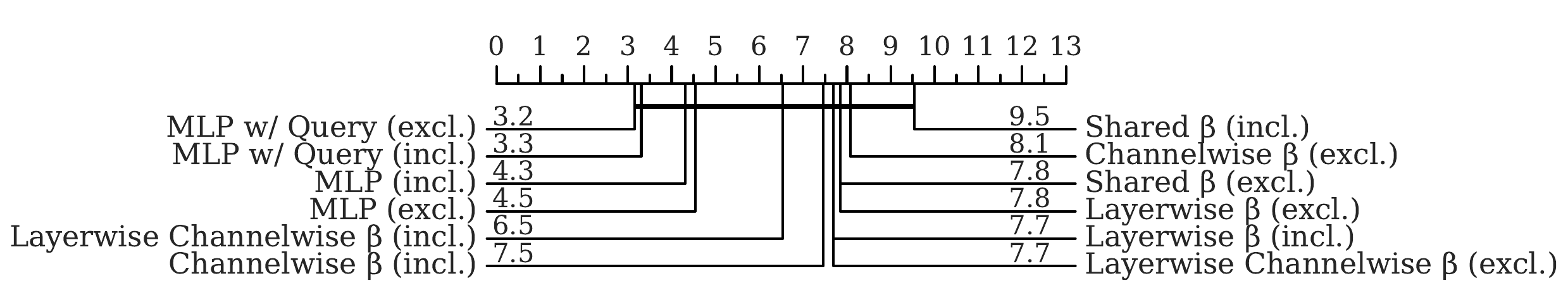}
    \caption{Critical Difference diagrams based on Friedman test ($\alpha = 0.2$) with post-hoc Wilcoxon signed-rank tests (Holm-corrected). Horizontal bars connect methods with no statistically significant performance differences. Lower ranks indicate better performance. For \PatchTST, \MLP-based gates achieve lower forecast error across datasets (top ranks 3.2-4.5).}
    \label{fig:mica_variant_cd_patchtst}
\end{figure}

\newpage
\subsection{MICA Weighted Channel Aggregation Ablation}\label{apd:mica_channel_weight_ablation}
\begin{table*}[ht!]
\newcommand{\prevdatasetwc}{}
\centering
\caption{Ablation of channel weighting mechanism in \MICA. We compare vanilla baselines against \MICA\ variants with uniform (U; default), static (S), and dynamic (D) weighted channels. Forecasting MAE averaged over 5 random seeds with standard deviation in parentheses. Best results are shown in \textbf{bold}. Second-best results are \underline{underlined}.}
\scriptsize
\resizebox{0.62\textwidth}{!}{
\begin{tabular}{ll|cccc}
\toprule
\multirow{2}{*}{\textbf{Dataset}} & \multirow{2}{*}{\textbf{Freq.}} & \multicolumn{4}{c}{\textbf{PatchTST}} \\
{} & {} & Baseline & \MICA\ (U) & \MICA\ (S) & \MICA\ (D) \\
\midrule
\begin{filecontents*}{tables/mae_channel_weight_table.csv}
Dataset,Frequency,patchtst_vanilla,patchtst_vanilla_se,patchtst_mica,patchtst_mica_se,patchtst_mica_sw,patchtst_mica_sw_se,patchtst_mica_dw,patchtst_mica_dw_se
Simglucose,5min,5.593,0.577,\textbf{4.241},0.097,\underline{4.268},0.141,4.302,0.172
COVID Deaths,D,141.437,43.419,\underline{135.718},40.678,\textbf{130.176},27.018,143.915,86.827
Iowa IHOP SMEX02,5min,1.765,0.019,\textbf{1.662},0.008,\underline{1.665},0.002,1.682,0.008
Iowa PLOWS,5min,1.382,0.009,\textbf{1.327},0.006,1.334,0.003,\underline{1.333},0.007
Jena Weather,H,9.911,0.225,9.543,0.298,\underline{9.517},0.176,\textbf{9.476},0.223
,D,14.005,0.185,\textbf{13.799},0.328,13.944,0.260,\underline{13.889},0.382
M-DENSE,H,95.861,2.105,88.020,1.220,\textbf{86.743},1.164,\underline{87.534},1.604
,D,53.723,0.236,\underline{51.959},0.203,52.078,0.302,\textbf{51.594},1.182
Loop-Seattle,D,3.246,0.011,\underline{3.009},0.003,3.021,0.012,\textbf{3.006},0.012
ETT1,H,\underline{5.454},0.035,\textbf{5.403},0.199,5.506,0.186,5.470,0.051
,D,\textbf{155.534},2.508,\underline{157.871},2.663,159.448,2.065,158.186,1.721
,W,1003.563,29.109,\underline{959.206},42.454,\textbf{938.360},22.515,965.903,27.555
ETT2,H,\underline{7.452},0.127,\textbf{7.334},0.013,7.574,0.072,7.662,0.325
,D,260.281,11.639,\underline{254.152},5.445,\textbf{253.327},5.868,255.058,9.701
,W,3016.452,686.651,\textbf{2249.077},188.625,2436.890,214.840,\underline{2324.161},157.982
Solar,H,\textbf{12.073},1.072,12.847,0.402,12.750,0.262,\underline{12.162},0.618
,D,258.389,1.663,\textbf{252.718},1.232,254.602,1.290,\underline{252.852},2.516
,W,1127.889,20.369,\textbf{1058.990},38.140,1067.190,25.051,\underline{1060.330},46.974
\end{filecontents*}
\csvreader[before reading=\gdef\prevdatasetwc{},late after line=\\,late after last line=\vspace{3pt}\\\bottomrule]{tables/mae_channel_weight_table.csv}
{Dataset=\dataset,
Frequency=\freq,
patchtst_vanilla=\patchtstVanilla,
patchtst_vanilla_se=\patchtstVanillaSE,
patchtst_mica=\patchtstMica,
patchtst_mica_se=\patchtstMicaSE,
patchtst_mica_sw=\patchtstMicaSW,
patchtst_mica_sw_se=\patchtstMicaSWSE,
patchtst_mica_dw=\patchtstMicaDW,
patchtst_mica_dw_se=\patchtstMicaDWSE
}
{%
\IfStrEq{\dataset}{}{}{%
  \IfStrEq{\prevdatasetwc}{}{}{\\\addlinespace[-3pt]\hline\addlinespace[6pt]}%
  \gdef\prevdatasetwc{\dataset}%
}%
\multirow{2}{*}{\dataset} & \multirow{2}{*}{\freq} & \patchtstVanilla & \patchtstMica & \patchtstMicaSW & \patchtstMicaDW \\
{} & {} & \SE{\patchtstVanillaSE} & \SE{\patchtstMicaSE} & \SE{\patchtstMicaSWSE} & \SE{\patchtstMicaDWSE}
}
\multicolumn{2}{l|}{\textbf{Average Rank}} & 3.389 & \textbf{1.722} & 2.500 & \underline{2.389} \\
\bottomrule
\end{tabular}}
\label{tab:mae_weighted_channels_ablation}
\end{table*}

\newpage
\section{Training Efficiency Analysis}
\label{sec:training_efficiency}
\subsection{Estimated Training Efficiency}
We estimate per-step training duration (in milliseconds) to compare the training efficiency of each forecasting model. All experiments were conducted on NVIDIA A100-SXM4-80GB GPUs (80GB memory). Each model is trained for 100 steps, and the average per-step duration is obtained by dividing total training time by 100. We repeat this for 5 random seeds and report the mean and standard deviation in Table~\ref{tab:runtimes_main}.

\newcommand{\prevdatasetsix}{}
\begin{table*}[ht!]
\centering
\caption{Training runtimes (in millseconds) averaged over 5 random seeds with standard deviation in parentheses.}
\scriptsize
\resizebox{1.0\textwidth}{!}{
\begin{tabular}{ll|cccccccccc}
\toprule
\textbf{Dataset} & \textbf{Freq.} & \textbf{\Moment} & \textbf{\Moment-\MICA} & \textbf{\PatchTST} & \textbf{\PatchTST-\MICA} & \textbf{iTransformer} & \textbf{\iTransformer-T5} & \textbf{\Crossformer} & \textbf{\TimerXL} & \textbf{\TSMixer} & \textbf{\MLP} \\
\midrule
\addlinespace[6pt]
\begin{filecontents*}{tables/runtime_table.csv}
Dataset,Frequency,moment_u,moment_u_se,moment_mlpquery_incl,moment_mlpquery_incl_se,patchtst_u,patchtst_u_se,patchtst_mlpquery_incl,patchtst_mlpquery_incl_se,itransformer_m,itransformer_m_se,itransformer_t5_m,itransformer_t5_m_se,crossformer_m,crossformer_m_se,timerxl_m,timerxl_m_se,tsmixer_m,tsmixer_m_se,mlp_m,mlp_m_se
Simglucose,5min,1542.253,748.493,751.172,344.318,645.576,329.095,1266.716,615.015,40.361,9.060,38.964,9.029,113.615,7.357,48.406,10.467,29.314,2.814,22.262,1.810
COVID Deaths,D,261.874,1.793,427.477,3.592,223.357,1.746,308.873,3.114,41.437,0.869,60.300,0.766,979.454,1.670,1052.689,6.065,23.094,1.848,19.102,2.199
Iowa IHOP SMEX02,5min,75.364,7.279,132.074,20.070,108.759,58.686,633.883,285.402,46.972,8.870,36.730,1.717,186.547,5.420,71.946,10.039,37.408,6.924,28.786,5.236
Iowa PLOWS,5min,86.457,2.500,765.361,317.321,73.250,1.347,795.642,354.137,53.417,5.649,57.782,9.143,246.230,4.351,100.323,5.496,48.942,6.345,34.529,3.261
Jena Weather,H,54.739,0.793,72.742,2.074,35.343,0.319,46.337,1.187,26.475,5.562,22.744,2.378,162.135,0.645,57.224,4.397,17.210,0.418,13.582,1.237
,D,38.820,0.529,53.322,0.974,24.649,0.362,32.285,0.175,21.158,0.721,21.627,2.106,110.091,1.067,35.010,2.309,16.681,1.174,11.327,0.243
M-DENSE,H,71.270,4.730,100.865,6.792,56.525,7.070,82.435,8.605,32.682,11.534,26.603,1.722,224.483,5.901,83.418,5.859,25.396,6.256,20.500,4.229
,D,48.342,3.679,66.967,0.888,31.548,0.323,42.251,0.777,22.282,2.856,21.108,0.732,145.199,7.676,41.671,2.149,16.606,0.885,12.287,0.840
Loop-Seattle,D,313.727,3.394,513.692,3.511,269.124,2.160,371.626,3.418,48.593,1.572,69.562,1.344,1176.349,6.049,1507.540,8.858,23.857,0.712,18.925,0.400
ETT1,H,36.385,7.654,45.730,8.251,21.018,1.577,35.806,7.644,28.645,6.025,22.573,0.809,91.920,4.195,31.400,1.621,20.179,5.863,19.673,6.026
,D,26.145,1.728,34.813,5.147,18.383,0.428,27.022,1.604,21.916,1.528,20.004,0.771,83.392,0.822,35.143,8.108,16.914,2.618,13.803,3.961
,W,25.714,1.589,35.166,5.588,18.993,0.597,24.832,1.491,21.667,1.462,19.728,0.502,82.722,1.997,31.490,4.939,16.046,0.754,12.288,2.375
ETT2,H,29.965,0.778,44.032,9.915,28.813,8.591,30.741,6.827,23.703,1.684,21.678,0.787,91.309,3.106,36.337,8.126,21.725,4.183,15.777,3.318
,D,26.084,1.079,34.372,5.105,18.846,0.784,26.832,3.013,22.523,2.541,19.391,0.498,83.854,0.783,33.741,5.279,15.517,0.219,12.475,2.492
,W,24.819,0.790,32.999,1.890,19.117,0.540,25.309,2.064,24.410,3.498,19.937,0.312,84.468,2.337,29.109,2.378,15.967,0.442,11.345,0.554
Solar,H,234.600,2.046,352.187,3.822,188.787,1.222,251.369,2.749,34.981,2.193,41.849,0.398,833.532,4.484,766.430,3.812,27.604,3.393,22.212,0.301
,D,143.557,2.988,228.856,0.785,118.515,0.458,164.896,1.467,31.141,4.894,36.555,1.249,518.066,3.279,324.011,7.545,19.302,1.245,15.359,1.241
,W,72.472,0.545,133.396,1.165,60.431,0.576,90.043,0.714,27.724,2.037,36.126,1.143,225.311,1.371,86.765,4.434,18.413,0.454,13.858,0.381
\end{filecontents*}
\csvreader[before reading=\gdef\prevdatasetsix{},late after line=\\,late after last line=\vspace{3pt}\\\bottomrule]{tables/runtime_table.csv}
{Dataset=\dataset,
Frequency=\freq,
moment_u=\momentU,
moment_u_se=\momentUSE,
moment_mlpquery_incl=\momentM,
moment_mlpquery_incl_se=\momentMSE,
patchtst_u=\patchtstU,
patchtst_u_se=\patchtstUSE,
patchtst_mlpquery_incl=\patchtstM,
patchtst_mlpquery_incl_se=\patchtstMSE,
itransformer_m=\itransformerM,
itransformer_m_se=\itransformerMSE,
itransformer_t5_m=\itransformerTfiveM,
itransformer_t5_m_se=\itransformerTfiveMSE,
crossformer_m=\crossformerM,
crossformer_m_se=\crossformerMSE,
timerxl_m=\timerxlM,
timerxl_m_se=\timerxlMSE,
tsmixer_m=\tsmixerM,
tsmixer_m_se=\tsmixerMSE,
mlp_m=\mlpM,
mlp_m_se=\mlpMSE,
}
{%
\IfStrEq{\dataset}{}{}{%
  \IfStrEq{\prevdatasetsix}{}{}{\\\addlinespace[-3pt]\hline\addlinespace[6pt]}%
  \gdef\prevdatasetsix{\dataset}%
}%
\multirow{2}{*}{\dataset} & \multirow{2}{*}{\freq} & \momentU & \momentM & \patchtstU & \patchtstM & \itransformerM & \itransformerTfiveM & \crossformerM & \timerxlM & \tsmixerM & \mlpM \\
{} & {} & \SE{\momentUSE} & \SE{\momentMSE} & \SE{\momentUSE} & \SE{\patchtstUSE} & \SE{\itransformerMSE} & \SE{\itransformerTfiveMSE} & \SE{\crossformerMSE} & \SE{\timerxlMSE} & \SE{\tsmixerMSE} & \SE{\mlpMSE} 
}
\end{tabular}}
\label{tab:runtimes_main}
\end{table*}

\newpage
\subsection{Training Error Trajectories}

\begin{figure}[H]
    \centering
    \begin{minipage}[t]{0.38\textwidth}
        \centering
        \includegraphics[width=\textwidth, trim=0 15 0 0, clip]{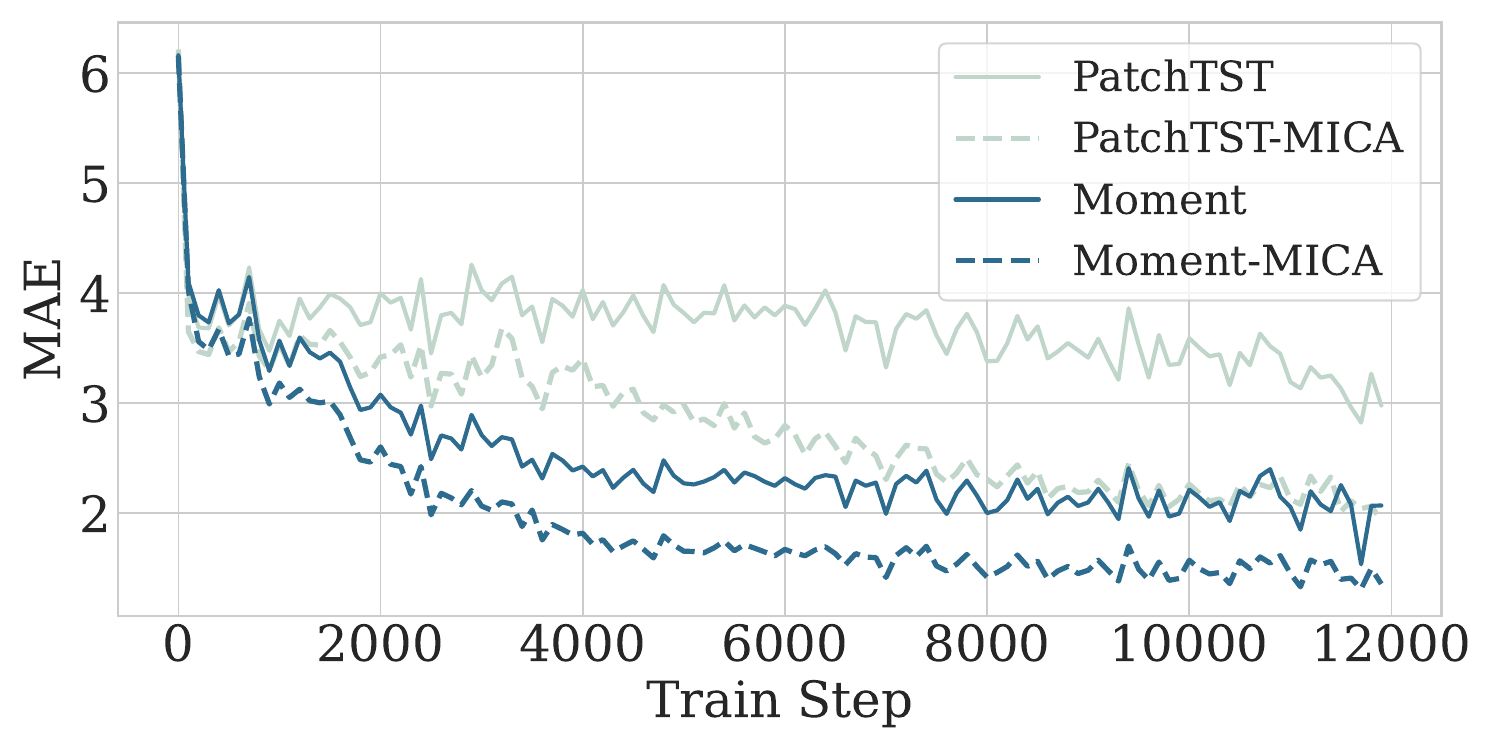}
        \par\vspace{2pt}
        {\small (a) COVID Deaths - Train set}
    \end{minipage}
    \hspace{2em}
    \begin{minipage}[t]{0.38\textwidth}
        \centering
        \includegraphics[width=\textwidth, trim=0 15 0 0, clip]{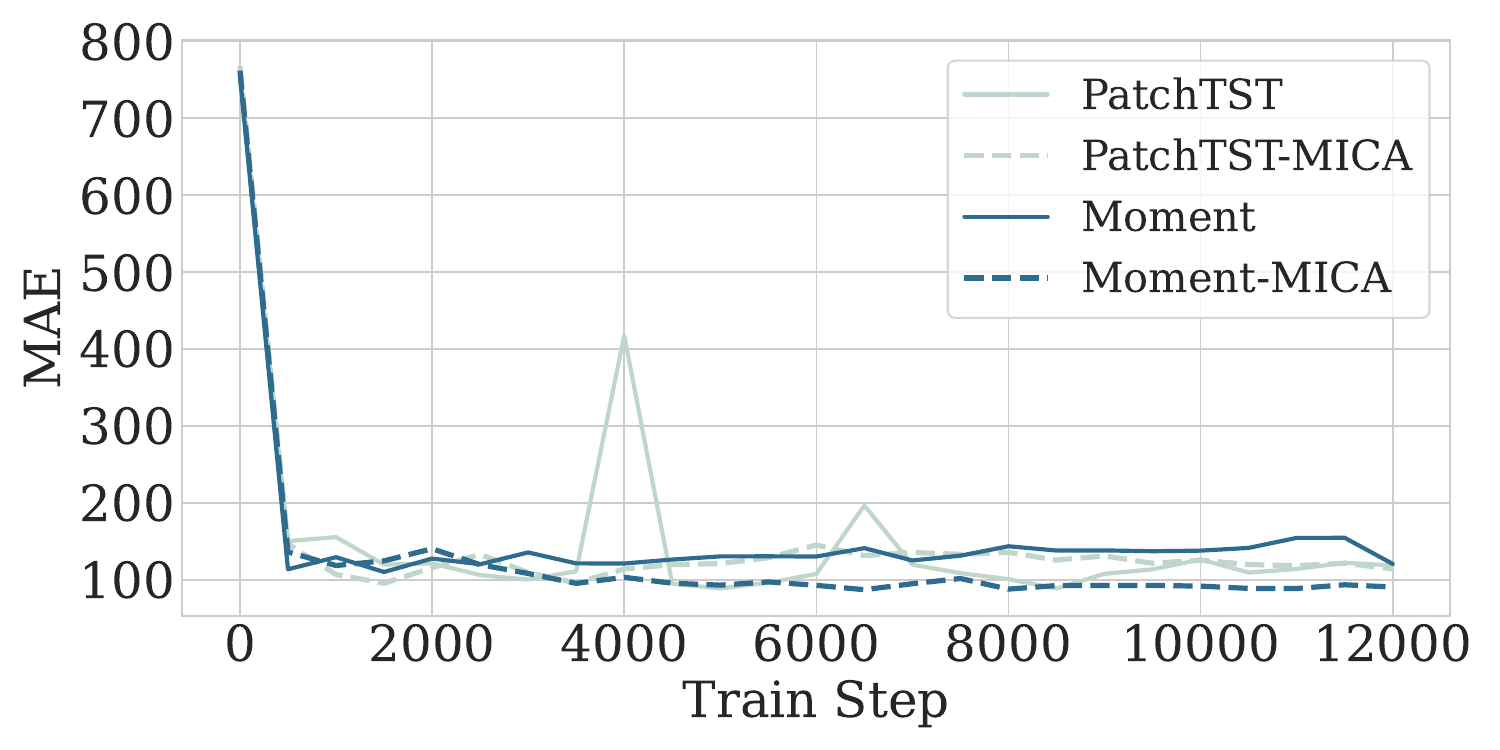}
        \par\vspace{2pt}
        {\small (b) COVID Deaths - Validation set}
    \end{minipage}
    
    \vspace{1em}
    
    \begin{minipage}[t]{0.38\textwidth}
        \centering
        \includegraphics[width=\textwidth, trim=0 15 0 0, clip]{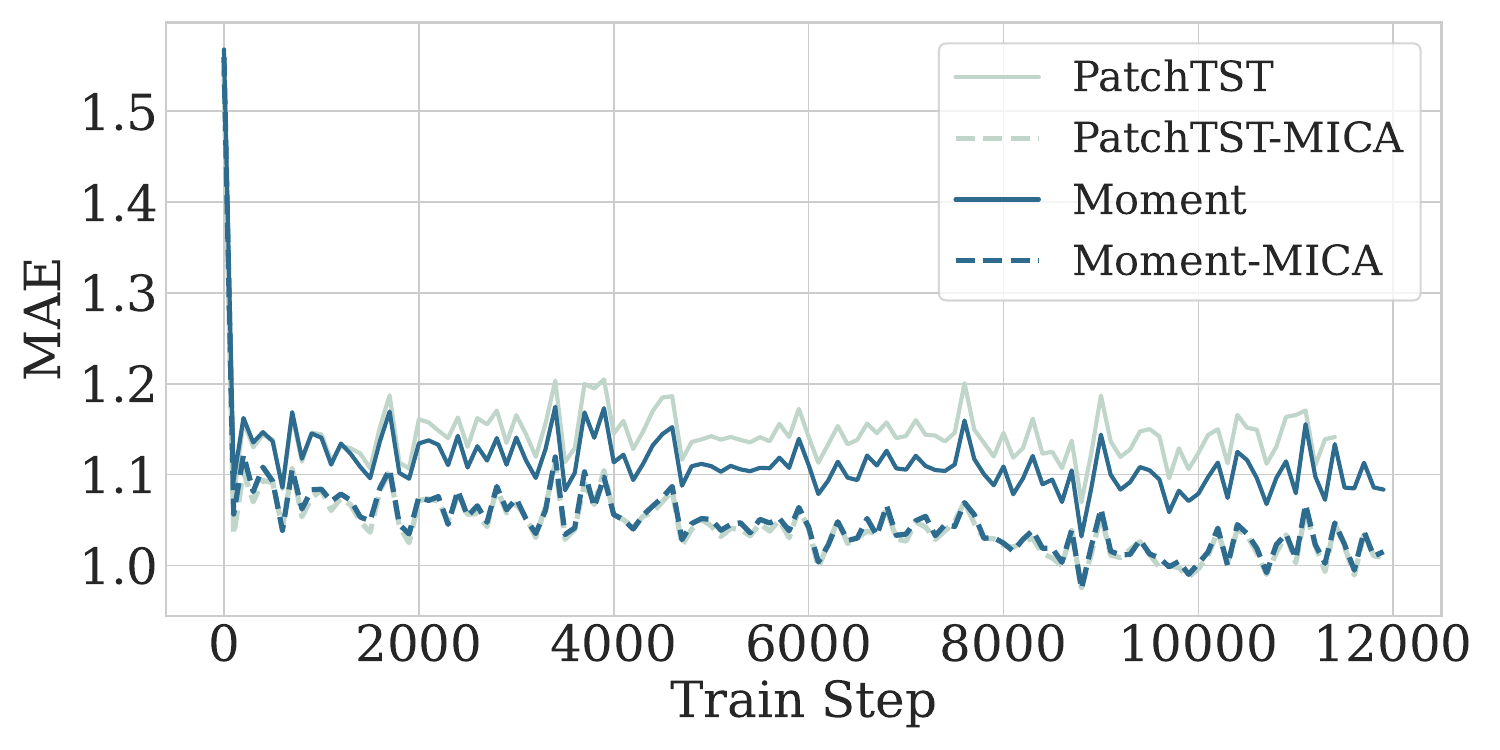}
        \par\vspace{2pt}
        {\small (c) Iowa IHOP SMEX02 - Train set}
    \end{minipage}
    \hspace{2em}
    \begin{minipage}[t]{0.38\textwidth}
        \centering
        \includegraphics[width=\textwidth, trim=0 15 0 0, clip]{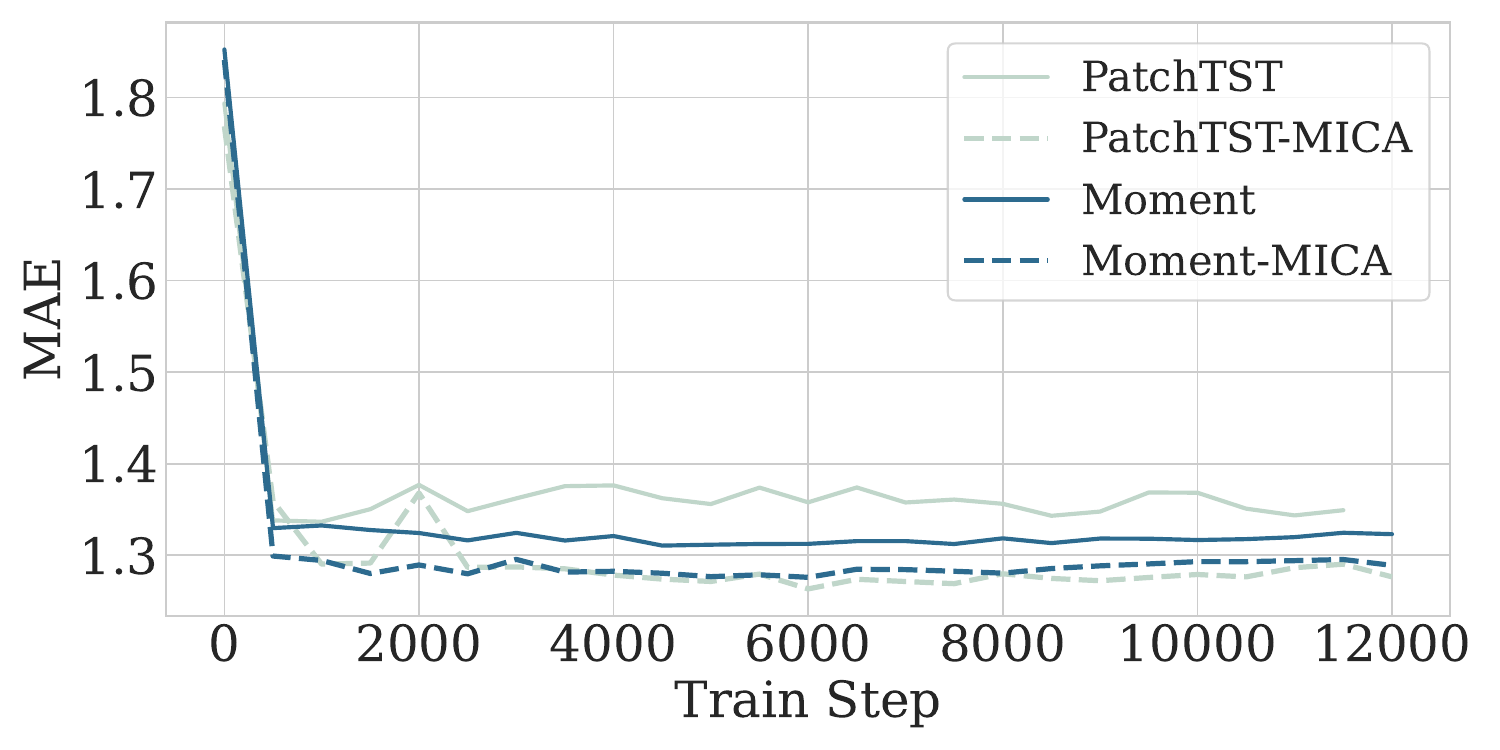}
        \par\vspace{2pt}
        {\small (d) Iowa IHOP SMEX02 - Validation set}
    \end{minipage}
    
    \vspace{1em}
    
    \begin{minipage}[t]{0.38\textwidth}
        \centering
        \includegraphics[width=\textwidth, trim=0 15 0 0, clip]{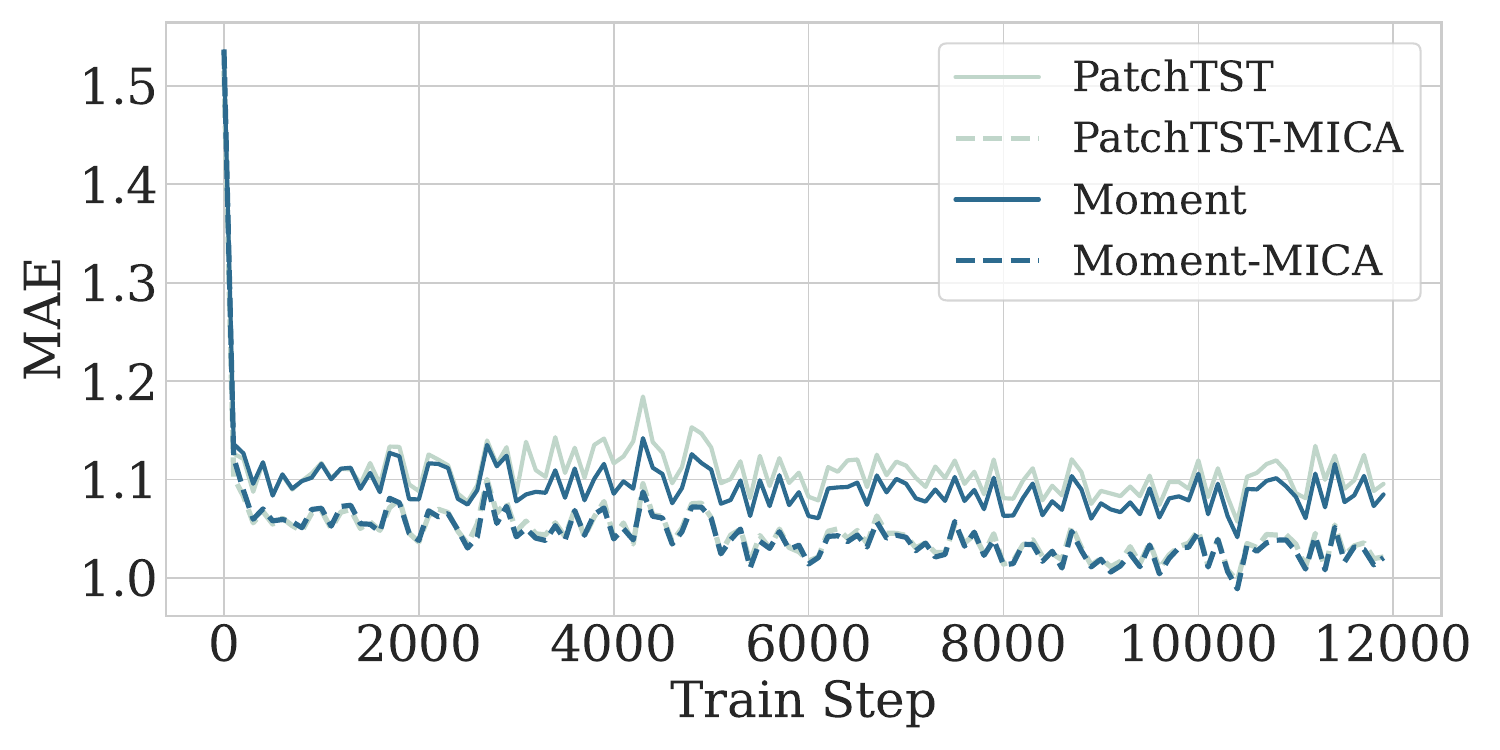}
        \par\vspace{2pt}
        {\small (e) Iowa PLOWS - Train set}
    \end{minipage}
    \hspace{2em}
    \begin{minipage}[t]{0.38\textwidth}
        \centering
        \includegraphics[width=\textwidth, trim=0 15 0 0, clip]{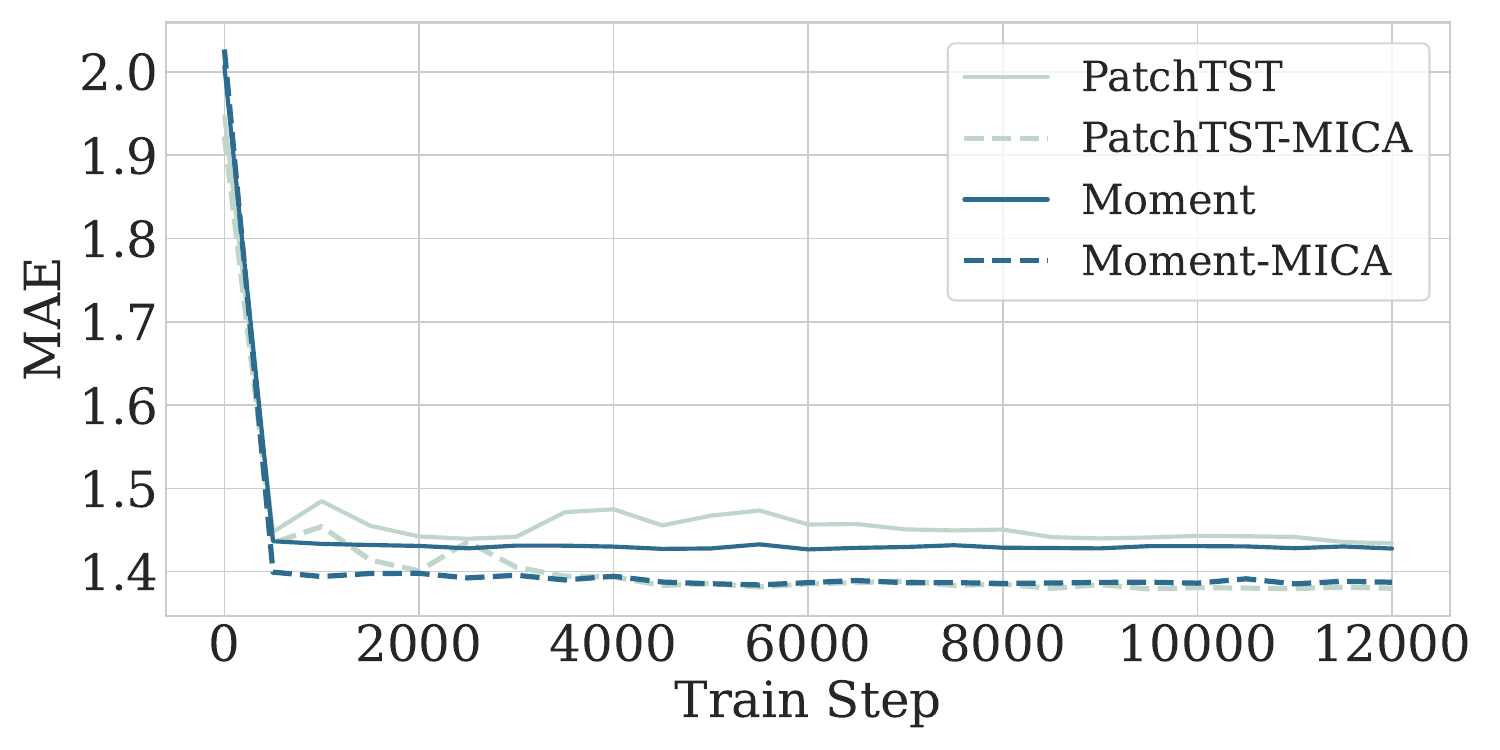}
        \par\vspace{2pt}
        {\small (f) Iowa PLOWS - Validation set}
    \end{minipage}
    
    \vspace{1em}
    
    \begin{minipage}[t]{0.38\textwidth}
        \centering
        \includegraphics[width=\textwidth, trim=0 15 0 0, clip]{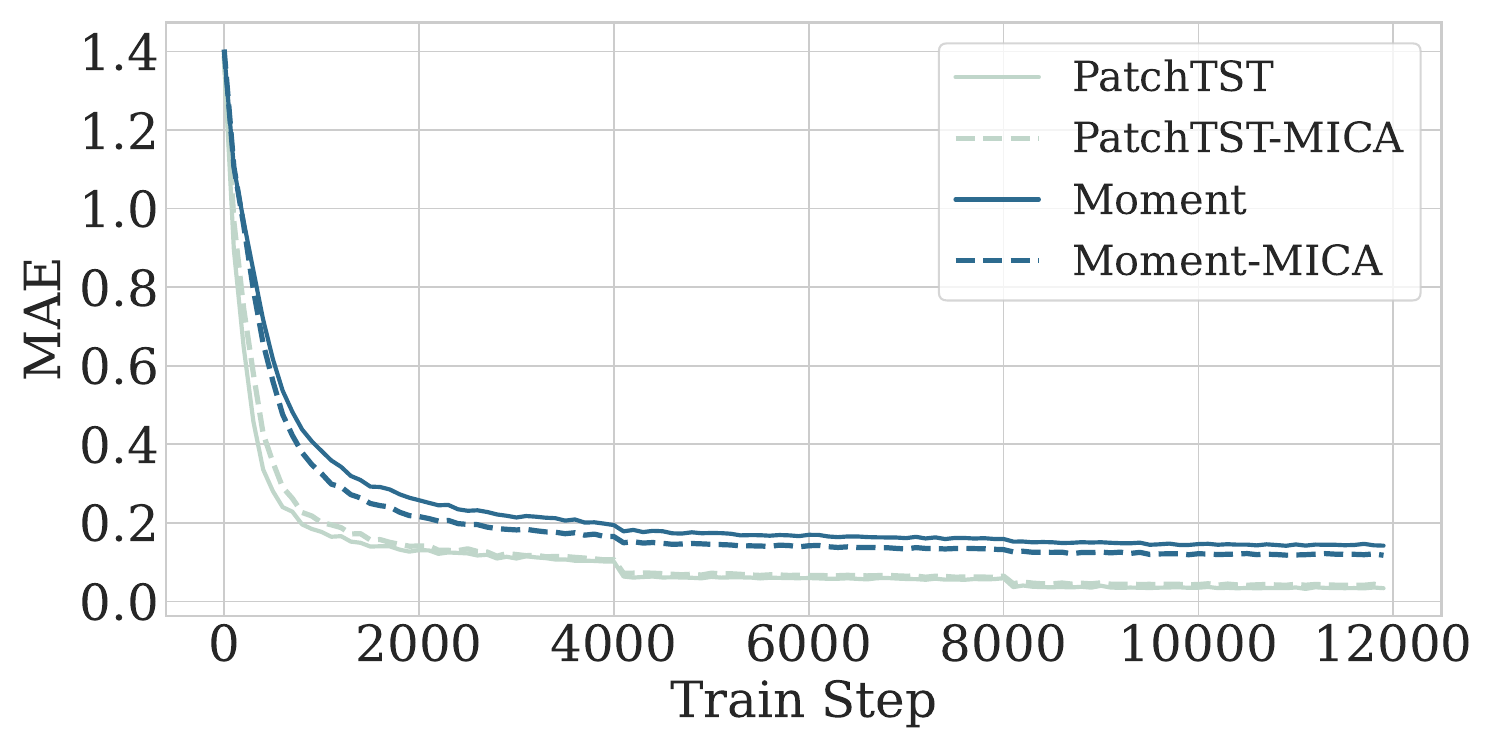}
        \par\vspace{2pt}
        {\small (g) ETT1 (Daily) - Train set}
    \end{minipage}
    \hspace{2em}
    \begin{minipage}[t]{0.38\textwidth}
        \centering
        \includegraphics[width=\textwidth, trim=0 15 0 0, clip]{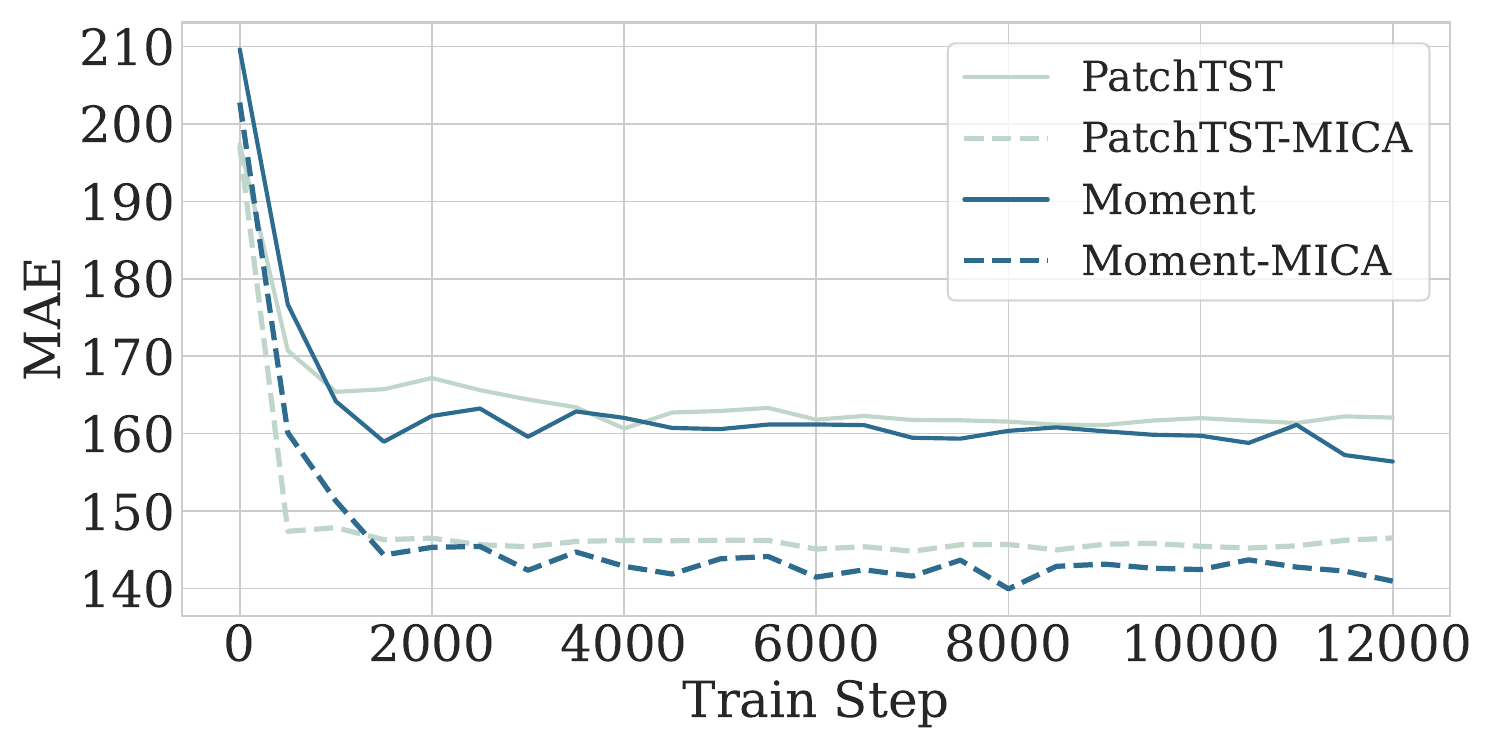}
        \par\vspace{2pt}
        {\small (h) ETT1 (Daily) - Validation set}
    \end{minipage}
    
    \vspace{1em}
    
    \begin{minipage}[t]{0.38\textwidth}
        \centering
        \includegraphics[width=\textwidth, trim=0 15 0 0, clip]{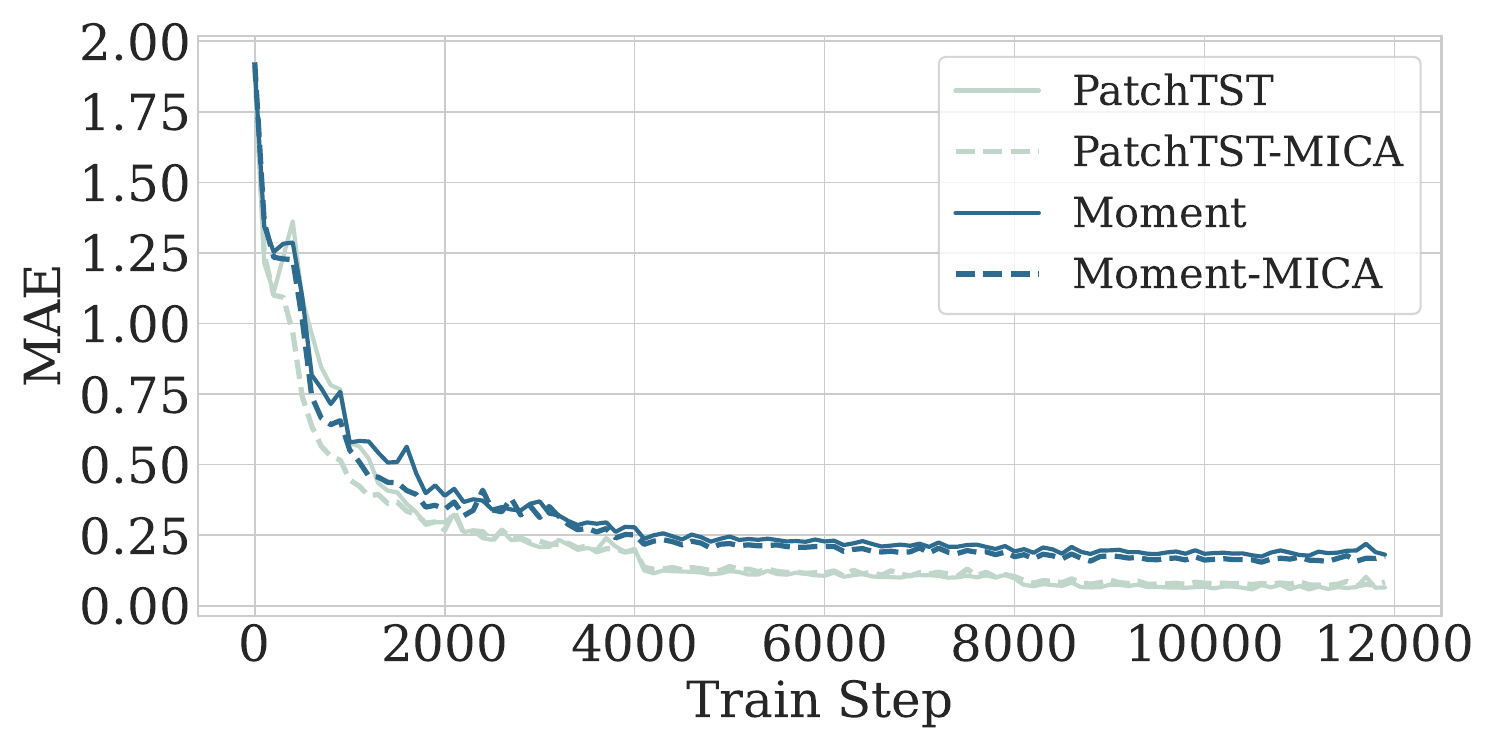}
        \par\vspace{2pt}
        {\small (i) ETT2 (Daily) - Train set}
    \end{minipage}
    \hspace{2em}
    \begin{minipage}[t]{0.38\textwidth}
        \centering
        \includegraphics[width=\textwidth, trim=0 15 0 0, clip]{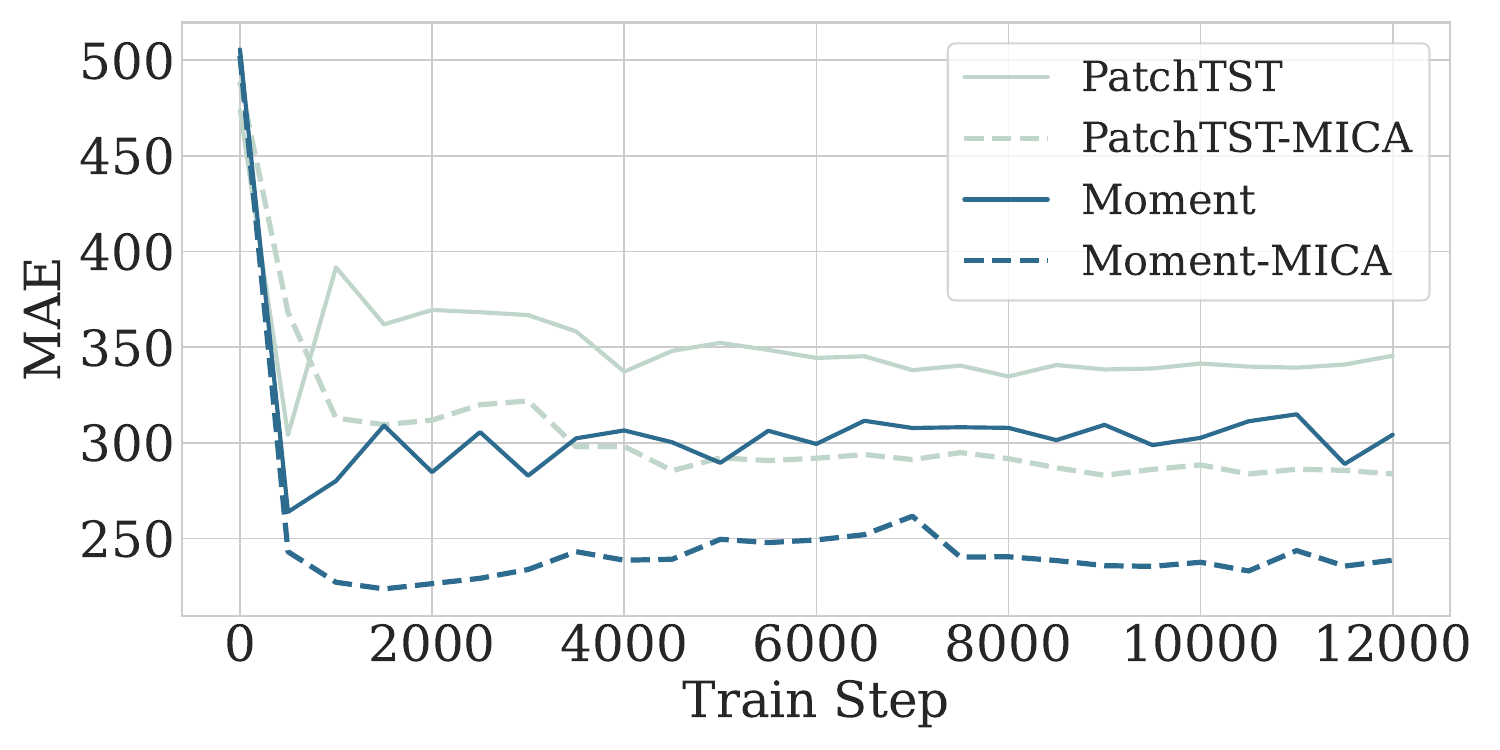}
        \par\vspace{2pt}
        {\small (j) ETT2 (Daily) - Validation set}
    \end{minipage}
    
    \caption{Training and validation loss (MAE) trajectories for a subset of datasets. Trajectories are averaged over random seeds at each training step. We compare \PatchTST\ and \Moment\ (solid lines) with their \MICA\ variants (MLP with query gate; dashed lines). The \MICA\ models generally achieve comparable or lower loss on both train and validation sets, indicating that cross-channel attention improves both optimization and generalization.}
    \label{fig:all_plots}
\end{figure}

\end{document}